\pdfoutput=1

\documentclass[11pt]{article}

\usepackage[]{acl}

\usepackage{times}
\usepackage{latexsym}

\usepackage[T1]{fontenc}

\usepackage[utf8]{inputenc}

\usepackage{microtype}

\usepackage{inconsolata}

\usepackage{hyperref}
\usepackage{url}
\usepackage{booktabs}       
\usepackage{amsfonts}       
\usepackage{nicefrac}       
\usepackage{microtype}      
\usepackage{xcolor}         
\usepackage{subfig}
\usepackage{amsmath}
\usepackage{multirow}
\usepackage{multicol}
\usepackage{wrapfig}
\usepackage[ruled,linesnumbered,boxed]{algorithm2e}
\usepackage{graphicx}
\usepackage{lipsum}
\usepackage{adjustbox}
\usepackage{subfig,graphicx}
\usepackage{tabularx}
\usepackage{array}
\usepackage{color}
\usepackage{colortbl}

\title{Learn or Recall? Revisiting Incremental Learning with Pre-trained Language Models}

\author{Junhao Zheng, Shengjie Qiu, Qianli Ma* \\
  School of Computer Science and Engineering, \\
  South China University of Technology, Guangzhou, China\\
  \texttt{junhaozheng47@outlook.com}, 
  \texttt{shengjieqiu6@gmail.com},
  \texttt{qianlima@scut.edu.cn}\thanks{*Corresponding author}}

\begin{document}
\maketitle
\addtocontents{toc}{\protect\setcounter{tocdepth}{0}}
\begin{abstract}
Incremental Learning (IL) has been a long-standing problem in both vision and Natural Language Processing (NLP) communities.
In recent years, as Pre-trained Language Models (PLMs) have achieved remarkable progress in various NLP downstream tasks, utilizing PLMs as backbones has become a common practice in recent research of IL in NLP.
Most assume that catastrophic forgetting is the biggest obstacle to achieving superior IL performance and propose various techniques to overcome this issue.
However, we find that this assumption is problematic.
Specifically, we revisit more than 20 methods on four classification tasks (Text Classification, Intent Classification, Relation Extraction, and Named Entity Recognition) under the two most popular IL settings (Class-Incremental and Task-Incremental) and reveal that most of them severely underestimate the inherent anti-forgetting ability of PLMs.
Based on the observation, we propose a frustratingly easy method called SEQ* for IL with PLMs.
The results show that SEQ* has competitive or superior performance compared with state-of-the-art (SOTA) IL methods yet requires considerably less trainable parameters and training time.
These findings urge us to revisit the IL with PLMs and encourage future studies to have a fundamental understanding of the catastrophic forgetting in PLMs.
The data, code and scripts are publicly available \footnote{\href{https://github.com/zzz47zzz/codebase-for-incremental-learning-with-llm}{https://github.com/zzz47zzz/codebase-for-incremental-learning-with-llm}}.
\end{abstract}

\begin{figure}
    \centering
    \includegraphics[width=\linewidth]{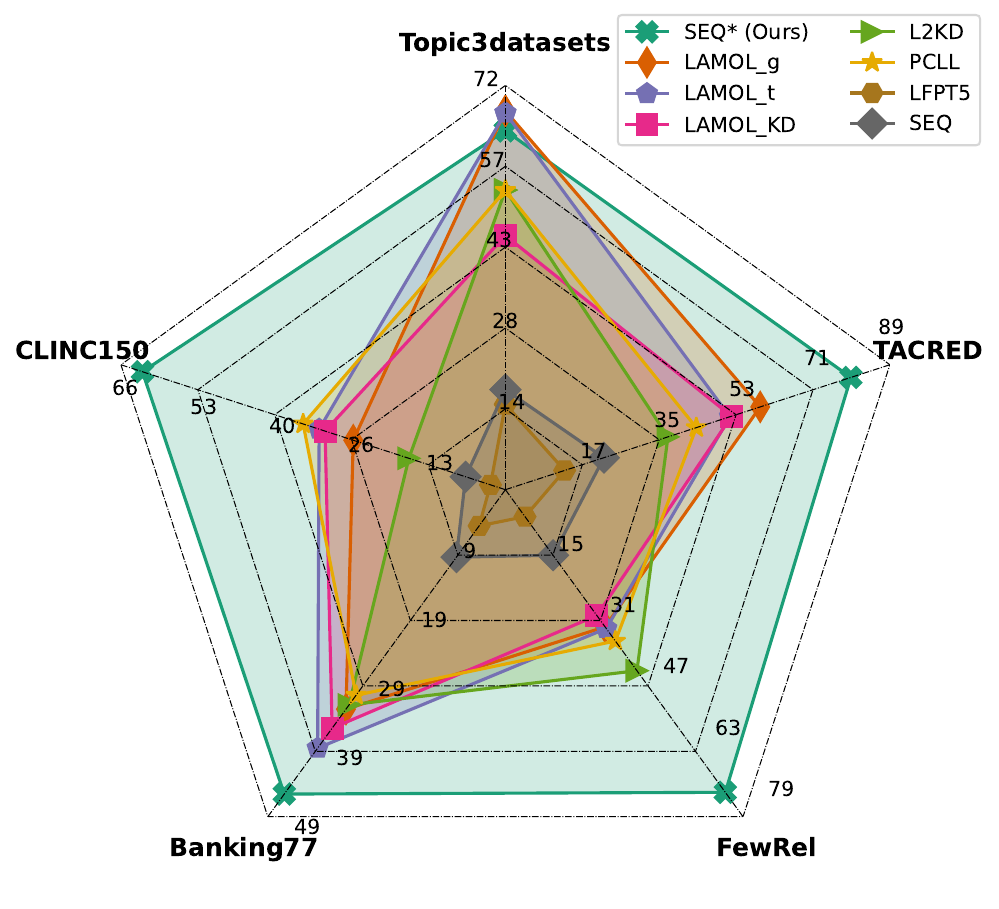}
    \caption{The comparison between the proposed SEQ* and SOTA IL methods on five class-incremental tasks. We report the average accuracy after learning the final task. The detailed results are provided in Table \ref{tab:sota_main_gen_pythia410m}.}
    \label{fig:radar_sota}
\end{figure}

\section{Introduction}
\label{sec:introduction}

Learning knowledge incrementally without much forgetting is an essential ability of human beings but still an unsolved challenge for neural networks in achieving human-level intelligence \cite{french1999catastrophic}.
Incrementally learning a sequence of tasks can be formulated into the paradigm of Incremental Learning (IL) and has been impeded by catastrophic forgetting \cite{kirkpatrick2017overcoming}.
Catastrophic forgetting refers to neural networks forgetting previous knowledge after learning new tasks \cite{mccloskey1989catastrophic}.

Recent years have witnessed significant breakthroughs in Pre-trained Language Models (PLMs) in vision and NLP tasks.
Most recent studies of IL use PLMs as the backbone and design various methods for alleviating catastrophic forgetting in NLP tasks.
However, is forgetting really catastrophic in PLMs? 
More specifically, how can we quantify forgetting and how much knowledge is forgotten in various IL scenarios when using various backbones and methods on various tasks?
More recently, \cite{tao2023can} reveal for the first time that BERT-like models have a strong anti-forgetting ability in the task-incremental setting. 
Why does this happen? 
Does it hold for a more challenging setting, such as class-incremental learning, and for other model architectures, such as GPT-like models?

To answer the above questions, we carry out extensive experiments to explore forgetting in more than 20 methods on four classification tasks (Text Classification, Intent Classification, Relation Extraction, and Named Entity Recognition) under the two most popular IL settings (Class-Incremental and Task-Incremental) with various model architecture (encoder only and decoder only) and scales (from 19M to 1.21B number of parameters).
Through extensive experiments, we have several major findings:
\begin{itemize}
    \item The popular assumption that PLMs suffer from catastrophic forgetting does not hold. Even under sequential fine-tuning (SEQ), the PLMs maintain the knowledge without much forgetting (Sec. \ref{sec:revisiting_forgetting_from_probing_is_lower_bound}). From the probing perspective, most existing IL methods do not learn incremental knowledge for PLMs (Sec. \ref{sec:comparing_sota_methods_with_SEQ*}). 
    \item By combining SEQ with simple strategies (Sec. \ref{sec:revisiting_sota_methods_boosting_SEQ}), we propose SEQ* and find that SEQ* has competitive or even superior performance than SOTA IL methods (Figure \ref{fig:radar_sota}, Sec. \ref{sec:comparing_sota_methods_with_SEQ*}). 
    \item The inherent anti-forgetting ability of PLMs comes from both the pre-training stage as well as the architecture of Transformer (Sec. \ref{sec:revisiting_forgetting_from_probing_pretraining}). Randomly initialised PLMs learn incrementally when SEQ is performed on a sequence of tasks.
    \item The forgetting of SEQ is due to the deviation of the classifier from the PLM rather than the loss of old knowledge in the PLM. (Sec. \ref{sec:revisiting_forgetting_from_probing_what_is_forgetten}).
\end{itemize}
Our study urges the NLP community to revisit and deepen the understanding of the forgetting in PLMs.

\section{Experimental Settings}
\label{sec:experimental_settings}

\subsection{Problem formulation}
\label{sec:experimental_settings_problem_formulation}
Formally, the goal of IL is to learn a model $f_\theta:\mathbf{x}\rightarrow y \in \mathcal{Y}$ from a sequence of tasks  $\mathcal{D}=\{\mathcal{D}_1,\mathcal{D}_2,\cdots,\mathcal{D}_T\}$, where the $t$-th task $\mathcal{D}_t = \{ (\mathbf{x}_i^t,y_i^t)\}_{i=1}$ contains input samples $\mathbf{x}_i^t \in \mathcal{X}_t$ and labels $y_i^t \in \mathcal{Y}_t$.
In Class-Incremental Learning (CIL), the label sets of different tasks are exclusive: $\mathcal{Y}_1 \cap \mathcal{Y}_2 \cdots \mathcal{Y}_T = \emptyset$, and the task id is unknown during inference.
In Task-Incremental Learning (TIL), the label sets of different tasks may be overlapping: $\mathcal{Y}_1 \cap \mathcal{Y}_2 \cdots \mathcal{Y}_T \neq \emptyset$, and the task id is required during inference.
In general, CIL is much more challenging than TIL because PLMs suffer from inter-task forgetting much more seriously than intra-task forgetting \cite{tao2023can}.
Appendix \ref{sec:appendix_problem_formulation} provides detailed description and evaluation metrics.

\subsection{Tasks and Datasets}
\label{sec:experimental_settings_tasks_datasets}
We consider four types of downstream tasks in our experiments:
Text classification, intent classification, relation extraction, and named entity recognition.
We use the following eight datasets: Topic3Datasets (containing AGNews, DBPedia, and YaHoo \cite{zhang2015character}) for text classification; CLINC150 \cite{larson-etal-2019-evaluation} and Backing77 \cite{casanueva-etal-2020-efficient} for intent classification; FewRel \cite{han-etal-2018-fewrel} and TACRED \cite{zhang-etal-2017-position} for relation extraction; OntoNotes5 \cite{hovy-etal-2006-ontonotes}, I2B2 \cite{murphy2010serving}, Few-NERD \cite{ding-etal-2021-nerd} for named entity recognition.
Detailed descriptions are provided in Appendix \ref{sec:appendix_datasets}.

\subsection{Backbones}
\label{sec:experimental_settings_backbones}
We consider two popular architectures as backbone PLMs: encoder-only and decoder-only.
For encoder-only backbones, we use bert-base-cased and bert-large-cased \cite{devlin-etal-2019-bert}, the most popular choices in previous IL studies.
The encoder-only backbones are typically used as discriminant models, and linear layers are added for downstream tasks. 
We use the GPT2 \cite{radford2019language} and Pythia suite \cite{biderman2023pythia} for decoder-only backbones.
Pythia is based on GPT-NeoX \cite{black-etal-2022-gpt}, which contains 8 model sizes and 154 pre-training checkpoints, enabling research in interpretability and learning dynamics.
The decoder-only backbones are typically used as generative models, and no additional linear layers are required since the output target is natural language.
The detailed description is provided in Appendix \ref{sec:appendix_backbones}.

\begin{figure*}
    \centering
    \includegraphics[width=\linewidth]{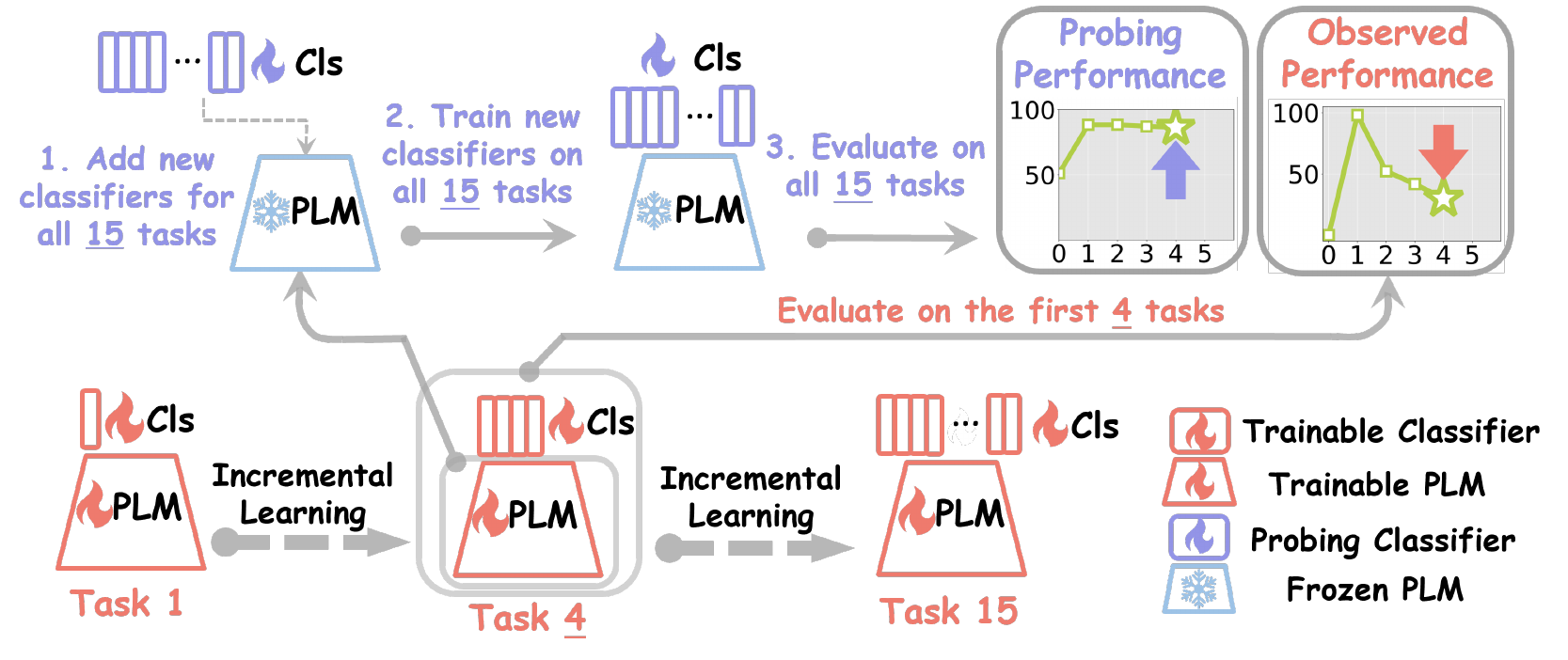}
    \caption{An illustration of how we obtain the probing and the observed performance of the model when learning the fourth task if there are a total of 15 tasks. The observed performance is used as a metric of forgetting in existing studies. The probing performance indicates how the encoder forgets. However, it is overlooked by previous studies.}
    \label{fig:probing_illustration}
\end{figure*}

\section{Revisiting the Forgetting from the Probing Perspective}
\label{sec:revisiting_forgetting_from_probing}

\subsection{How to Measure the Forgetting in PLMs?}
\label{sec:revisiting_forgetting_from_probing_how_to_measure}
This subsection describes how to measure the forgetting inside PLMs during IL.
Specifically, we utilize the probing technique, an effective method to evaluate the representation ability of backbones on target tasks \cite{chenforgetting,tao2023can,davari2022probing,wu2021pretrained}.

To probe the knowledge in PLMs for all tasks in IL, we add \textit{probing classifiers} on top of the PLM and train the probing classifiers on \textit{all tasks} in IL.
Then, we evaluate the PLM and the probing classifiers on all tasks and obtain the \textit{probing performance}.
The probing performance is the upper bound performance when the classifiers do not forget.
For clarity, the performance evaluated with the original model is called the \textit{observed performance}.
We note that measuring probing performance will not affect the training process of IL since the backbone PLM is frozen when training probing classifiers. 
Furthermore, the original classifiers only predict the classes of \textit{learned} tasks, while the probing classifiers predict the classes of \textit{all} tasks in IL.
We provide an illustration in Figure \ref{fig:probing_illustration} and the formal definition in Appendix \ref{sec:appendix_problem_formulation}.

We consider four metrics for our probing study: linear probing, cosine linear probing, prototype probing, and cosine prototyping.
Linear probing has been widely adopted in previous works.
In linear probing, the probing classifier is a linear layer.
Cosine linear probing adopts a cosine linear layer as the probing classifier.
Specifically, the logits are computed as the cosine similarities between classifier weights and extracted features.
\citet{hou2019learning} show that utilizing cosine linear layers mitigates bias towards new classes in IL.
In prototype probing, the probing classifier is a linear layer whose weight matrix is calculated as the class feature centres.
Previous IL studies \cite{zhou2023revisiting,chen-etal-2023-consistent,ma-etal-2023-learning} show that using class feature centres as prototypes for classification is effective.
Cosine prototype probing further utilizes cosine normalization when calculating logits.
Further discussion is provided in Appendix \ref{sec:appendix_probing_study_four_metrics_introduction}.

\subsection{Is Sequential Fine-tuning Really the Lower Bound?}
\label{sec:revisiting_forgetting_from_probing_is_lower_bound}
Sequential fine-tuning (SEQ) has long been regarded as the lower bound of IL.
In this subsection, we revisit SEQ from the probing perspective, and we find that SEQ is severely underestimated when using PLMs for IL.

The backbone was small and randomly initialized in early studies exploring IL \cite{kirkpatrick2017overcoming,french1999catastrophic,mccloskey1989catastrophic}.
They find that SEQ usually results in models forgetting all previous knowledge when learning new tasks.
Recent IL studies in NLP \cite{razdaibiedina2023progressive,zheng2024concept,huang-etal-2021-continual,sun2019lamol,qiu2024incremental} also observe that SEQ leads to worse performance.
However, in the era of PLMs, fine-tuning has proven to be effective for adapting PLMs to different domains or downstream tasks \cite{aghajanyan2020better,devlin-etal-2019-bert,radford2018improving,zheng-etal-2023-preserving}.
If fine-tuning really causes PLMs to forget nearly all previous knowledge in IL, it should also cause PLMs to forget all pre-trained knowledge when adapting to new tasks.
Obviously, this assumption is not true since fine-tuning is still effective for PLMs \cite{OpenAI2023GPT4TR}.
Furthermore, prior research \cite{tao2023can} also finds that BERT-like models suffer from little forgetting under the task-incremental setting by the probing study.

\begin{figure}[!t]
    \centering
    \subfloat[Observed Performance]{
        \includegraphics[width=0.49\linewidth]{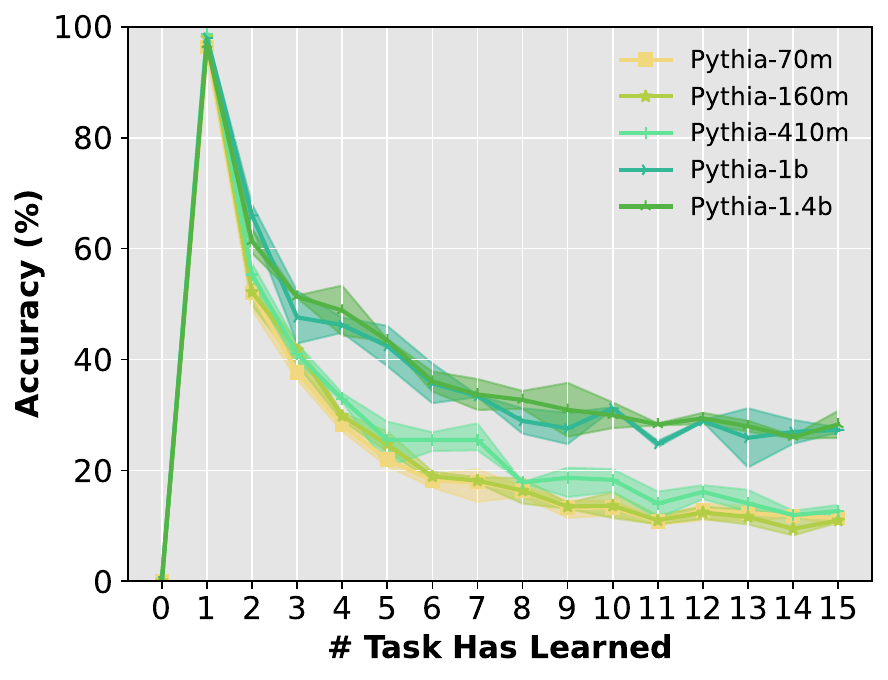}
        \label{fig:probing_CIL_IC_GEN_a}
    }
    \subfloat[Training Loss]{
        \includegraphics[width=0.49\linewidth]{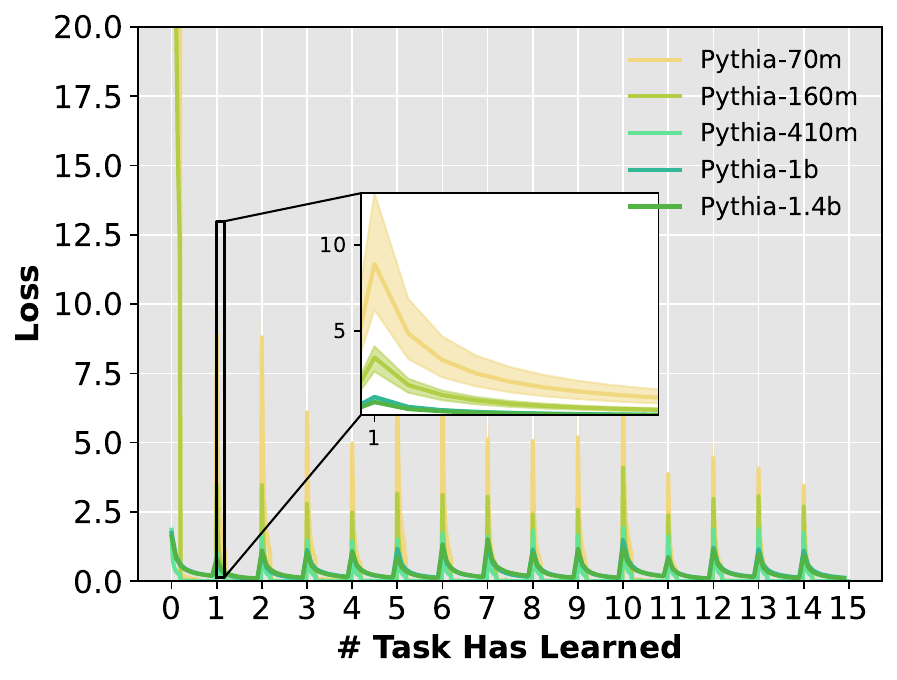}
        \label{fig:probing_CIL_IC_GEN_b}
    }
    
    \subfloat[Lin. Prob]{
        \includegraphics[width=0.49\linewidth]{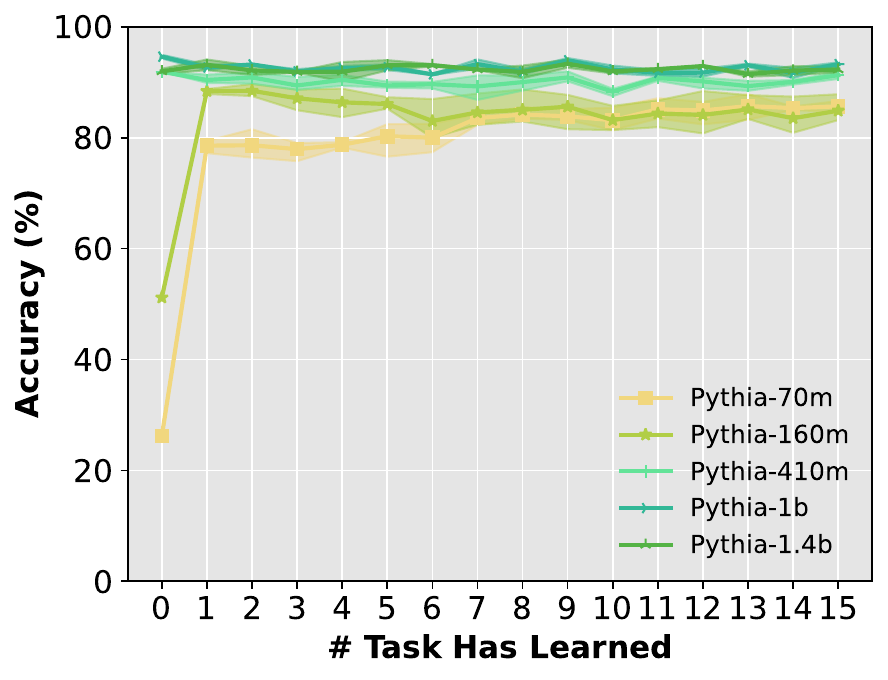}
        \label{fig:probing_CIL_IC_GEN_c}
    }
    \subfloat[Cos.Lin. Prob]{
        \includegraphics[width=0.49\linewidth]{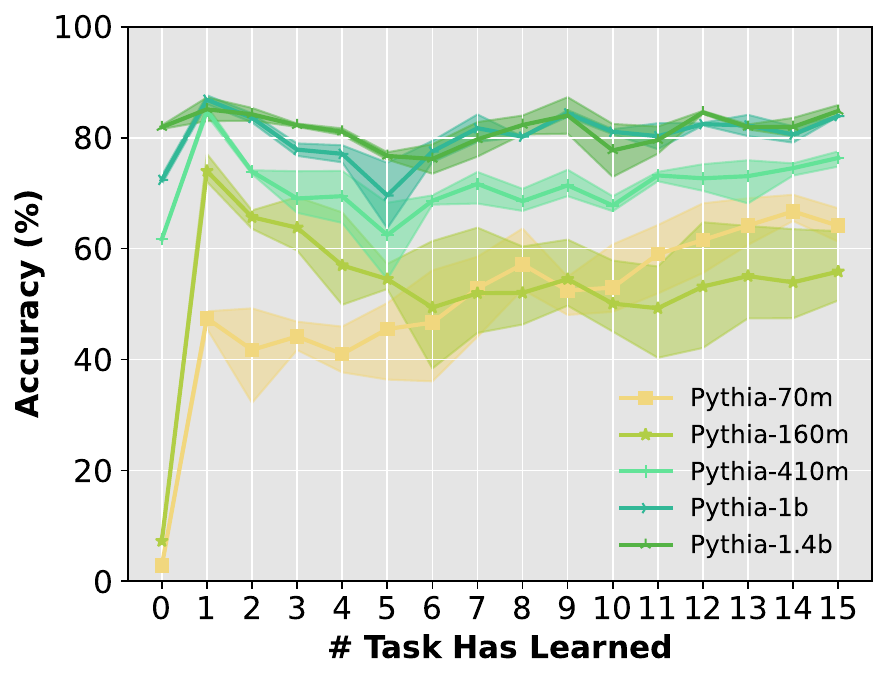}
        \label{fig:probing_CIL_IC_GEN_d}
    }
    
    \subfloat[Proto. Prob]{
        \includegraphics[width=0.49\linewidth]{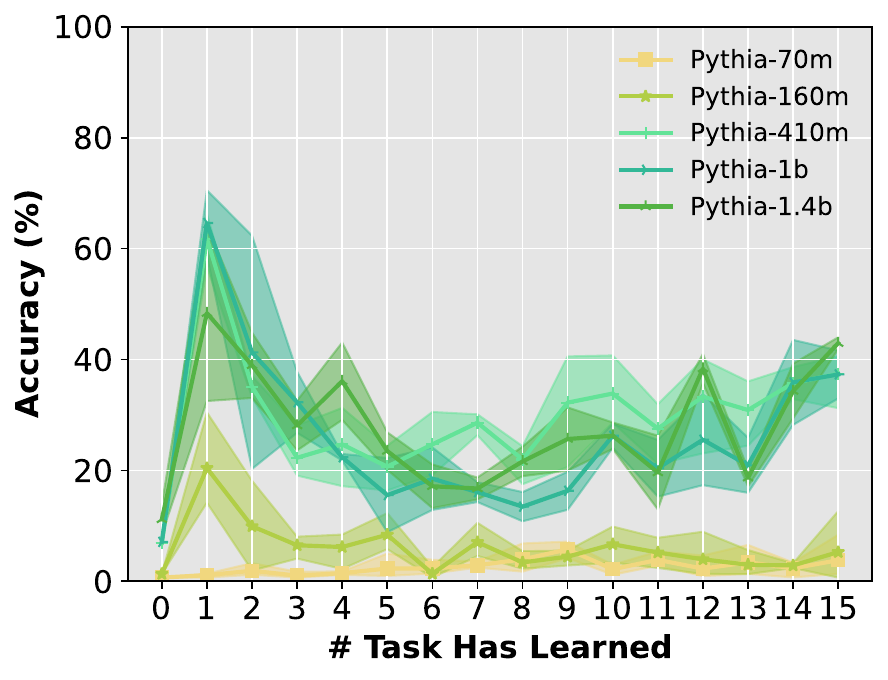}
        \label{fig:probing_CIL_IC_GEN_e}
    }
    \subfloat[Cos.Proto. Prob]{
        \includegraphics[width=0.49\linewidth]{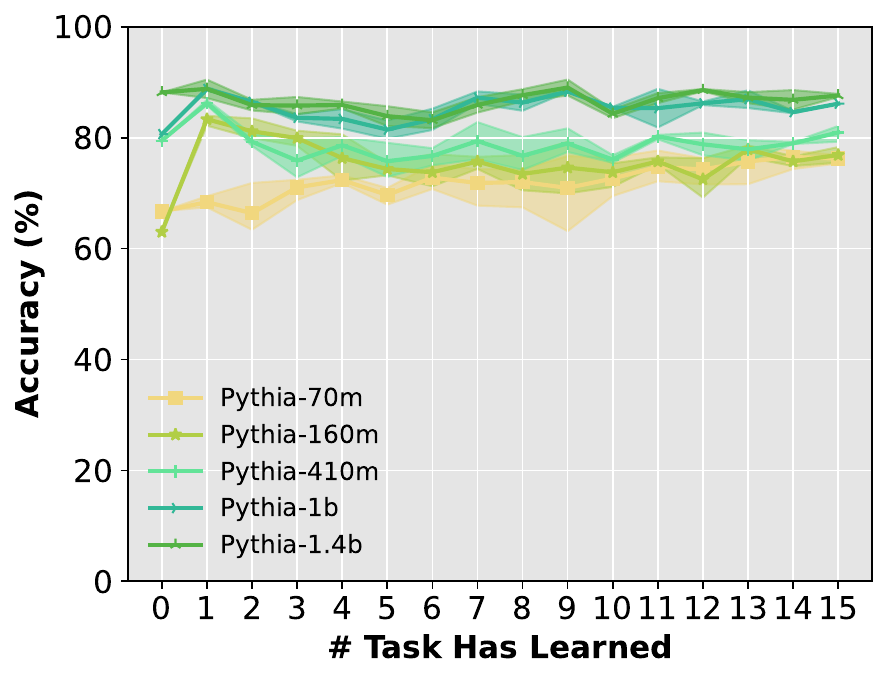}
        \label{fig:probing_CIL_IC_GEN_f}
    }
    \caption{The observed and probing performance on Class-Incremental Intent Classification. The dataset is CLINC150. The backbones are generative models. (a)(b) are the observed performance and training loss during IL training. (c)-(f) are the probing performance when different metrics are adopted.}
    \label{fig:probing_CIL_IC_GEN}
\end{figure}

The observed and probing performance on class-incremental intent classification with generative models are summarized in Figure \ref{fig:probing_CIL_IC_GEN}.
The results on other IL settings, downstream tasks and backbones are in Appendix \ref{sec:appendix_probing_study_four_metrics_result}.
Figure \ref{fig:probing_CIL_IC_GEN_a}  shows that the observed performance drops dramatically from approximately 98\% to 10\% as more new tasks are learned, in line with our understanding of catastrophic forgetting.
However, Figure \ref{fig:probing_CIL_IC_GEN_c} describes an entirely different phenomenon.
The PLMs achieve high probing performance after learning the first task.
And the linear probing performance has barely decreased since the second task.
In other words, PLMs preserve the knowledge to classify all 15 tasks even when adapting to only new tasks sequentially.
This phenomenon is contradictory to what we know about catastrophic forgetting and SEQ.

Indeed, the probing performance is high since all tasks' data is available when training the probing classifiers, while the observed performance is poor since the original classifiers only train on the data from the current task.
However, with the above observation, we can boost the observed performance with simple strategies, which will be described in Section \ref{sec:revisiting_sota_methods_boosting_SEQ}.

\subsection{What is the Best Metric for Probing Performance?}
\label{sec:revisiting_forgetting_from_probing_best_metric}
In Figure \ref{fig:probing_CIL_IC_GEN}, we find that the ranking of four probing metrics is as follows: linear > cosine linear, cosine prototype > prototype probing.
This subsection will explain why linear probing is the best metric for probing study.

\begin{figure}[htbp]
    \centering
    \subfloat[Feature Sim]{
        \includegraphics[width=0.49\linewidth]{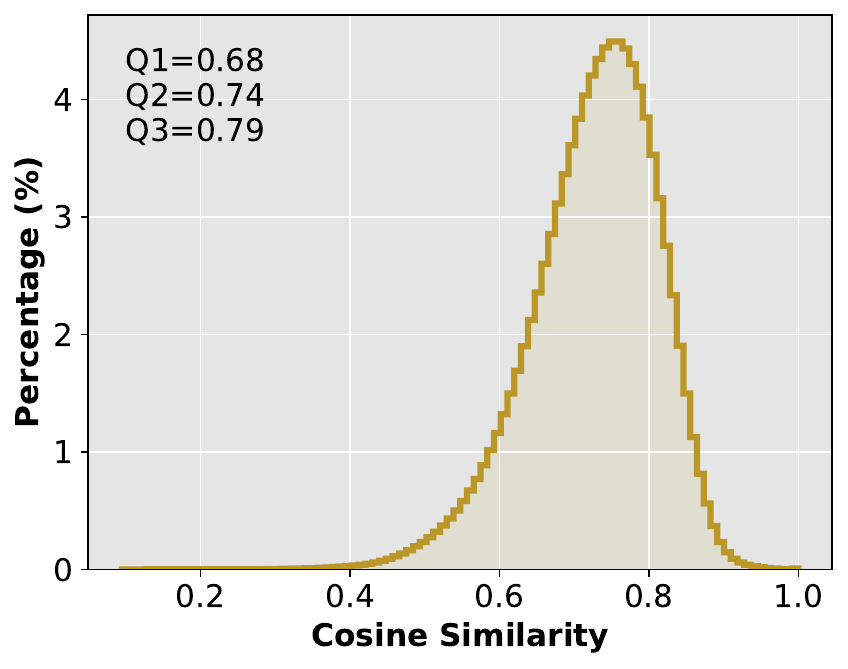}
        \label{fig:feature_embed_hist_a}
    }
    \subfloat[Feat-WordEmbed Sim]{
        \includegraphics[width=0.49\linewidth]{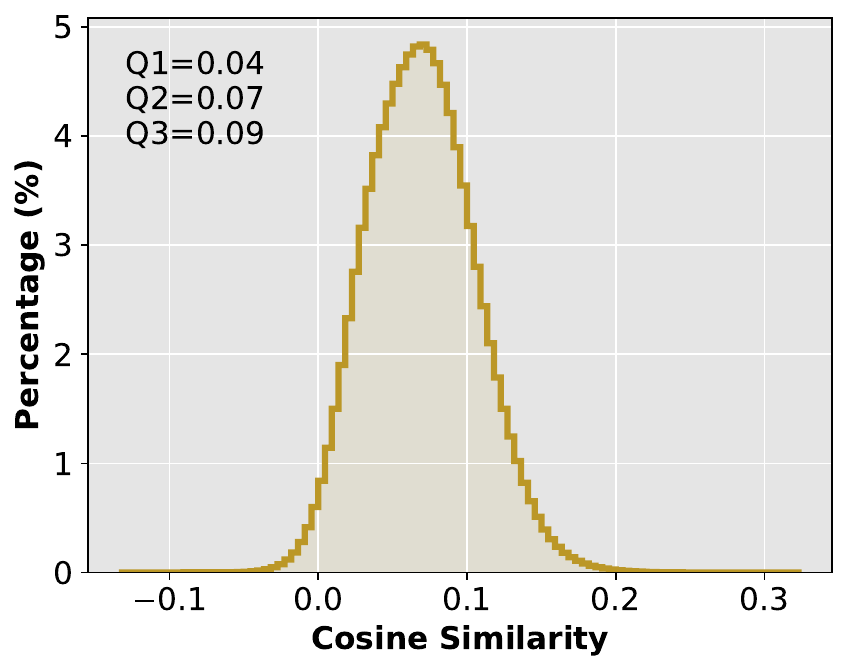}
        \label{fig:feature_embed_hist_b}
    }

    \subfloat[Feature Norm]{
        \includegraphics[width=0.49\linewidth]{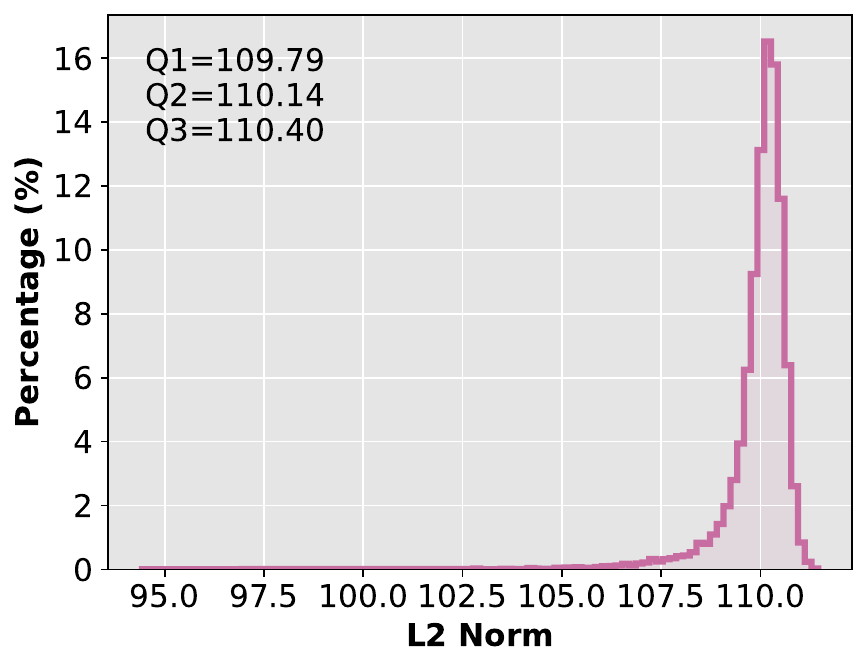}
        \label{fig:feature_embed_hist_c}
    }
    \subfloat[WordEmbed Norm]{
        \includegraphics[width=0.49\linewidth]{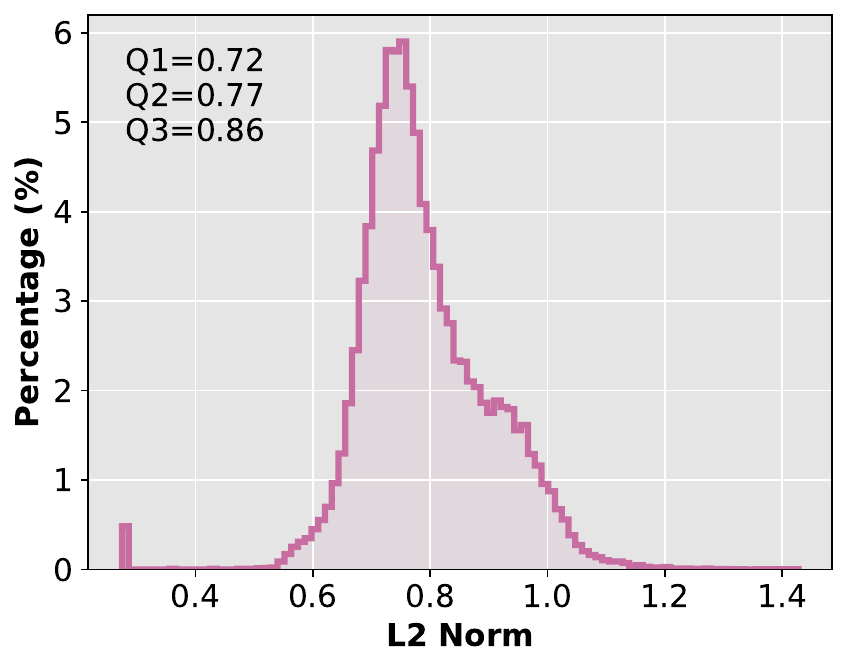}
        \label{fig:feature_embed_hist_d}
    }
    
    \caption{The histogram of features and different embeddings of Pythia-410m. The features are calculated on the training set of CLINC150, and the output word embeddings are loaded from pre-trained weights. The class embeddings refer to the row vectors of the weight matrix in the probing classifier on CLINC150. The class prototypes refer to the class feature centres estimated on the training set of CLINC150.}
    \label{fig:feature_embed_hist}
\end{figure}

First, we need to understand what the features (i.e., last hidden states), word embeddings of PLMs, and the class embeddings in probing classifiers ``look like'' respectively.
The detailed description is in Appendix \ref{sec:appendix_probing_study_four_metrics_analysis}.
The histograms of the L2 norm and the cosine similarity of features, word embeddings and class embeddings are in Figure \ref{fig:feature_embed_hist}.
Figure \ref{fig:feature_embed_hist_a} shows that the features occupy a narrow cone in the vector space rather than being uniform in all directions, which has been discussed in \citep{ethayarajh-2019-contextual}.
More surprisingly, Figure \ref{fig:feature_embed_hist_b} shows that the learned (output) word embeddings are nearly orthogonal to the features.
We infer that the cross-entropy loss encourages all word embeddings except the ground truth one to become farther away from the feature during pre-training.
In other words, the cross-entropy loss encourages a large difference in logits, and the word embeddings to be orthogonal to the features in order to distinguish logits better.
Therefore, it is not surprising that linear probing has the best performance, considering that the word embedding layer is essentially a linear layer.
From this point of view, it is also not surprising that the performance of prototype probing is poor since the prototypes (class feature centres) also fall in the narrow cone space, and it is not an optimal solution for distinguishing logits.

Then, why does cosine normalization degrade the performance of linear probing but improve prototype probing?
Figure \ref{fig:feature_embed_hist_c} and \ref{fig:feature_embed_hist_d} are the L2 norm of the features and word embedings.
We find that the norm of word embeddings has a larger discrepancy than features.
It indicates that the norm of word embeddings contains the prior knowledge obtained from pre-training. 
Therefore, the cosine linear probing ignores the difference in the norm of features and thus has poorer performance compared with linear probing.
For prototype probing, the prototype falls in a narrow cone space, and the similarity between the prototype and features is large and close to each other.
In this case, cosine normalization can eliminate the interference of the norm and establish the relationship between logits and cosine similarity between features.
We provide the detailed analysis and full results with different backbones in Appendix \ref{sec:appendix_probing_study_four_metrics_analysis}. An illustration of different types of probing metrics is in Figure \ref{fig:metrics_illustration}.

\begin{figure}[htbp]
    \centering
    \subfloat[Intent+Before SEQ]{
        \includegraphics[width=0.49\linewidth]{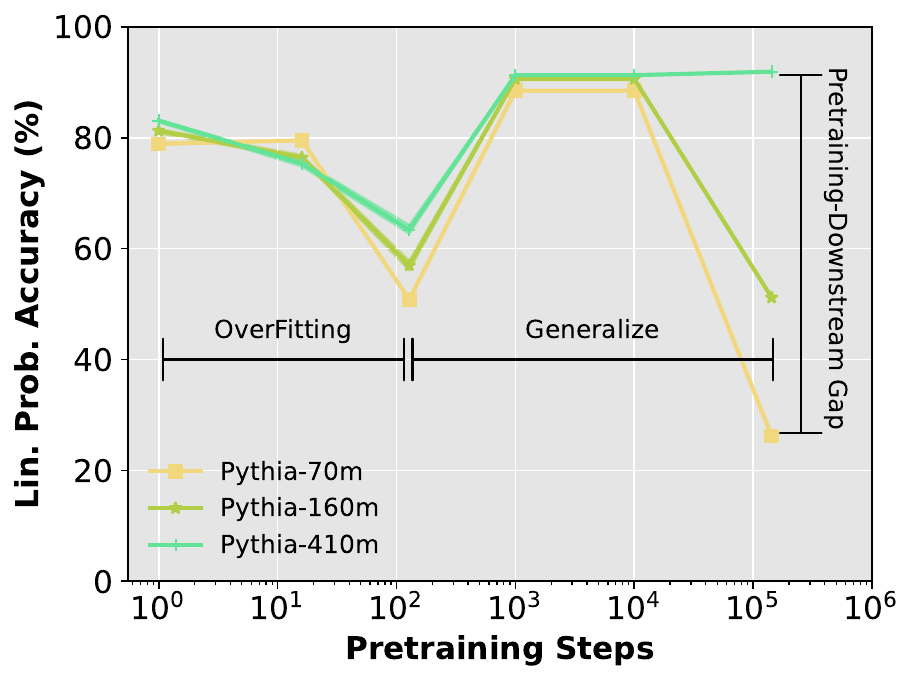}
        \label{fig:probing_pretraining_a}
    }
    \subfloat[Intent+After SEQ]{
        \includegraphics[width=0.49\linewidth]{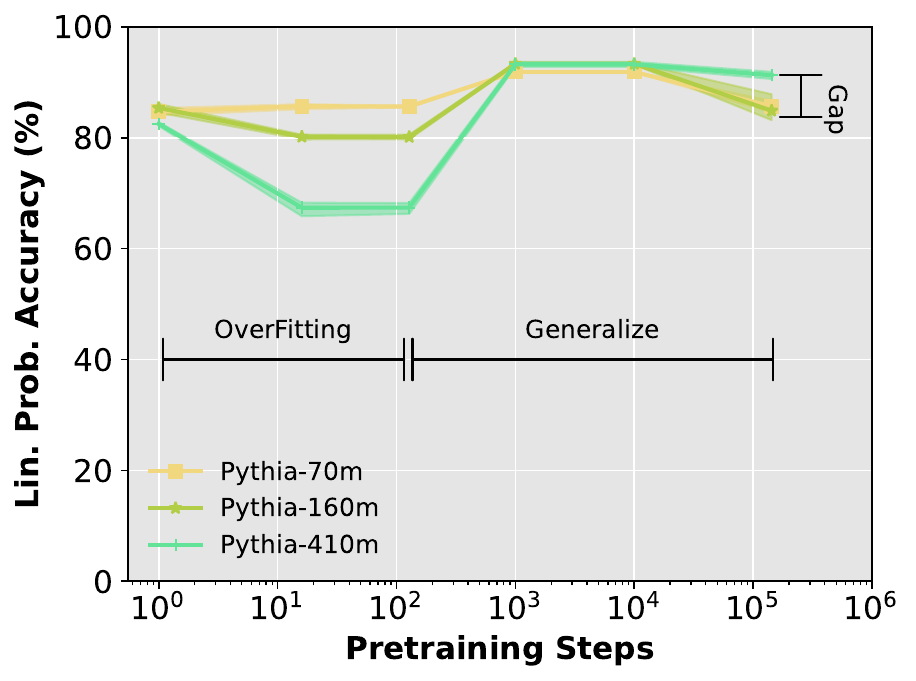}
        \label{fig:probing_pretraining_b}
    }

    \subfloat[RE+Before SEQ]{
        \includegraphics[width=0.49\linewidth]{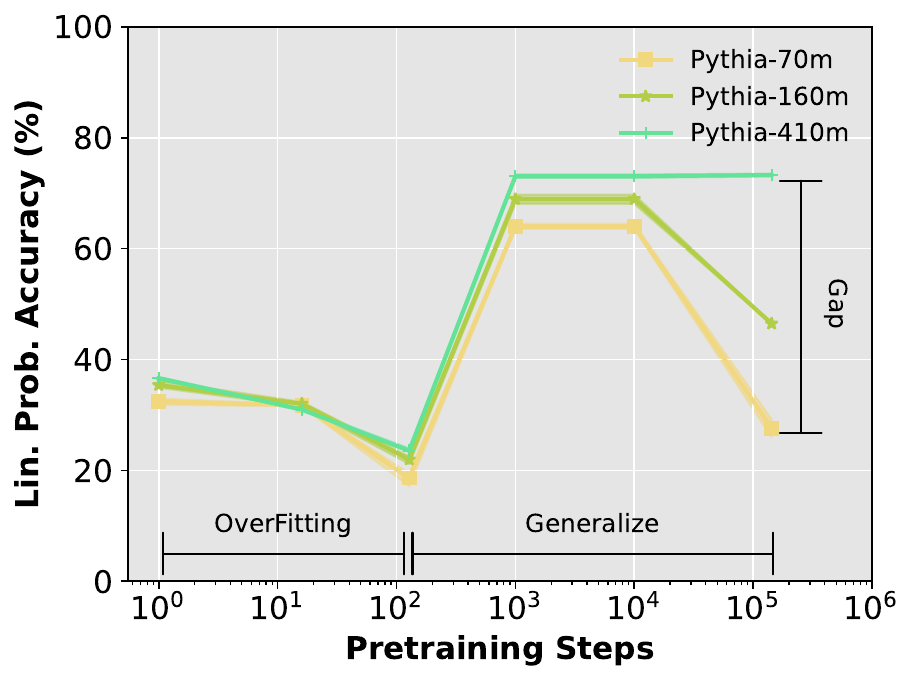}
        \label{fig:probing_pretraining_c}
    }
    \subfloat[RE+After SEQ]{
        \includegraphics[width=0.49\linewidth]{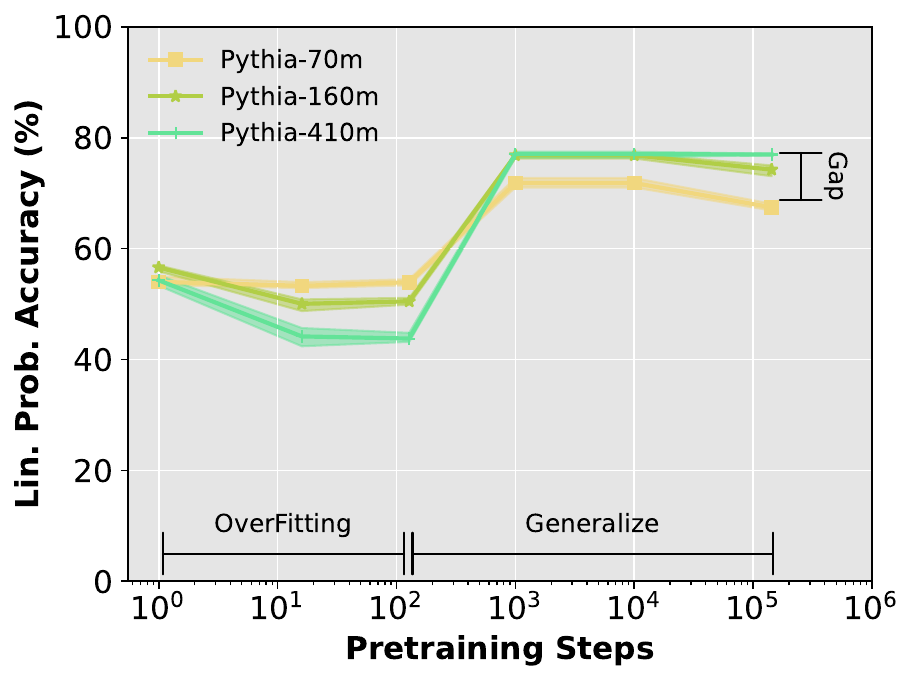}
        \label{fig:probing_pretraining_d}
    }
    
    \caption{The linear probing performance on checkpoints with different pre-training steps. (a) and (b) are evaluated before and after incremental learning using SEQ. ``Intent'' and ``RE'' represent the model is evaluated on the Class-Incremental Intent Classification or Relation Extraction.}
    \label{fig:probing_pretraining}
\end{figure}

\subsection{What is the Role of Pre-training in IL?}
\label{sec:revisiting_forgetting_from_probing_pretraining}
In this subsection, we reveal that the key to the anti-forgetting ability of PLMs lies in both the Transformers' architecture and the pre-training knowledge.

We evaluate the linear probing performance on checkpoints with a different number of pre-training steps: \{0,16,128,1k,10k,143k(final)\}.
We load the pre-trained checkpoints (or randomly-initialized checkpoints at step 0) and evaluate their linear probing performance before and after IL using SEQ.
Figure \ref{fig:probing_pretraining} shows two main phases in pre-training: overfitting and generalization. 
In the first phase (step 0 - step 128), the model memorizes the pre-training corpus, and the linear probing performance decreases. 
In the second phase (step 1k - step 143k), the model gradually learns the pre-training knowledge and the linear probing performance increases. 
However, when the model further generalizes to the pre-training corpus (step 10k - step 143k), the linear probing performance of small backbones (Pythia-70 m and 160m) decreases again due to the gap between pre-training and downstream tasks.
This gap can be eliminated when adapting to downstream tasks (Figure \ref{fig:probing_pretraining_a} and Figure \ref{fig:probing_CIL_IC_GEN_c}).
For larger backbones (Pythia-410 m, 1b, and 1.4b), the model can be adapted to new tasks directly without this gap.

Besides, we have the following interesting findings:
(1) Pre-training indeed improves the linear probing performance in IL (Figure \ref{fig:probing_pretraining_b} and \ref{fig:probing_pretraining_d}).
(2) Apart from pre-training, the architecture of the Transformer is also a key factor in the high linear probing accuracy during SEQ. 
When the downstream task is relatively simple, such as intent classification, even the randomly-initialized models achieve high linear probing performance (Figure \ref{fig:probing_pretraining_b}).
Pre-training brings considerable improvements when the downstream task is more complex, such as relation extraction (Figure \ref{fig:probing_pretraining_d}).
(3) More surprisingly, SEQ improves the linear probing performance of models from nearly all pre-training steps (Figure \ref{fig:probing_pretraining_a} v.s. \ref{fig:probing_pretraining_b}; Figure \ref{fig:probing_pretraining_c} v.s. \ref{fig:probing_pretraining_d}).
This shows that Transformers' architecture can incrementally absorb new knowledge even when just sequential fine-tuning on new tasks.
The detailed settings, visualization of features, and additional results on text classification are provided in Appendix \ref{sec:appendix_role_of_pretraining}.

\begin{figure}[htbp]
    \centering
    \subfloat[Observed Classifier]{
        \includegraphics[width=0.49\linewidth]{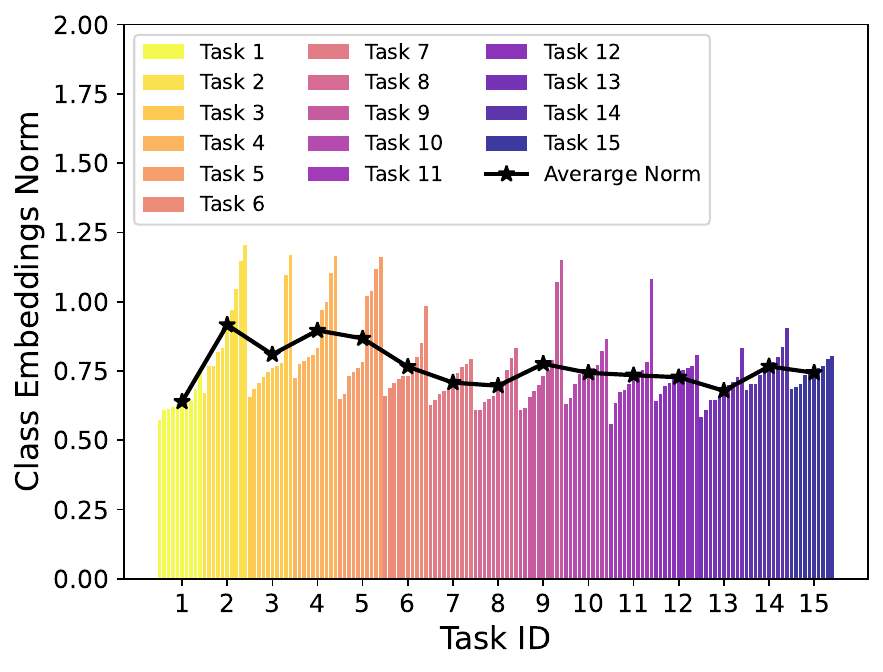}
        \label{fig:comparison_class_norm_moving_distance_a}
    }
    \subfloat[Probing Classifier]{
        \includegraphics[width=0.49\linewidth]{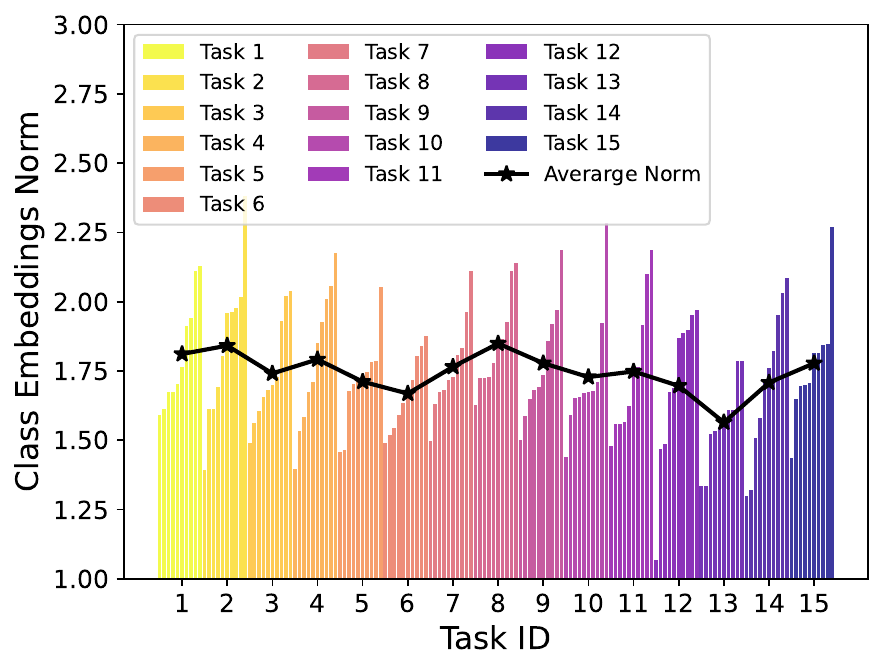}
        \label{fig:comparison_class_norm_moving_distance_b}
    }

    \subfloat[Observed Classifier]{
        \includegraphics[width=0.49\linewidth]{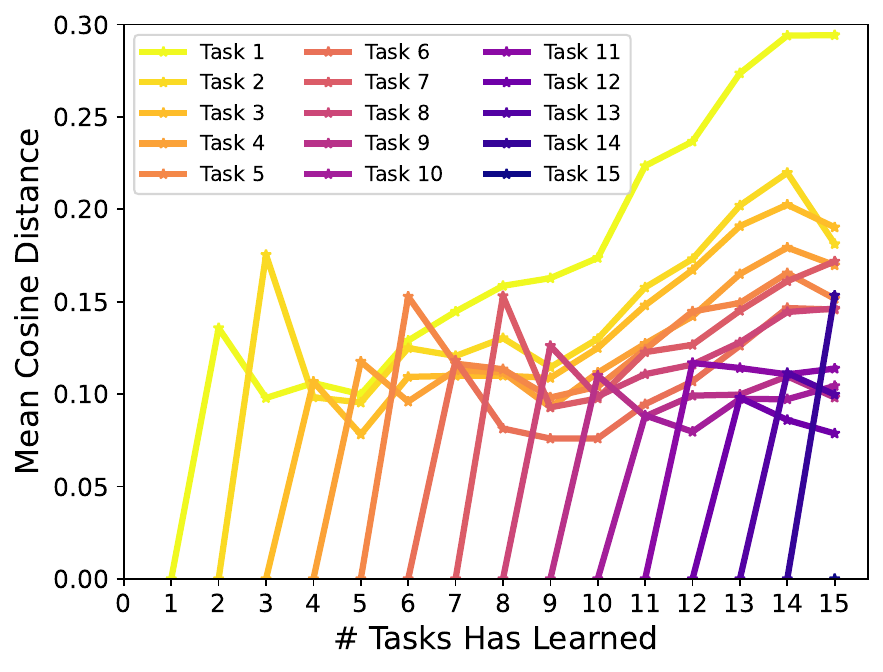}
        \label{fig:comparison_class_norm_moving_distance_c}
    }
    \subfloat[Probing Classifier]{
        \includegraphics[width=0.49\linewidth]{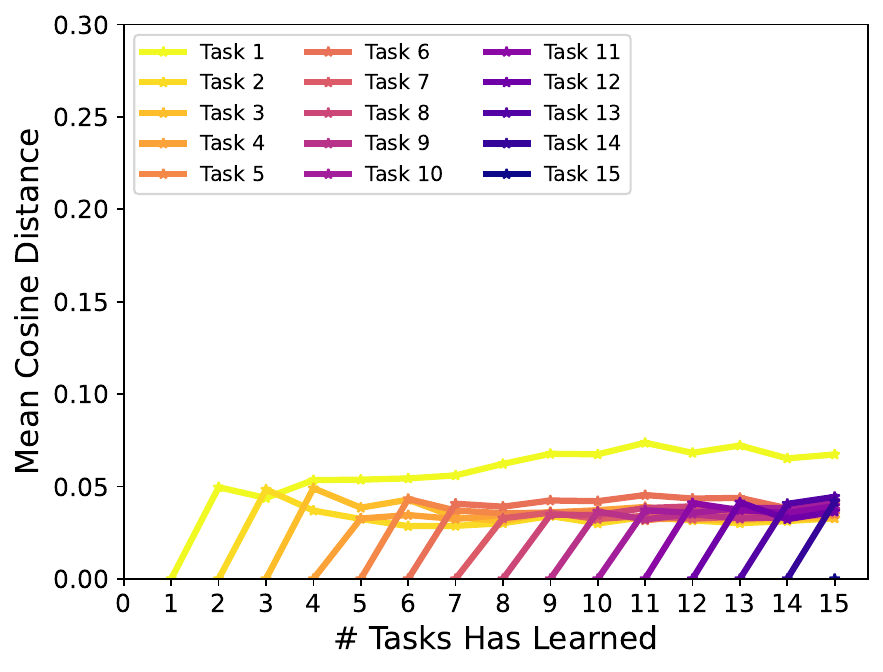}
        \label{fig:comparison_class_norm_moving_distance_d}
    }
    
    \caption{Comparison between the observed linear classifier and the linear probing classifier during SEQ on class-incremental intent classification. The backbone is Pythia-410m. (a)(b) show the average norm of the class embeddings of each task; (c)(d) show the average moving distance of the class embeddings of each task.}
    \label{fig:comparison_class_norm_moving_distance}
\end{figure}

\subsection{What is really forgotten in SEQ?} 
\label{sec:revisiting_forgetting_from_probing_what_is_forgetten}
As discussed in Sec. \ref{sec:revisiting_forgetting_from_probing_is_lower_bound}, SEQ's linear probing performance showed little degradation or even improvement across all settings.
Therefore, the reason for forgetting must lie in the classifier.
In this subsection, we will take a closer look at the forgetting in classifiers.

Existing studies \cite{wu2019large,hou2019learning,zheng2024balancing} find that the model tends to predict new classes during IL.
They refer this phenomenon as the class imbalance problem between old and new classes.
In SEQ, the class imbalance problem is more severe since only new classes are learned.
We also observe that the logits of new classes are much larger than those of old classes in a SEQ model. 
Because both features and class embeddings determine the magnitude of logits, the features occupy a narrow cone space, and their norms are relatively close, we can infer that the forgetting is caused by either (1) the norm of class embeddings or (2) the cosine similarity between features and class embeddings.

For the first reason (i.e., class norm), we compare the class embedding norm between learned linear classifiers and linear probing classifiers in Figure \ref{fig:comparison_class_norm_moving_distance_a} and \ref{fig:comparison_class_norm_moving_distance_b}.
Surprisingly, the class embedding norm of new tasks is not larger than those of old tasks in the observed classifier of SEQ.
It indicates that the class norm is not the primary reason for the forgetting in SEQ.

For the second reason (i.e., cosine similarity), we compare the moving distance of class embeddings between the observed and probing classifiers in Figure \ref{fig:comparison_class_norm_moving_distance_c} and \ref{fig:comparison_class_norm_moving_distance_d}.
The moving distance of the class embeddings of task $t$ at task $t+k$ is computed as follows:
(1) When the model finishes training on task $t$, we compute the cosine distance between all pairs of class embeddings from task $t$ and class feature centres from \textit{all tasks} and obtain a cosine similarity matrix $\mathbf{C}_{t}^{t}$.
(2) When model finishes training on task $t+k$, we compute the cosine distance between all pairs of class embeddings from task $t$ and class feature centres from \textit{all tasks} and obtain a cosine similarity matrix $\mathbf{C}_{t+k}^{t}$.
(3) Then, the moving distance of task $t$'s class embeddings is calculated as the average absolute difference between the cosine similarity matrix $\mathbf{C}_{t}^{t}$ and $\mathbf{C}_{t+k}^{t}$.
The moving distance measures how the class embeddings move relatively to \textit{all} class feature centres since they have been learned.
If the classifier does not forget a class, the distance from its class embeddings to all class feature centres should remain constant.
In other words, its moving distance will be zero if the classifier does not forget how to classify this class with the features extracted by PLMs.
We provide an illustration in Figure \ref{fig:moving_distance_illustration}.
The detailed settings, definition, and additional results with frozen bert-large-cased are provided in Appendix \ref{sec:appendix_forgetting_in_classifiers}.

Figure \ref{fig:comparison_class_norm_moving_distance_c} and \ref{fig:comparison_class_norm_moving_distance_d} show that the class embeddings of observed classifiers change significantly compared with those of probing classifiers.
It indicates that the forgetting happens because the old class embeddings are pushed away from their initial and optimal position.
The cosine similarity matrices are visualized in Figure \ref{fig:cos_distance_matrix_examples}.

\section{Revisiting the SOTA Methods in IL\label{sec:revisiting_sota_methods}} 

\subsection{SEQ*: Boosting the Performance of SEQ} 
\label{sec:revisiting_sota_methods_boosting_SEQ}
In this subsection, we propose SEQ* based on the findings about the forgetting in SEQ. 

In the previous section, we have the following findings about SEQ:
(F1) The PLMs do not learn new knowledge in SEQ about the downstream IL tasks;
(F2) The PLMs achieve the highest probing performance once being adapted to downstream tasks, and there is little performance degradation when learning on more new tasks (See Figure \ref{fig:probing_CIL_IC_GEN_c} and Figure \ref{fig:probing_pretraining_a});
(F3) The classifier forgets dramatically, while the PLMs do not. The reason is that the class embeddings are pushed away from the initial learned optimal position.

Therefore, we propose the following strategies for closing the gap between the probing and observed performance in SEQ:
(S1) Freeze the PLMs after warm-up;
(S2) Freeze the old classifiers when learning new tasks;
(S3) Use cosine linear classifiers only when no old data is available in a CIL scenario. Otherwise, use linear classifiers;
(S4, optional) Pre-allocate future classifiers.
We call the method with the above strategies as SEQ*, and an illustration is provided in Figure \ref{fig:illustration_seqstar}.

The rationale for the above strategies is as follows:
(S1) is proposed according to (F1).
Furthermore, we propose to warm up (i.e., full-parameter fine-tuning) PLMs in only the first task according to (F2).
In practice, warm-up onlt for 1-3 epochs brings considerable improvement across backbones and datasets.
(S1) and (S2) preserve the relative position of class embeddings with respective to class feature centres to avoid the issue in (F3).
When both PLMs and classifiers of old tasks are frozen, only the norm of new class embeddings may lead to the biased prediction towards new classes.
Because cosine linear layers are not optimal for exploiting PLMs' knowledge, we propose (S3) to avoid bias prediction.
In other words, we use linear classifiers for the TIL scenario and the CIL scenario where old data is stored.
Finally, we propose (S4) for better forward compatibility \cite{zhou2022forward}.
(S4) is marked as an optional strategy since it requires additional information on the number of total tasks.
Therefore, we report the two variants of SEQ*, i.e., w/ and w/o (S4), when comparing with SOTA methods.
We provide detailed discussion and explanation in Appendix \ref{sec:appendix_strategies_for_SEQ}.

\begin{table}[!t]
  \centering
  \caption{Comparison between SOTA methods and SEQ* on sentence-level classification tasks. The backbone is Pythia-410m. The IL scenario is CIL. No old samples are stored for all models. \textit{Lin}: use linear classifiers; \textit{Cos}: use cosine linear classifiers; \textit{FixB}: fix backbone PLMs; \textit{FixC}: fix old classifiers; \textit{FixBC}: fix both backbone PLMs and old classifiers; \textit{W}: warm up backbone PLMs; \textit{P}: pre-allocate future classifiers. The best and second best results are \textbf{bold} and \underline{underlined}. The full result is in Table \ref{tab:sota_full_pythia410m_sentence}.}
  \resizebox{\linewidth}{!}{
    \begin{tabular}{lccccc}
    \toprule
          & \textbf{Topic3Datasets} & \textbf{CLINC150} & \textbf{Banking77} & \textbf{FewRel} & \textbf{TACRED} \\
\cmidrule{2-6}          & \boldmath{}\textbf{$\mathcal{A}_T$}\unboldmath{} & \boldmath{}\textbf{$\mathcal{A}_T$}\unboldmath{} & \boldmath{}\textbf{$\mathcal{A}_T$}\unboldmath{} & \boldmath{}\textbf{$\mathcal{A}_T$}\unboldmath{} & \boldmath{}\textbf{$\mathcal{A}_T$}\unboldmath{} \\
    \midrule
    \textbf{LFPT5} & 16.78  & 3.48  & 7.98  & 5.52  & 7.60  \\
    \textbf{L2KD} & 58.89  & 22.48  & 47.47  & 37.08  & 20.86  \\
    \textbf{LAMOL\_KD} & 49.94  & 41.99  & 52.60  & 25.77  & 29.03  \\
    \textbf{LAMOL\_g} & \textbf{74.45 } & 35.43  & 48.40  & 28.10  & 32.70  \\
    \textbf{LAMOL\_t} & \underline{74.05} & 43.37  & \underline{57.00} & 28.44  & 28.81  \\
    \textbf{PCLL} & 58.83  & 47.09  & 45.33  & 31.00  & 24.50  \\
    \midrule
    \rowcolor{black!10}\textbf{SEQ (Lin)} & 19.66  & 9.26  & 14.88  & 13.43  & 12.64  \\
    \rowcolor{black!10}\textbf{SEQ (Cos)} & 16.89  & 5.97  & 11.10  & 11.40  & 10.08  \\
    \rowcolor{black!10}\textbf{SEQ (FixB+Cos)} & 17.13  & 6.08  & 10.32  & 7.45  & 9.30  \\
    \rowcolor{black!10}\textbf{SEQ (FixC+Cos)} & 50.96  & 64.28  & 44.93  & 33.48  & 28.90  \\
    \rowcolor{black!10}\textbf{SEQ (FixBC+Cos)} & 53.18  & 62.72  & 44.09  & 33.58  & 28.02  \\
    \rowcolor{black!10}\textbf{SEQ (W+FixBC+Lin)} & 33.41  & 19.06  & 17.79  & 13.68  & 13.65  \\
    \rowcolor{black!10}\textbf{SEQ (P+W+FixBC+Lin)} & 33.70  & 27.20  & 15.09  & 17.08  & 14.54  \\
    \midrule
    \rowcolor{black!20}\textbf{SEQ* (W+FixBC+Cos)} & 50.77  & \underline{75.96} & 53.76  & \underline{46.12} & \underline{36.55} \\
    \rowcolor{black!20}\textbf{SEQ* (P+W+FixBC+Cos)} & 70.56  & \textbf{84.51 } & \textbf{67.12 } & \textbf{61.99 } & \textbf{44.34 } \\
    \bottomrule
    \end{tabular}%
    }
  \label{tab:sota_main_gen_pythia410m}%
\end{table}%

\begin{table}[!t]
  \centering
  \caption{Comparison between SOTA methods and SEQ* on word-level classification tasks. The backbone is bert-base-cased. The IL scenario is CIL. No old samples are stored for all models. Other notation is the same as Table \ref{tab:sota_main_gen_pythia410m}. The full result is in Table \ref{tab:sota_full_bert-base-cased_word}.}
  \resizebox{\linewidth}{!}{
    \begin{tabular}{lccc}
    \toprule
          & \textbf{Few-NERD} & \textbf{OntoNotes5} & \textbf{I2B2} \\
\cmidrule{2-4}          & \boldmath{}\textbf{$\mathcal{A}_T$}\unboldmath{} & \boldmath{}\textbf{$\mathcal{A}_T$}\unboldmath{} & \boldmath{}\textbf{$\mathcal{A}_T$}\unboldmath{} \\
    \midrule
    \textbf{SpanKL} & 18.26  & 40.10  & 6.12  \\
    \textbf{OCILNER} & 18.44  & 39.99  & 27.27  \\
    \textbf{ExtendNER} & 20.02  & 48.08  & 20.02  \\
    \textbf{DLD} & 20.75  & 47.23  & 30.50  \\
    \textbf{SelfTrain} & 23.46  & 51.08  & 23.60  \\
    \textbf{RDP} & 27.08  & 50.45  & 40.38  \\
    \textbf{CPFD} & \textbf{34.65 } & 55.58  & 43.52  \\
    \textbf{ICE\_O} & \underline{28.98} & 51.81  & 49.12  \\
    \textbf{ICE\_PLO} & 19.94  & 46.52  & 47.76  \\
    \textbf{CFNER} & 27.70  & 58.07  & 35.42  \\
    \midrule
    \rowcolor{black!10}\textbf{SEQ  (Lin)} & 2.97  & 4.38  & 5.26  \\
    \rowcolor{black!10}\textbf{SEQ (W+FixBC+Cos)} & 7.26  & 29.12  & 45.95  \\
    \rowcolor{black!10}\textbf{SEQ (P+W+FixBC+Cos)} & 3.17  & 29.70  & 47.10  \\
    \midrule
    \rowcolor{black!20}\textbf{SEQ* (W+FixBC+Lin)} & 28.13  & \underline{66.99} & \underline{71.76} \\
    \rowcolor{black!20}\textbf{SEQ* (P+W+FixBC+Lin)} & 28.21  & \textbf{67.39 } & \textbf{72.51 } \\
    \bottomrule
    \end{tabular}%
    }
  \label{tab:sota_main_dis_bert-base-cased}%
\end{table}%

\subsection{Comparing SOTA methods with SEQ*} 
\label{sec:comparing_sota_methods_with_SEQ*}
In this subsection, we evaluate SEQ* under extensive settings.
Despite its simplicity, SEQ* has competitive or even superior performance in most settings.

We provide the result under the CIL scenario in Figure \ref{fig:radar_sota}, Table \ref{tab:sota_main_gen_pythia410m} and \ref{tab:sota_main_dis_bert-base-cased} in the main manuscript due to the space limitation.
We provide the introduction and training details of SOTA methods in Appendix \ref{sec:appendix_introduction_training_details_baselins}.
The full results on other backbones, datasets, and IL settings are summarized in Table \ref{tab:sota_full_pythia410m_sentence}, \ref{tab:sota_full_pythia160m_sentence}, \ref{tab:sota_full_bert-large-cased_sentence}, \ref{tab:sota_full_bert-base-cased_sentence}, \ref{tab:sota_full_bert-large-cased_word}, \ref{tab:sota_full_bert-base-cased_word}, \ref{tab:sota_full_gpt2}, \ref{tab:sota_full_til} in Appendix \ref{sec:appendix_full_results_of_SEQ*}.
All baselines and SEQ* use the same backbone PLM for IL.
In all settings except for sentence-level classification tasks with discriminant backbones, SEQ* and all baselines store no old samples.

From the results in Table \ref{tab:sota_main_gen_pythia410m} and \ref{tab:sota_main_dis_bert-base-cased}, we have the following findings:
(1) SEQ* shows comparable or better performance on all datasets.
Using proper classifiers, fixing old classifiers, and pre-allocating future classifiers improve SEQ significantly.
We highlight that we do not aim to show SEQ* achieves SOTA performance across all settings.
Instead, we aim to demonstrate that SEQ* serves as a comparable baseline in most IL settings and should be considered in further IL studies.

(2) SEQ* does not perform best when the PLM is required to absorb new knowledge, or there are overlaps between new and old tasks.
For example, Few-NERD contains fine-grained entities, such as ``Airport'' and ``Hotel'', which PLMs may not have seen during pre-training.
In Topic3Datasets, ``Sci/Tech'' and ``Computers \& Internet'' belong to two different tasks.
Intuitively, the PLM need to adjust the class boundary to avoid overlapping between classes.

Furthermore, we compare the linear probing performance between SEQ* and SOTA methods in Figure \ref{tab:compare_linear_probing_sota_gen} and \ref{tab:compare_linear_probing_sota_dis}.
The results show that the difference in the linear probing performance is small compared with the observed performance.
The improvement between ``BeforeIL'' and ``AfterIL'' mainly comes from the adaptation from pre-training to downstream tasks (Figure \ref{fig:probing_CIL_IC_GEN},\ref{fig:probing_CIL_RE_GEN},\ref{fig:probing_CIL_NER_Ontonotes5_DIS},\ref{fig:probing_CIL_NER_I2B2_DIS}).
It explains why freezing PLMs after warm-up is effective.
It also explains why prompt-based methods \cite{razdaibiedina2023progressive,wang2022learning} are effective even if only a tiny portion of the parameters are learned.
Furthermore, the performance gap between SEQ* and linear probing performance still exists.
The reason is that backward knowledge transfer from new tasks to old tasks is prohibited in SEQ*.

We compare the training time and the number of trainable parameters between SEQ* and SOTA methods in Table \ref{tab:comparison_time_params}.
SEQ* requires much less training time and trainable parameters for each task.

\begin{table}[htbp]
  \centering
  \caption{The linear probing performance with Pythia-410m. Other settings are the same as Table \ref{tab:sota_main_gen_pythia410m}.}
  \resizebox{\linewidth}{!}{
    \begin{tabular}{lcccc}
    \toprule
          & \multicolumn{2}{c}{\textbf{CLINC150}} & \multicolumn{2}{c}{\textbf{FewRel}} \\
\cmidrule{2-5}          & \textbf{Before IL} & \textbf{After IL} & \textbf{Before IL} & \textbf{After IL} \\
    \midrule
    \textbf{SEQ (Lin)} & \multirow{5}[2]{*}{91.05\small{±0.65}} & 91.08\small{±0.20} & \multirow{5}[2]{*}{52.18\small{±0.50}} & 77.39\small{±0.28} \\
    \textbf{L2KD} &       & 90.88\small{±0.78} &       & 76.57\small{±0.46} \\
    \textbf{LAMOL\_t} &       & 91.30\small{±1.14} &       & 81.54\small{±0.66} \\
    \textbf{LAMOL\_g} &       & 91.42\small{±0.25} &       & 81.09\small{±0.71} \\
    \textbf{SEQ* (P+W+FixBC+Cos)} &       & 91.12\small{±0.52} &       & 77.47\small{±0.84} \\
    \bottomrule
    \end{tabular}%
    }
  \label{tab:compare_linear_probing_sota_gen}%
\end{table}%

\begin{table}[htbp]
  \centering
  \caption{The comparison of training time and trainable parameters for each task on CLINC150. $\dagger$: The model after warm-up.}
  \resizebox{\linewidth}{!}{
        \begin{tabular}{ccc}
    \toprule
          & \textbf{Time (Min)} & \textbf{\# Trainable Params each Task} \\
    \midrule
    \textbf{PCLL} & 199   & 410M \\
    \textbf{L2KD} & 179   & 405M \\
    \textbf{LAMOL\_KD} & 119   & 405M \\
    \textbf{LAMOL\_t} & 70    & 405M \\
    \textbf{LAMOL\_g} & 68    & 405M \\
    \rowcolor{black!10}\textbf{SEQ* } & 24    & 10.24K$^{\dagger}$ \\
    \bottomrule
    \end{tabular}%
     }
  \label{tab:comparison_time_params}%
\end{table}%

\section{Related Work}
\label{sec:related_work}
Previous studies have assessed catastrophic forgetting by measuring performance degradation on old tasks.
However, there is limited understanding of probing performance in incremental learning. 
\citet{davari2022probing} use linear probing to reveal that representations still experience significant drift due to parameter updates. 
\citet{wu2021pretrained} conduct layer-wise probing studies on BERT, revealing catastrophic forgetting in the top and middle layers. 
They observed that, although BERT maintains high representational ability at the last incremental step, the classifier loses the ability to classify previously learned classes. 
\citet{chenforgetting} conduct linear probing on k-shot samples from the next task, revealing a strong correlation between retaining past information and learning efficiency on new tasks.
\citet{tao2023can} utilize linear probing to illustrate that BERT is inherently resilient to catastrophic forgetting, even without buffer data in task-incremental learning.  
In this study, we further investigate the influence of probing metrics, backbone and classifier architecture, pre-training steps, datasets, and IL methods on the probing performance.
Furthermore, we analyze the forgetting of classifiers from the perspective of norm and cosine similarity.
Finally, we propose simple but effective strategies for SEQ and conduct extensive experiments to validate its effectiveness.

\section{Conclusion}
Incremental learning is a key pillar of human cognition and intelligence.
As PLMs have become popular in recent years, more and more IL studies adopt PLMs as the backbone model.
However, we reveal that existing studies ignore the inherent anti-forgetting of PLMs and design methods based on a problematic assumption.
Our findings encourage the IL community to revisit the assumption of catastrophic forgetting in PLMs and re-evaluate the proposed IL algorithms by comparing them with frozen-based methods such as SEQ*.

We suggest two future directions for IL with PLMs: 
(1) design IL benchmark where domain-specific knowledge is required;
(2) design IL algorithms that update the knowledge of PLMs with limited time, computation cost and memory budget.

\section*{Limitations}

There are two limitations of this study:
(1) We only focused on the IL of classification tasks and did not explore the forgetting of general forms of knowledge in PLMs;
(2) We did not fully understand the internal mechanism of how PLMs incrementally learn the knowledge under SEQ.

\section*{Ethical Considerations}
The ethical considerations of our research are carefully addressed to ensure compliance with relevant standards and transparency. To this end, we provide the following clarifications for reproducibility:
\begin{itemize}
    \item We provide a detailed setting of our experiments.
    \item The source code, data, and scripts will all be publicly available. 
    \item Our findings are in alignment with observed empirical outcomes.
\end{itemize}

\section*{Acknowledgements}
We thank the anonymous reviewers for their helpful feedbacks.The work described in this paper was partially funded by the National Natural Science Foundation of China (Grant No. 62272173), the Natural Science Foundation of Guangdong Province (Grant Nos. 2024A1515010089, 2022A1515010179), the Science and Technology Planning Project of Guangdong Province (Grant No. 2023A0505050106), and the National Key R\&D Program of China (Grant No. 2023YFA1011601).

\bibliography{anthology,custom}

\appendix
\renewcommand*\contentsname{Appendix}
\clearpage
\addtocontents{toc}{\protect\setcounter{tocdepth}{2}}
\begin{center}
    \tableofcontents
\end{center}

\begin{table*}[!t]
  \centering
  \caption{The statistics on eight datasets for incremental learning. Granularity: the classification granularity. For example, named entity recognition models classify each word into an entity type or non-entity, while intent classification models classify sentences into intent categories. \# base classes: the number of classes to learn in the first task; \# Inc. classes: the number of classes to learn in the incremental task (i.e., the second and subsequent tasks).}
  \resizebox{\linewidth}{!}{
    \begin{tabular}{ccccccccc}
    \toprule
    \textbf{Granularity} & \textbf{Task} & \textbf{Dataset} & \textbf{\# Classes} & \textbf{\# Tasks} & \textbf{\# Base Classes} & \textbf{\# Inc. Classes} & \textbf{\# Training Instances} & \textbf{\# Test Instances} \\
    \midrule
    \multirow{5}[6]{*}{Sentence Level} & Text Classification & Topic3Datasets & 25    & 5     & 5     & 5     & 75000 & 46000 \\
\cmidrule{2-9}          & \multirow{2}[2]{*}{Intent Classification} & CLINC150 & 150   & 15    & 10    & 10    & 15000 & 4500 \\
          &       & Banking77 & 77    & 7     & 11    & 11    & 7191  & 2800 \\
\cmidrule{2-9}          & \multirow{2}[2]{*}{Relation Extraction} & FewRel & 80    & 8     & 10    & 10    & 33600 & 11200 \\
          &       & TACRED & 40    & 8     & 5     & 5     & 5909  & 1259 \\
    \midrule
    \multirow{3}[2]{*}{Word Level} & \multirow{3}[2]{*}{Named Entity Recognition} & Few-NERD & 66    & 11    & 6     & 6     & 131758 & 230025 \\
          &       & Ontonotes5 & 18    & 6     & 8     & 2     & 59922 & 23836 \\
          &       & I2B2  & 16    & 5     & 8     & 2     & 59376 & 41397 \\
    \bottomrule
    \end{tabular}%
    }
  \label{tab:datasets_statistics}%
\end{table*}%

\begin{table*}[!t]
  \centering
  \caption{The details of the 9 backbones. $\dagger$: Non-embedding parameters according to \citet{biderman2023pythia}.}
  \resizebox{\linewidth}{!}{
    \begin{tabular}{ccccccc}
    \toprule
    \textbf{Architecture} & \textbf{Model Class} & \textbf{Pretrained Weights} & \textbf{Parameters} & \textbf{Layers} & \textbf{Hidden Dim} & \textbf{Link} \\
    \midrule
    \multirow{2}[2]{*}{Encoder-Only} & \multirow{2}[2]{*}{BERT} & bert-base-cased & 109M  & 12    & 768   & \href{https://huggingface.co/bert-base-cased}{Link} \\
          &       & bert-large-cased & 335M  & 24    & 1024  & \href{https://huggingface.co/bert-large-cased}{Link} \\
    \midrule
    \multirow{7}[4]{*}{Decoder-Only} & \multirow{5}[2]{*}{GPT-NeoX} & Pythia-70m & 19M$^{\dagger}$   & 6     & 512   & \href{https://huggingface.co/EleutherAI/pythia-70m-deduped}{Link} \\
          &       & Pythia-160m & 85M$^{\dagger}$   & 12    & 768   & \href{https://huggingface.co/EleutherAI/pythia-160m-deduped}{Link} \\
          &       & Pythia-410m & 302M$^{\dagger}$  & 24    & 1024  & \href{https://huggingface.co/EleutherAI/pythia-410m-deduped}{Link} \\
          &       & Pythia-1b & 805M$^{\dagger}$  & 16    & 2048  & \href{https://huggingface.co/EleutherAI/pythia-1b-deduped}{Link} \\
          &       & Pythia-1.4b & 1.21B$^{\dagger}$ & 24    & 2048  & \href{https://huggingface.co/EleutherAI/pythia-1.4b-deduped}{Link} \\
\cmidrule{2-7}          & \multirow{2}[2]{*}{GPT2} & gpt2-base & 124M  & 12    & 768   & \href{https://huggingface.co/gpt2}{Link} \\
          &       & gpt2-large & 774M  & 36    & 1280  & \href{https://huggingface.co/gpt2-large}{Link} \\
    \bottomrule
    \end{tabular}%
    }
  \label{tab:statistics_backbones}%
\end{table*}%

\section{Problem Formulation}
\label{sec:appendix_problem_formulation}

\subsection{Overview of IL}
Incremental Learning (IL) aims to learn a model on new tasks incrementally without forgetting previous knowledge.
In this paper, we only consider classification tasks, which are popular and challenging settings in existing studies.
Formally, IL aims to learn a model $f_\theta:\mathbf{x}\rightarrow y \in \mathcal{Y}$ from the sequence of tasks $\mathcal{D}=\{\mathcal{D}_1,\mathcal{D}_2,\cdots,\mathcal{D}_T\}$, where the $t$-th task $\mathcal{D}_t = \{ (\mathbf{x}_i^t,y_i^t)\}_{i=1}$ contains input samples $\mathbf{x}_i^t \in \mathcal{X}_t$ and labels $y_i^t \in \mathcal{Y}_t$.
There are three popular scenarios in IL: Class-Incremental Learning (CIL), Task-Incremental Learning (TIL), and Domain-Incremental Learning (DIL).
In CIL, the label sets of different tasks are exclusive: $\mathcal{Y}_1 \cap \mathcal{Y}_2 \cdots \mathcal{Y}_T = \emptyset$, and the task identity is unknown during inference.
In TIL, the label sets of different tasks may be overlapping: $\mathcal{Y}_1 \cap \mathcal{Y}_2 \cdots \mathcal{Y}_T \neq \emptyset$, and the task identity is required during inference.
In DIL, the label sets of different tasks are the same: $\mathcal{Y}_1 = \mathcal{Y}_2 = \mathcal{Y}_T$.
Under the data replay setting, a buffer $\mathcal{M}$ is introduced for storing old representative instances.
In the main experiments of this research, we consider the most challenging scenario, CIL, where catastrophic forgetting occurs most severely.
The result on TIL is also reported when compared with state-of-the-art methods.

\subsection{Evaluation Metric for IL}
We adopt average accuracy \cite{chaudhry2018riemannian} as the metric for evaluation.
Specifically, the average accuracy at task $t$ is defined as the following
\begin{equation}
    \mathcal{A}_t = \frac{1}{t}\sum_{i=1}^{t}a_{t,i},
\end{equation}
where $a_{t, i}$ represents the accuracy evaluated on the test set of task $i$ after training the model incrementally from tasks $1$ to $t$.
The average accuracy indicates the performance on all learnt tasks.
In the main manuscript, we report the average accuracy after learning the final task, i.e., $\mathcal{A}_T$.
Besides, we report the \textit{average incremental accuracy} $\bar{\mathcal{A}}$ in the appendix.
The average incremental accuracy is computed as follows:
\begin{equation}
    \bar{\mathcal{A}} = \frac{1}{T}\sum_{t=1}^{T}\mathcal{A}_t
\end{equation}
It indicates the average of the average accuracy over all incremental steps.

We note that the probing accuracy is calculated as the average of the test accuracy on \textit{all} $T$ tasks:
\begin{equation}
    \mathcal{A}_{prob} = \frac{1}{T}\sum_{i=1}^{T}a_{prob,i},
\end{equation}
$a_{prob, i}$ represents the accuracy evaluated on the test set of task $i$ after training probing classifiers.
The probing accuracy is the performance when the classifier is optimal.
According to the probing experiments in Sec. \ref{sec:revisiting_forgetting_from_probing}, the PLMs almost do not forget.
Therefore, the probing performance can be regarded as the upper bound performance when using PLMs for IL.

\section{Datasets}
\label{sec:appendix_datasets}

The statistics of the eight datasets are summarized in Table \ref{tab:datasets_statistics}.
For text classification, we construct Topic3Datasets from AGNews, DBPedia, and YaHoo \cite{zhang2015character}.
We remove \textit{Sports}, \textit{Business \& Finance},\textit{Science \& Mathematics} from YaHoo since they overlap with the classes in AGNews.
We subsample 3000 training samples and 2000 test samples for each class (only 1900 test samples for the four classes in AGNews).
The class order is obtained by sorting class names alphabetically and shuffling them using the random seed 1.

We use the default class order for Clinc150, Banking77, FewRel, and TACRED. 
We follow \cite{shao2023class} to convert the class name to semantic labels for generative backbones.
For example, in TARCRED, \textit{org:founded\_by} is converted to \textit{organization related: founded by}.

For Few-NERD, Ontontes5, and I2B2, the class order is obtained by sorting class names alphabetically.
We use the BIO schema for tagging.
The \textit{\# class} represents the number of entities in Table \ref{tab:datasets_statistics}.

For sentence-level classification tasks, we report the accuracy.
For word-level classification tasks, we report the macro-f1 due to the class imbalance.

\section{Backbones}
\label{sec:appendix_backbones}
The statistics of the 9 backbones are summarised in Table \ref{tab:statistics_backbones}.
We download the pre-trained weights from Huggingface \citep{wolf2019huggingface}. 

\subsection{Discriminant Backbones}
For discriminant backbones (i.e., encoder-only backbones), we use the [CLS] feature for sentence-level classification tasks and the feature of last hidden states for word-level classification tasks.
We do not use prompts for discriminant backbones.
When learning each new task, we add a linear layer on top of the backbone for classification.
For example, in the bert-base case, we add a linear layer whose input dimension and output dimension are 768 and 10 for learning the first task in CLINC150.
Then, we add another linear layer with the same architecture as the previous one when learning the second task.
In the CIL scenario, the output logits over all learned categories are obtained by concatenating the logits from all classifiers.
In the TIL scenario, the output logits are the logits from the classifier with the same task ID as the input sample.

\subsection{Generative Backbones}
For generative backbones (i.e., decoder-only backbones), we train the model to output the class name with causal language modelling loss.
We do not use generative backbones for word-level classification tasks because it requires special design on input and output format \cite{zhao-etal-2022-prompt} and the evaluation is different from that of sequential labelling models \cite{monaikul2021continual}.
For text and intent classification, we use the following prompt:
``Input sentence: \{text\}\textbackslash n The label: \{label\}\{eos token\}''.
For relation extraction, we use the following prompt:
``Input sentence: \{text\}\textbackslash n The relationship between \{head entity\} and \{tail entity\} is \{label\}\{eos token\}''.
We note that we use the same prompt for all baseline models unless they have special designs on prompts.
Following \citet{sun2019lamol}, only the causal language modelling loss of ``\{label\}\{eos token\}'' is optimized.
For the probing study, we use the last hidden states of the last word as the feature.

The max sequence length is 256 in Topic3Datasets, 50 in CLINC150, 64 in Banking77, 100 in FewRel, and 128 in other datasets.
We use exactly the same backbone for all baselines unless they have special designs on backbones.

\begin{figure*}[!t]
    \centering
    \hspace{-0.4cm}
    \subfloat[Lin. Prob]{
        \includegraphics[width=0.25\linewidth]{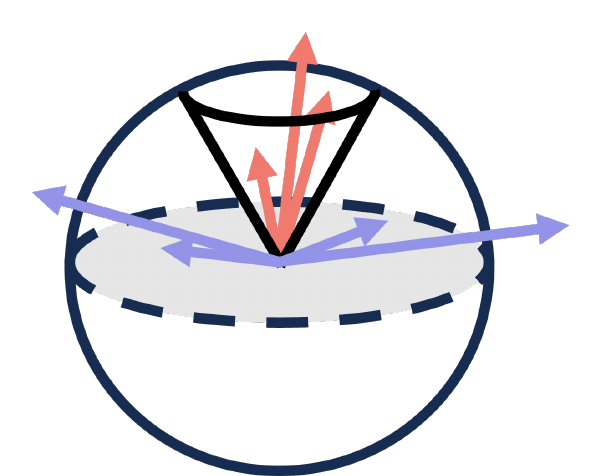}
        \label{fig:metirc_illustration_a}
    }
    \hspace{-0.3cm}
    \subfloat[Cos.Lin. Prob]{
        \includegraphics[width=0.2\linewidth]{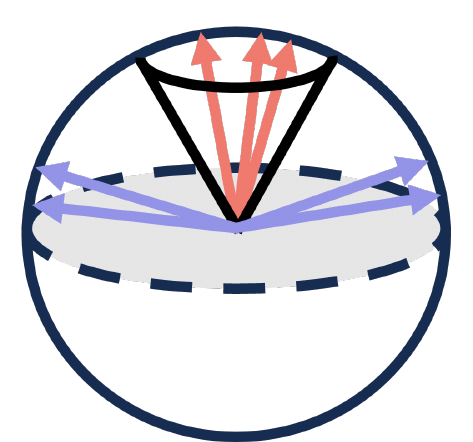}
        \label{fig:metirc_illustration_b}
    }
    \subfloat[Proto. Prob]{
        \includegraphics[width=0.185\linewidth]{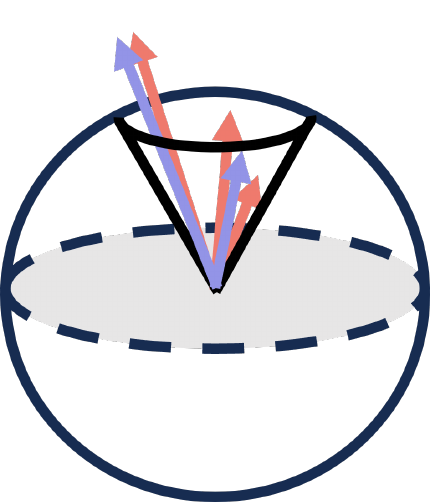}
        \label{fig:metirc_illustration_c}
    }
    \hspace{0.4cm}
    \subfloat[Cos.Proto Prob]{
        \includegraphics[width=0.185\linewidth]{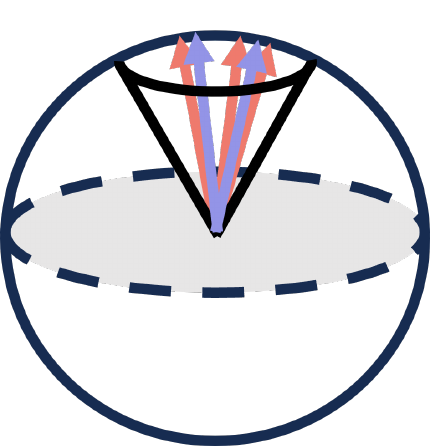}
        \label{fig:metirc_illustration_d}
    }

    \includegraphics[width=0.4\linewidth]{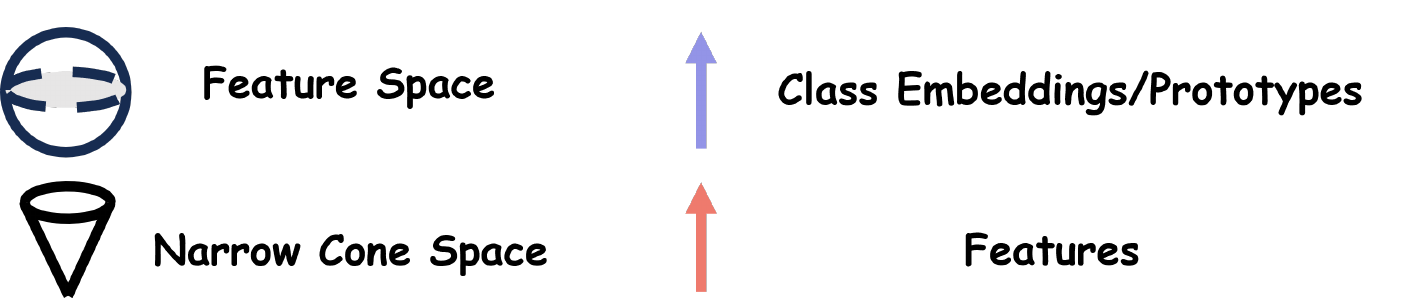}
    
    \caption{The illustration of different metrics for probing study.}
    \label{fig:metrics_illustration}
\end{figure*}

\begin{figure*}[!t]
    \centering
    \subfloat[Observed Performance]{
        \includegraphics[width=0.19\linewidth]{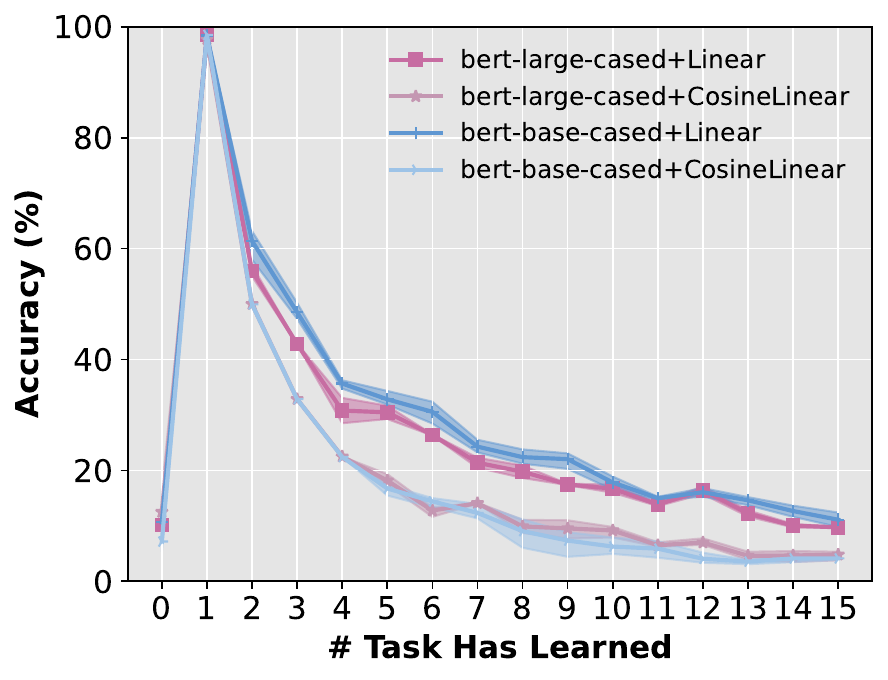}
    }
    \subfloat[Lin. Prob]{
        \includegraphics[width=0.19\linewidth]{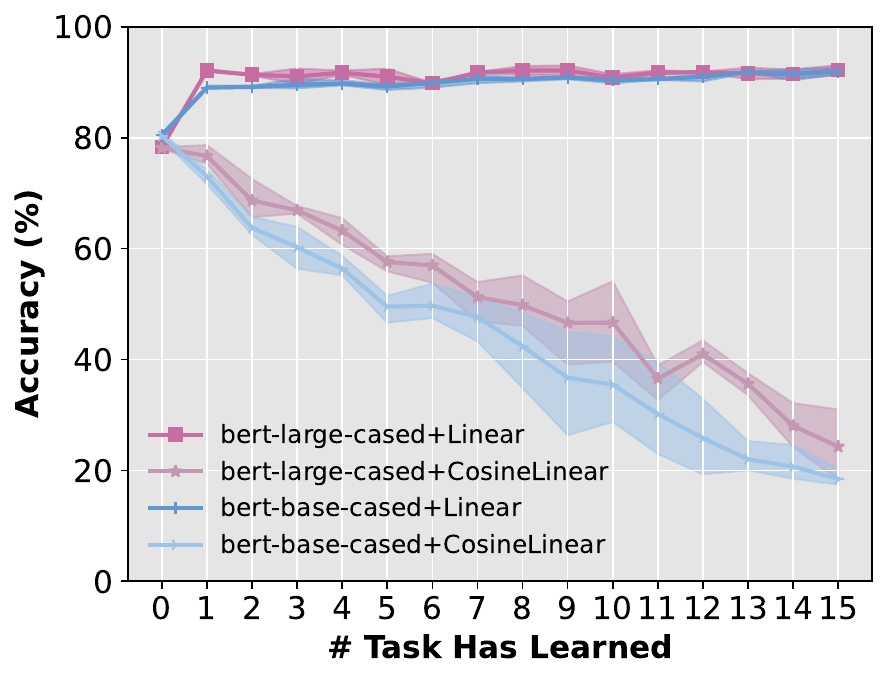}
    }
    \subfloat[Cos.Lin. Prob]{
        \includegraphics[width=0.19\linewidth]{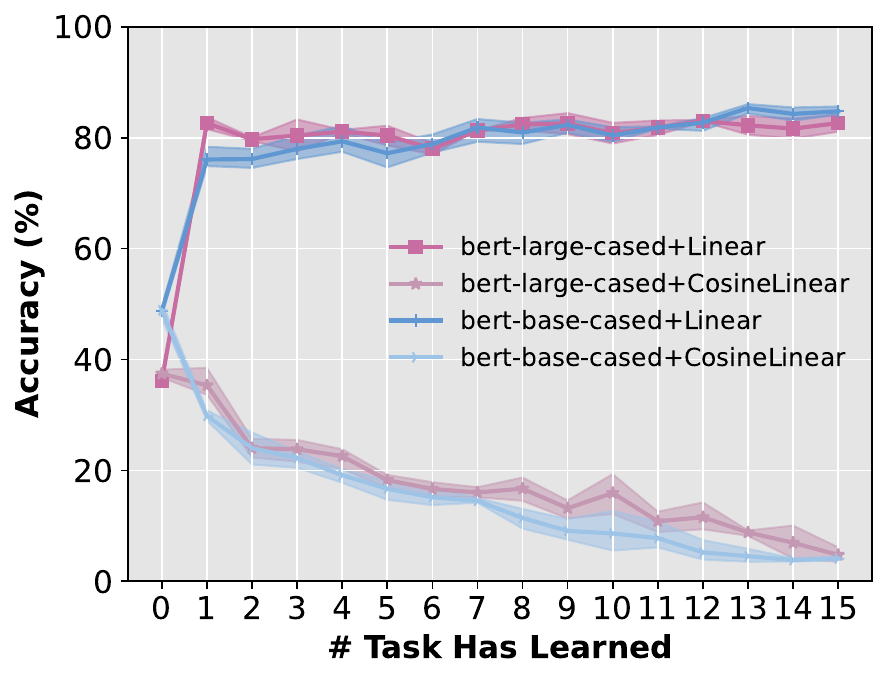}
    }
    \subfloat[Proto. Prob]{
        \includegraphics[width=0.19\linewidth]{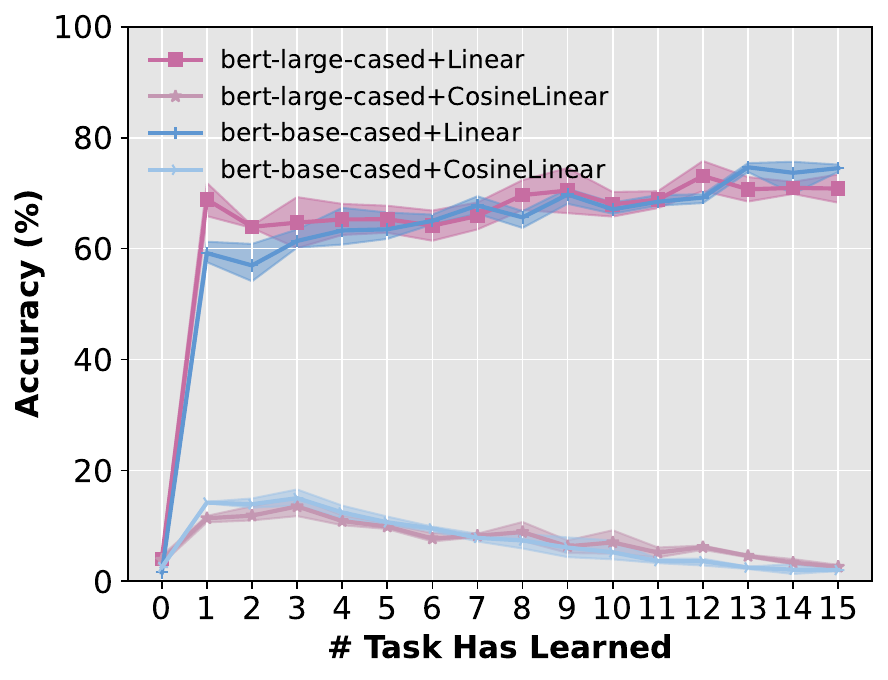}
    }
    \subfloat[Cos.Proto. Prob]{
        \includegraphics[width=0.19\linewidth]{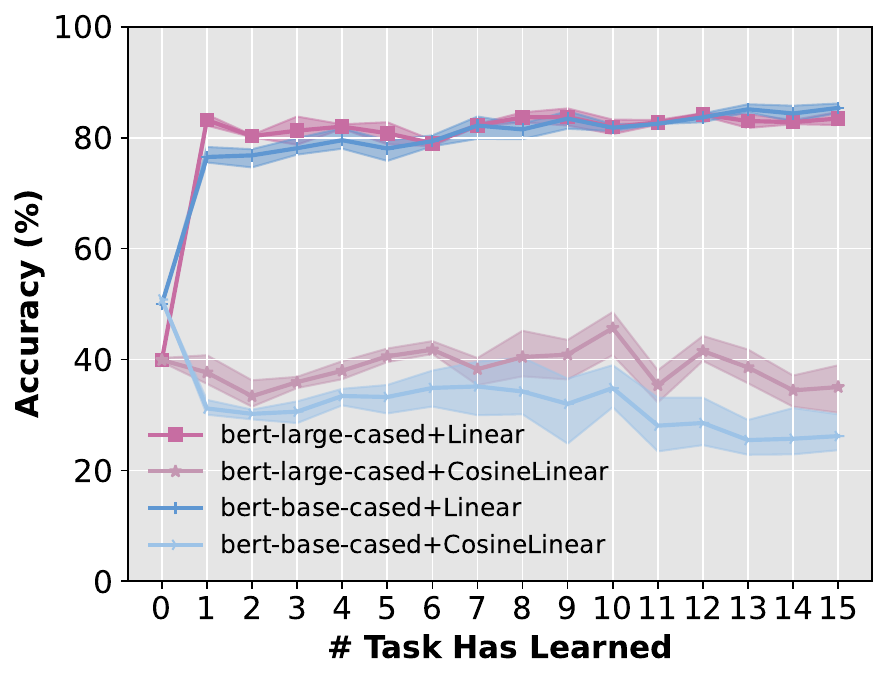}
    }
    \caption{The observed and probing performance on Class-Incremental Intent Classification. The dataset is CLINC150. The backbones are generative models. (a) shows the observed performance during IL training. (b)(c)(d)(e) show the probing performance when different metrics, including linear probing, cosine linear probing, prototype probing, and cosine prototype probing.}
    \label{fig:probing_CIL_IC_DIS}
\end{figure*}

\begin{figure*}[!t]
    \centering
    \subfloat[Observed Performance]{
        \includegraphics[width=0.19\linewidth]{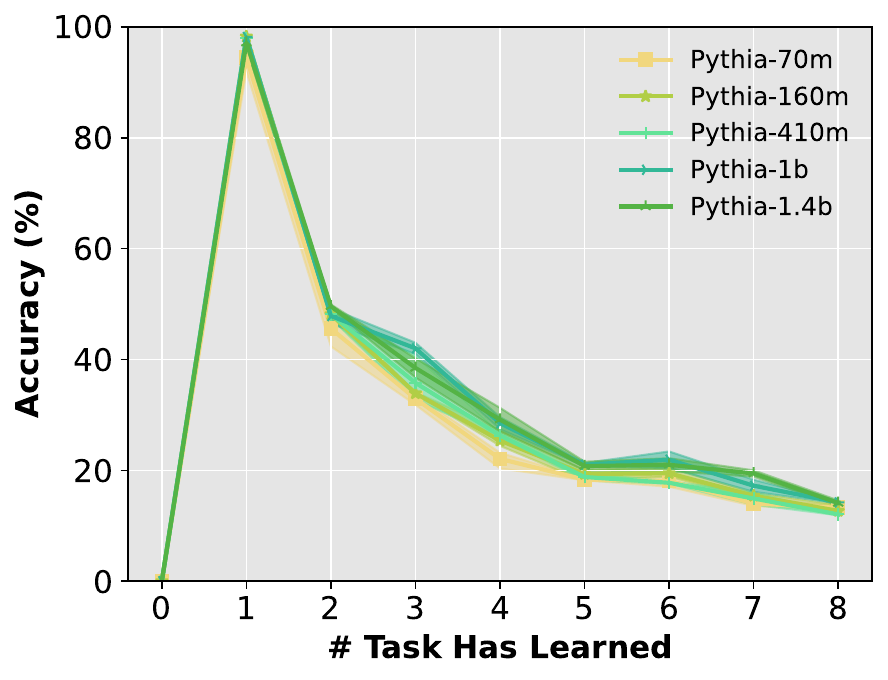}
    }
    \subfloat[Lin. Prob]{
        \includegraphics[width=0.19\linewidth]{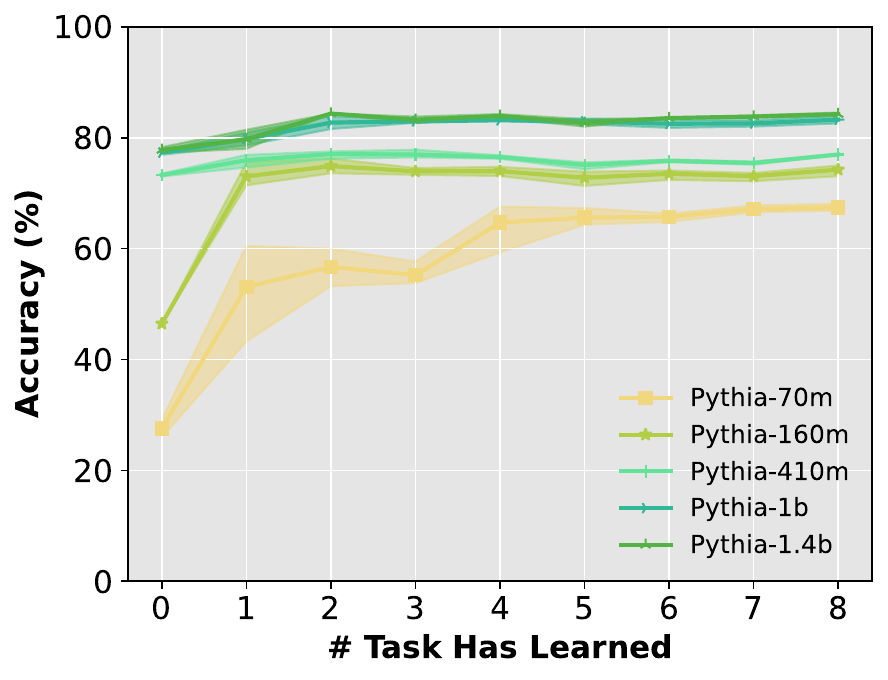}
    }
    \subfloat[Cos.Lin. Prob]{
        \includegraphics[width=0.19\linewidth]{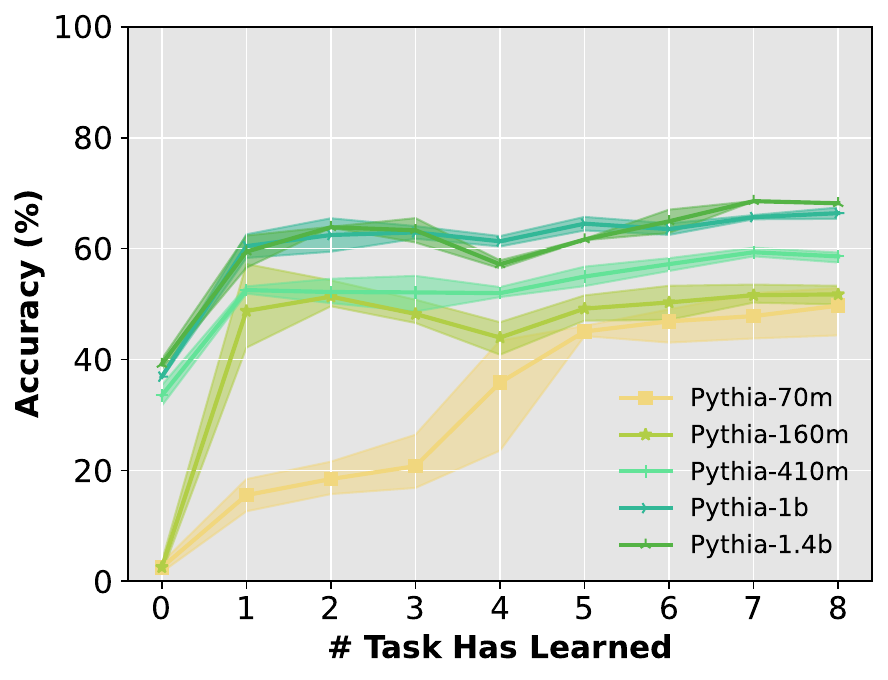}
    }
    \subfloat[Proto. Prob]{
        \includegraphics[width=0.19\linewidth]{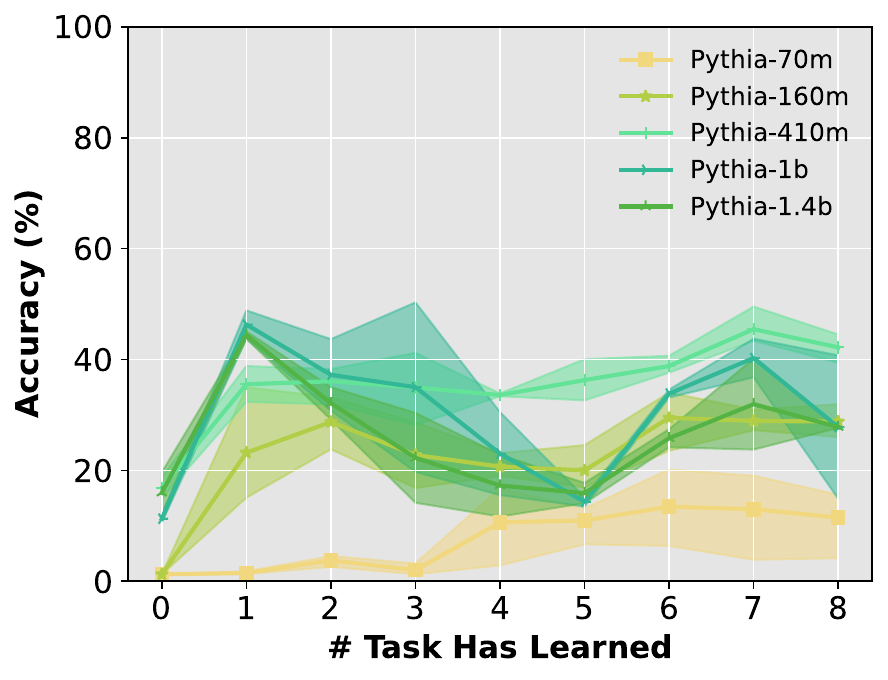}
    }
    \subfloat[Cos.Proto. Prob]{
        \includegraphics[width=0.19\linewidth]{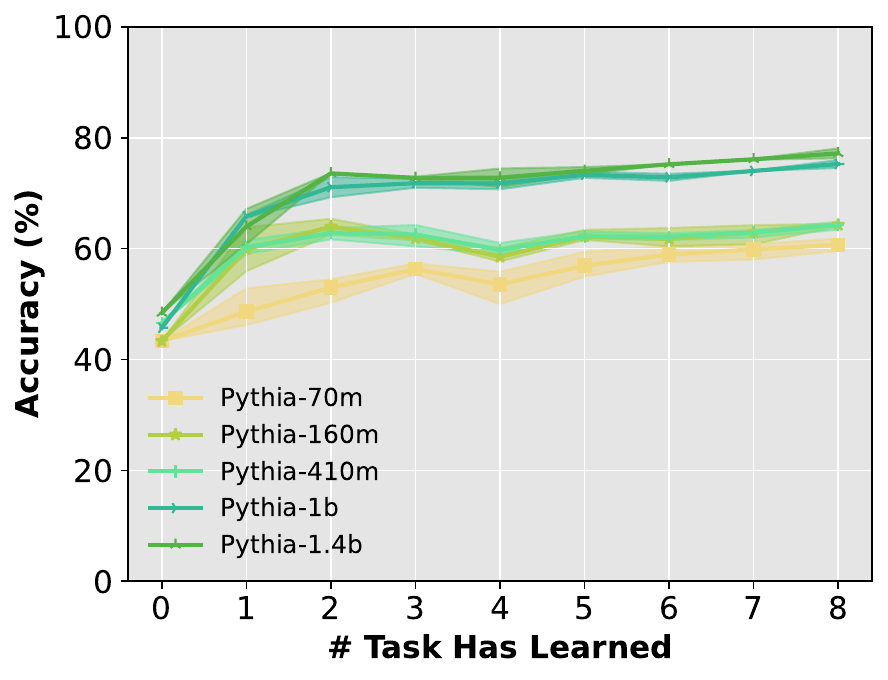}
    }
    \caption{The observed and probing performance on Class-Incremental Relation Extraction. The dataset is FewRel. The backbones are generative models. Other settings are the same as Figure \ref{fig:probing_CIL_IC_DIS}.}
    \label{fig:probing_CIL_RE_GEN}
\end{figure*}

\begin{figure*}[!t]
    \centering
    \subfloat[Observed Performance]{
        \includegraphics[width=0.19\linewidth]{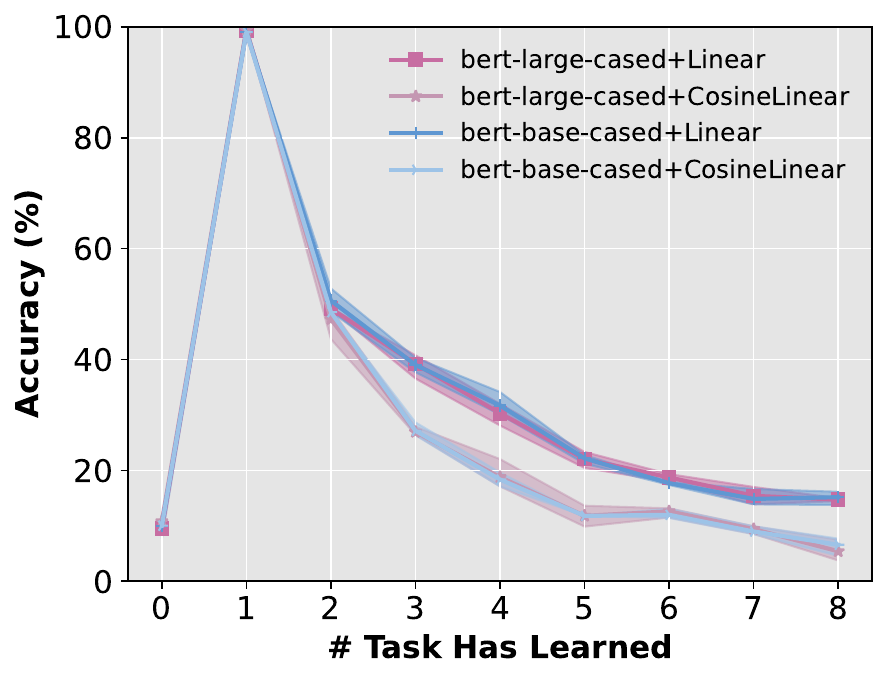}
    }
    \subfloat[Lin. Prob]{
        \includegraphics[width=0.19\linewidth]{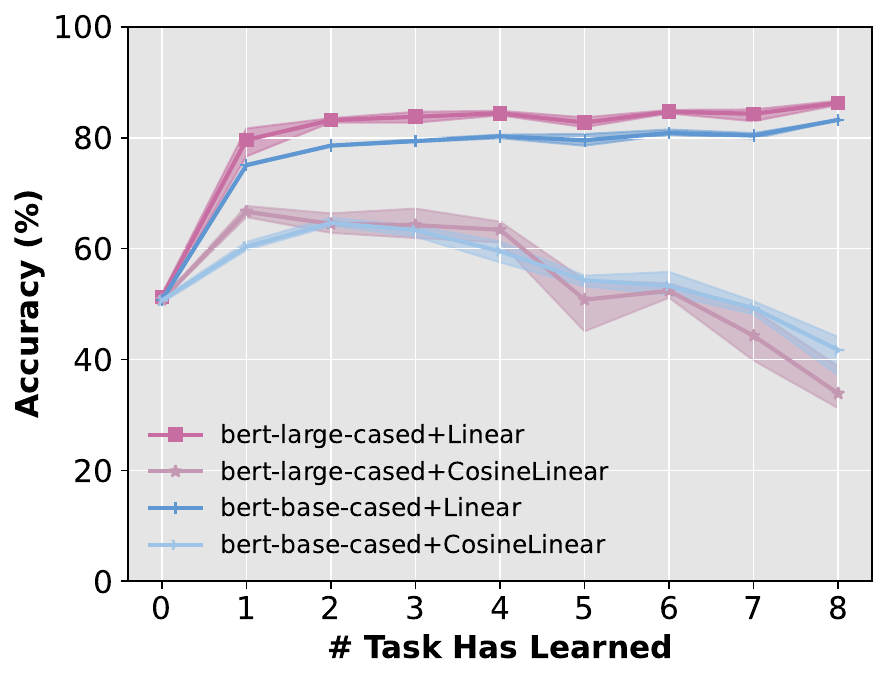}
    }
    \subfloat[Cos.Lin. Prob]{
        \includegraphics[width=0.19\linewidth]{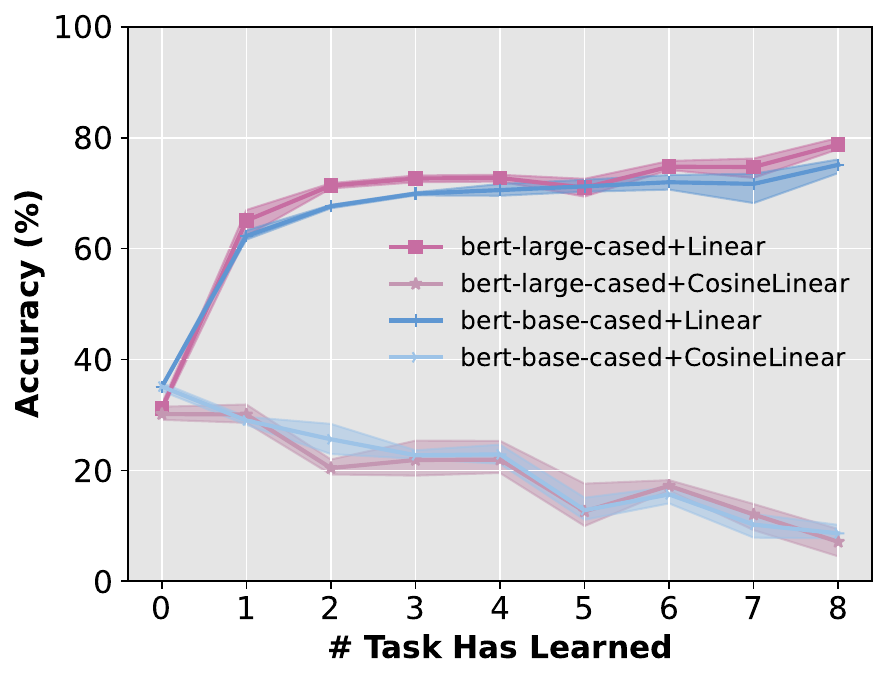}
    }
    \subfloat[Proto. Prob]{
        \includegraphics[width=0.19\linewidth]{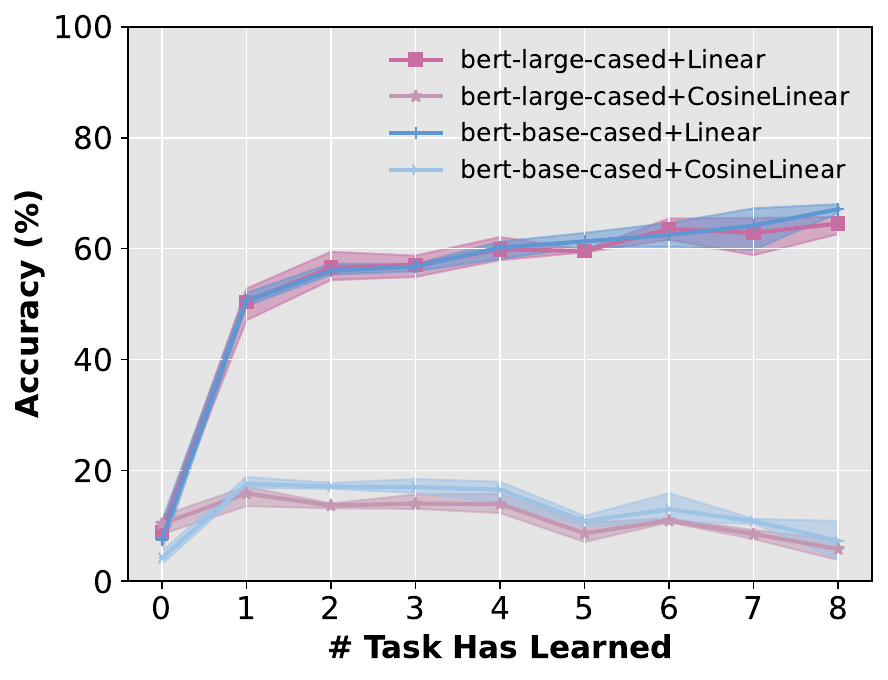}
    }
    \subfloat[Cos.Proto. Prob]{
        \includegraphics[width=0.19\linewidth]{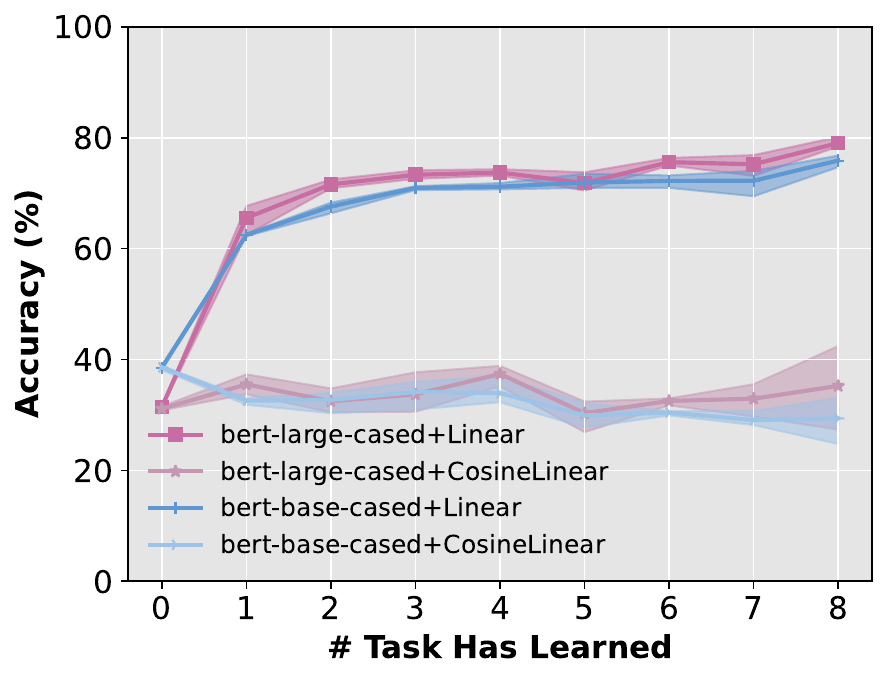}
    }
    \caption{The observed and probing performance on class-incremental relation extraction. The dataset is FewRel. The backbones are discriminant models. Other settings are the same as Figure \ref{fig:probing_CIL_IC_DIS}.}
    \label{fig:probing_CIL_RE_DIS}
\end{figure*}

\begin{figure*}[!t]
    \centering
    \subfloat[Observed Performance]{
        \includegraphics[width=0.19\linewidth]{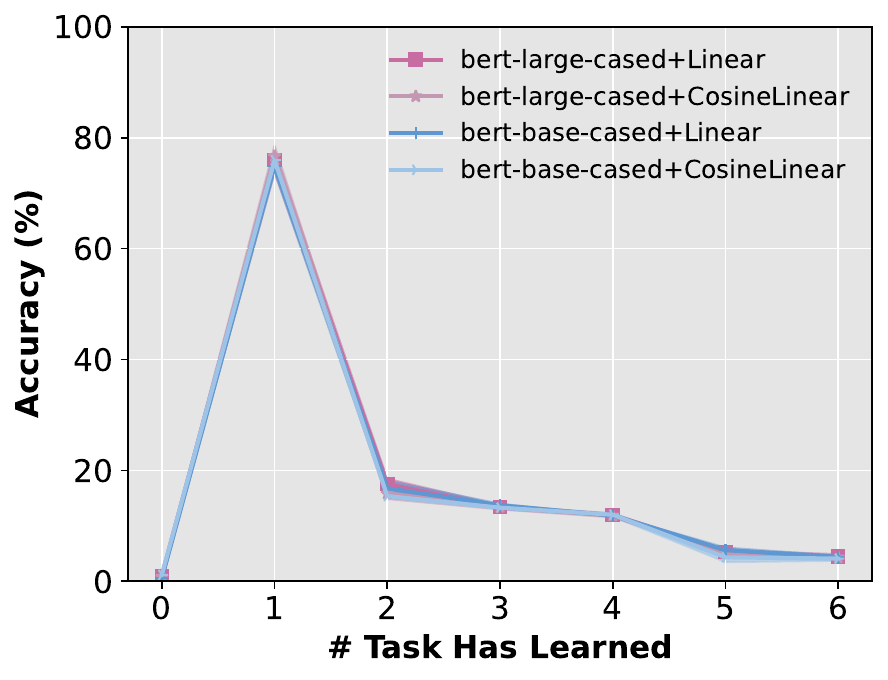}
    }
    \subfloat[Lin. Prob]{
        \includegraphics[width=0.19\linewidth]{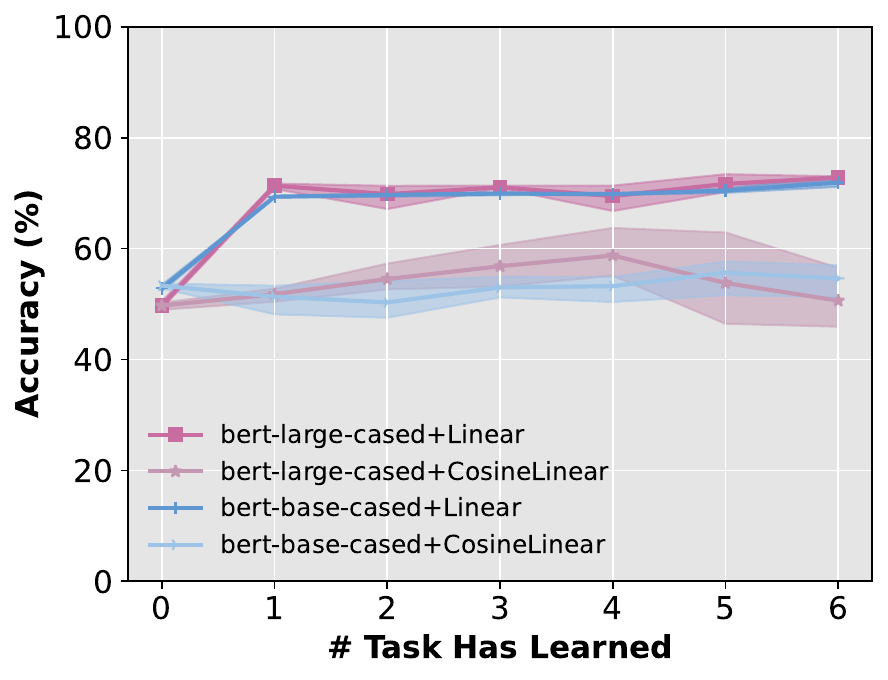}
    }
    \subfloat[Cos.Lin. Prob]{
        \includegraphics[width=0.19\linewidth]{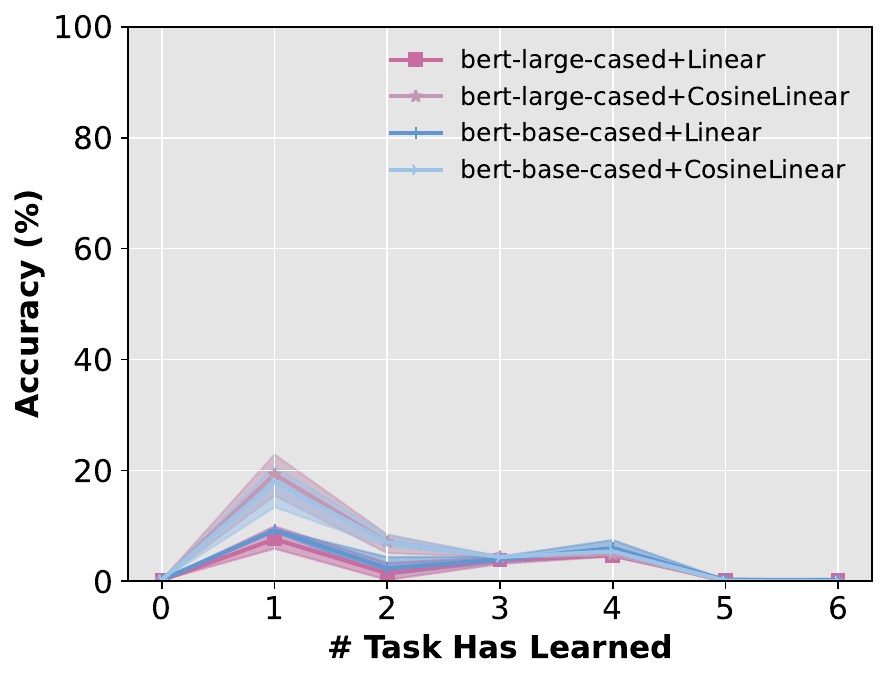}
    }
    \subfloat[Proto. Prob]{
        \includegraphics[width=0.19\linewidth]{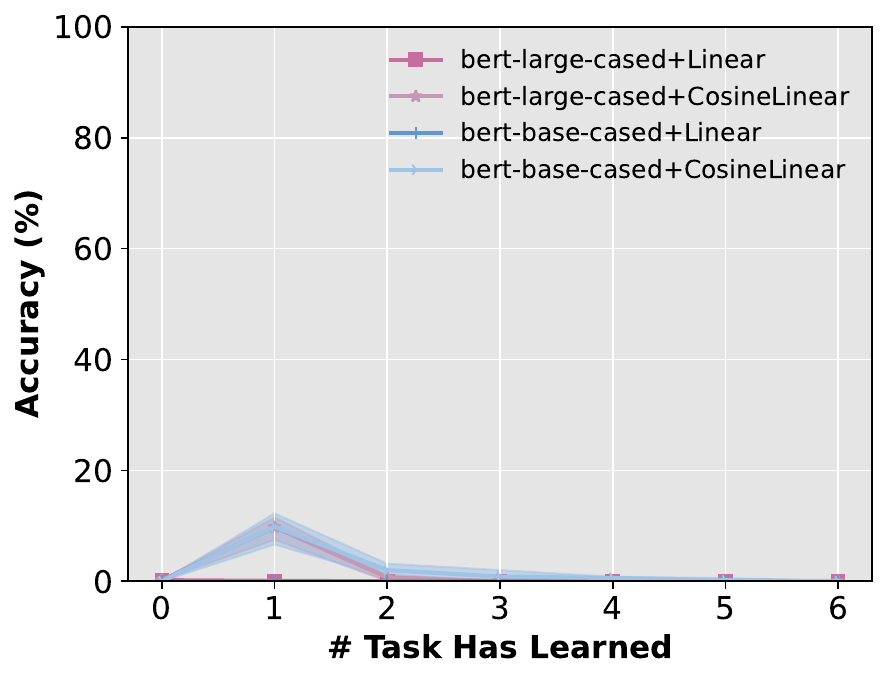}
    }
    \subfloat[Cos.Proto. Prob]{
        \includegraphics[width=0.19\linewidth]{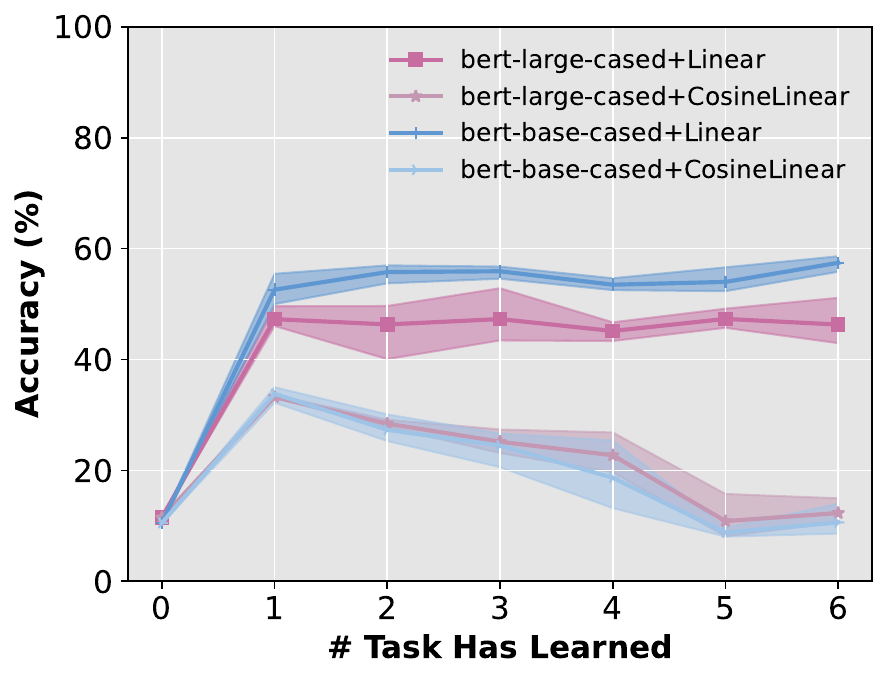}
    }
    \caption{The observed and probing performance on class-incremental named entity recognition. The dataset is Ontonotes5. The backbones are discriminant models. Other settings are the same as Figure \ref{fig:probing_CIL_IC_DIS}.}
    \label{fig:probing_CIL_NER_Ontonotes5_DIS}
\end{figure*}

\begin{figure*}[!t]
    \centering
    \subfloat[Observed Performance]{
        \includegraphics[width=0.19\linewidth]{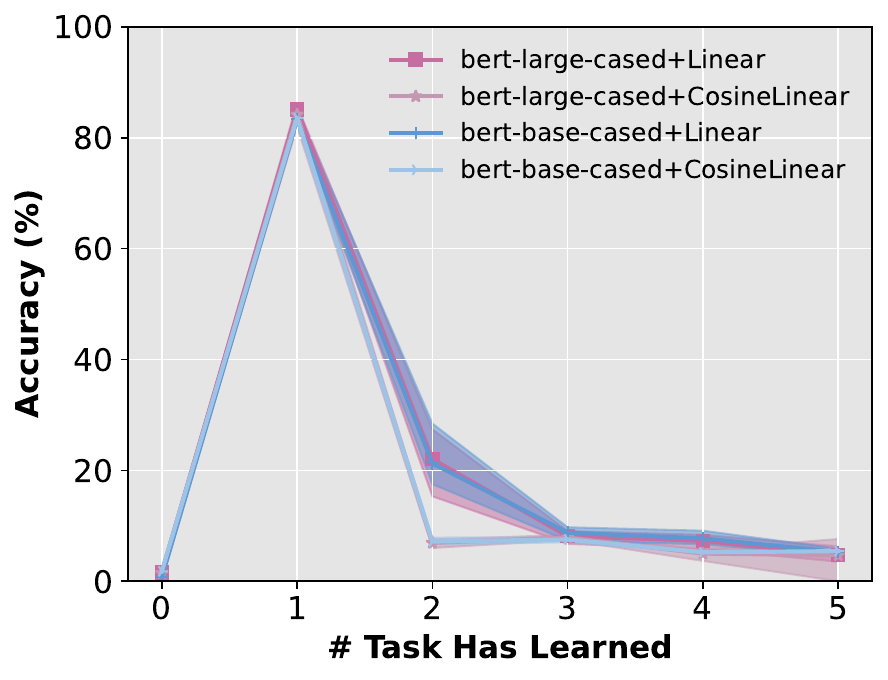}
    }
    \subfloat[Lin. Prob]{
        \includegraphics[width=0.19\linewidth]{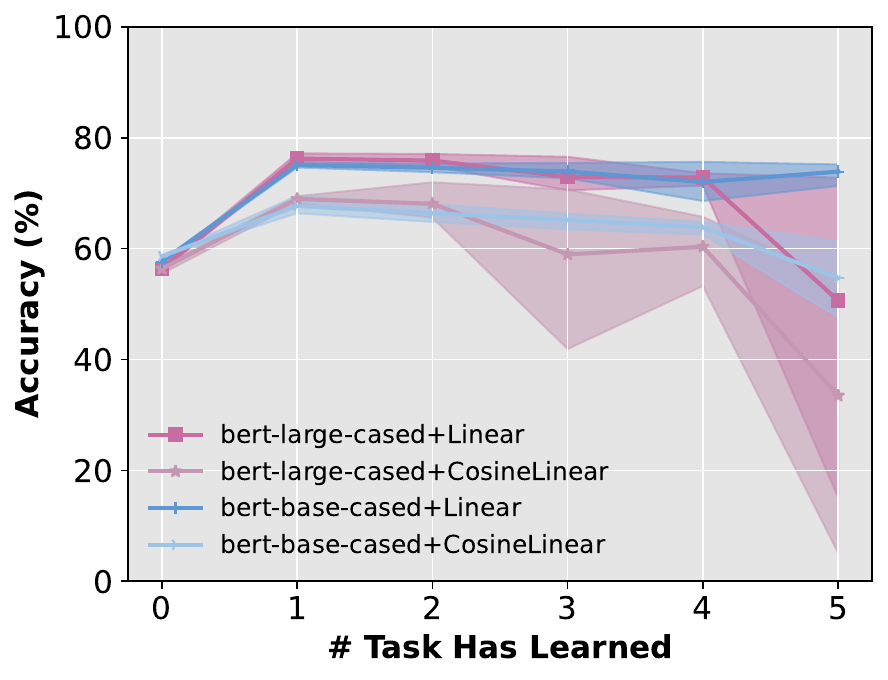}
    }
    \subfloat[Cos.Lin. Prob]{
        \includegraphics[width=0.19\linewidth]{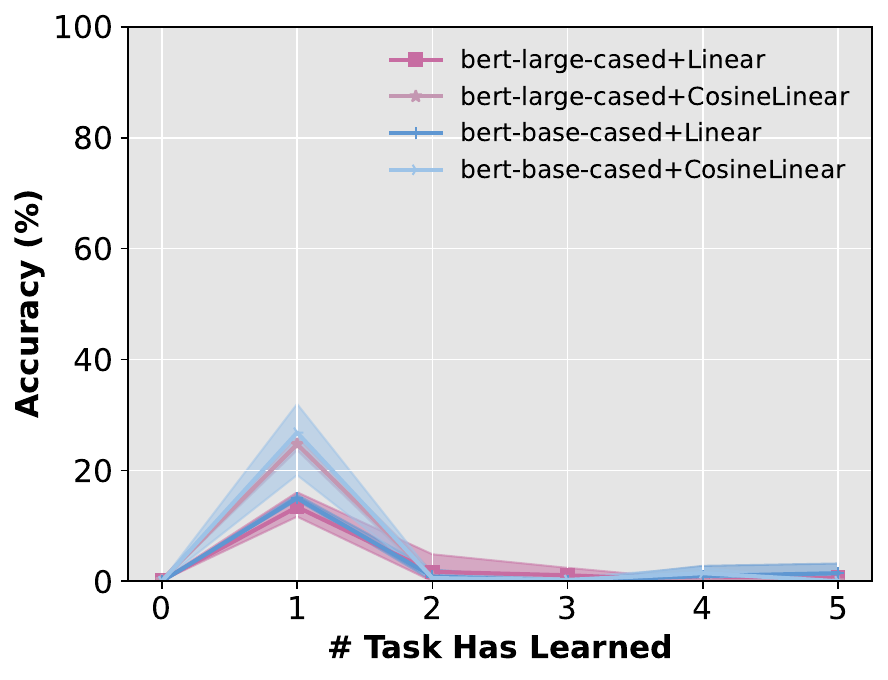}
    }
    \subfloat[Proto. Prob]{
        \includegraphics[width=0.19\linewidth]{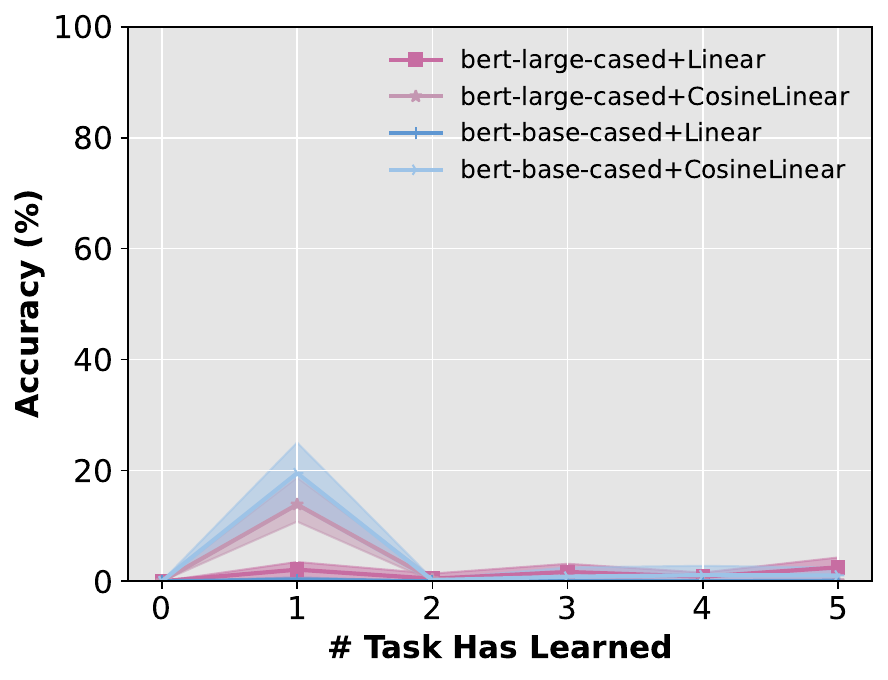}
    }
    \subfloat[Cos.Proto. Prob]{
        \includegraphics[width=0.19\linewidth]{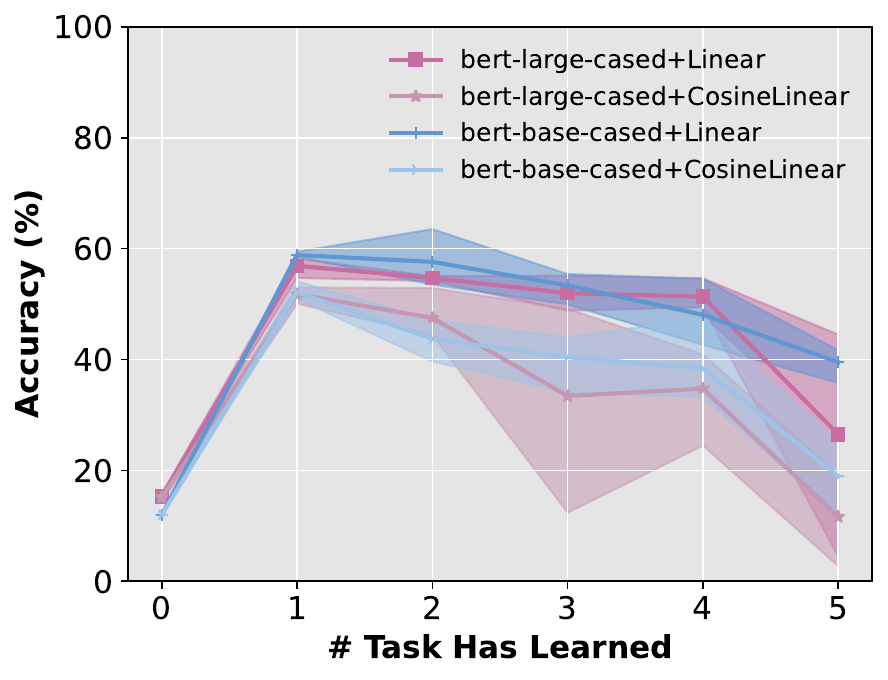}
    }
    \caption{The observed and probing performance on class-incremental named entity recognition. The dataset is I2B2. The backbones are discriminant models. Other settings are the same as Figure \ref{fig:probing_CIL_IC_DIS}.}
    \label{fig:probing_CIL_NER_I2B2_DIS}
\end{figure*}

\begin{figure}[!t]
    \centering
    \subfloat[Observed Performance]{
        \includegraphics[width=0.49\linewidth]{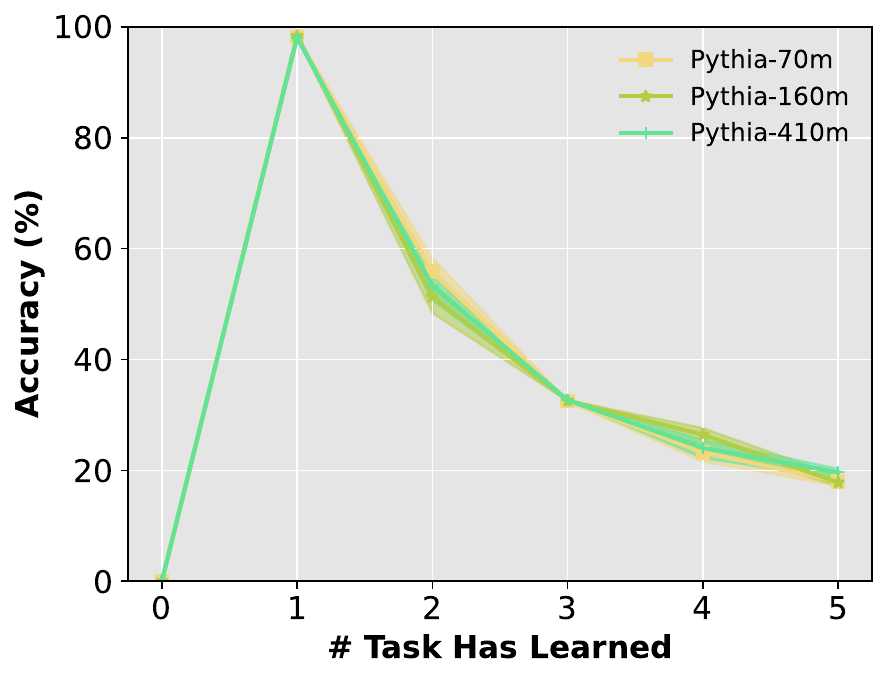}
    }
    \subfloat[Lin. Prob]{
        \includegraphics[width=0.49\linewidth]{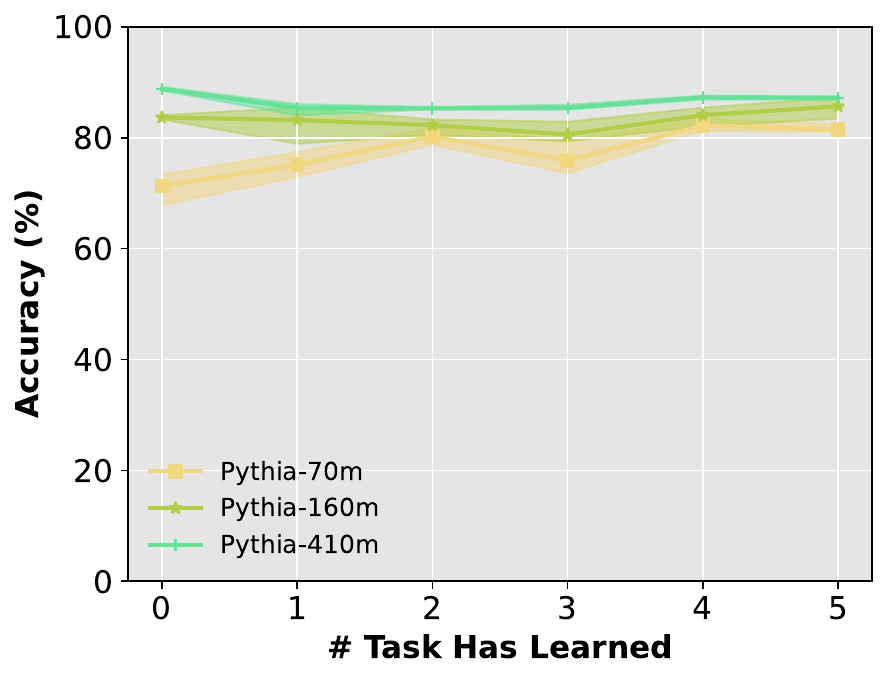}
    }

    \caption{The observed and probing performance on class-incremental text classification. The dataset is Topic3datasets. The backbones are generative models. Other settings are the same as Figure \ref{fig:probing_CIL_IC_DIS}.}
    \label{fig:probing_CIL_TC_GEN}
\end{figure}

\begin{figure}[!t]
    \centering
    \subfloat[Observed Performance]{
        \includegraphics[width=0.49\linewidth]{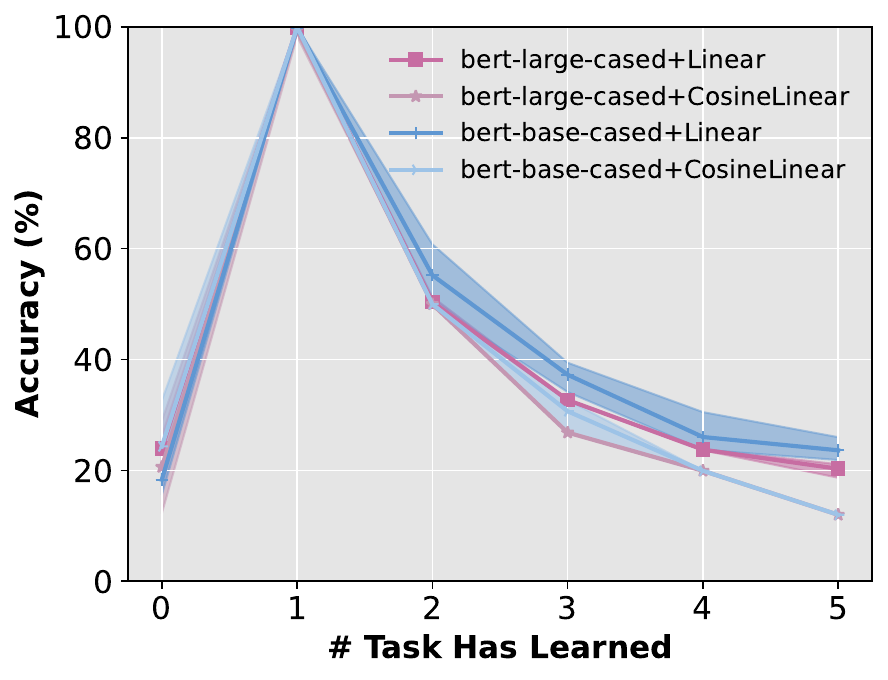}
    }
    \subfloat[Lin. Prob]{
        \includegraphics[width=0.49\linewidth]{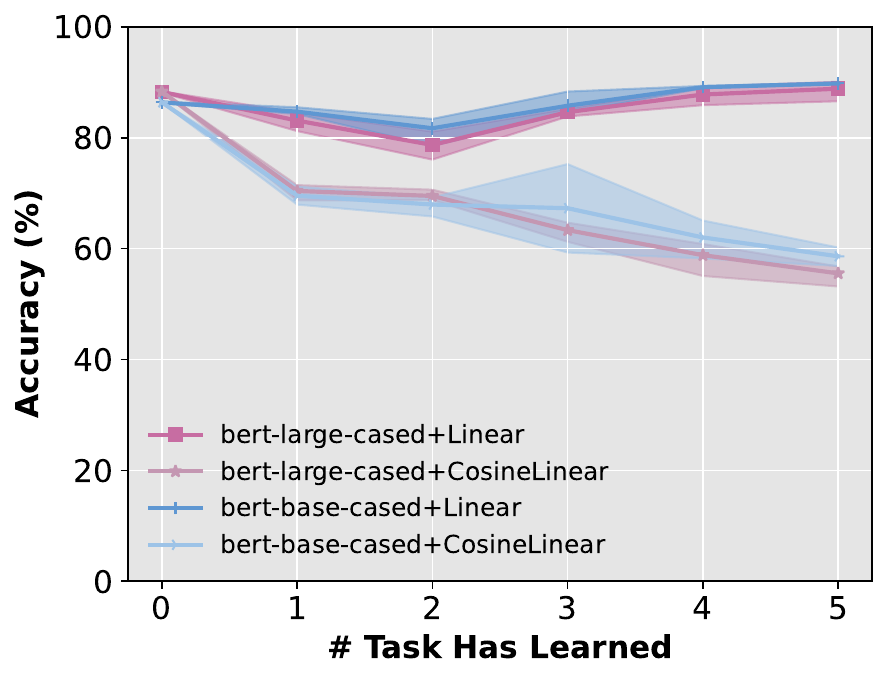}
    }

    \caption{The observed and probing performance on class-incremental text classification. The dataset is Topic3datasets. The backbones are discriminant models. Other settings are the same as Figure \ref{fig:probing_CIL_IC_DIS}.}
    \label{fig:probing_CIL_TC_DIS}
\end{figure}

\section{Revisiting IL with Probing Study}
\label{sec:appendix_probing_study_four_metrics}

\subsection{Four Probing Metrics}
\label{sec:appendix_probing_study_four_metrics_introduction}
\textbf{Linear Probing} trains a new linear layer on top of the backbone model. 
We do not use bias for linear probing classifiers because there is no significant difference between the probing performance.
We train the linear probing classifier for 20 epochs with an initial learning rate of 0.001.
The linear probing classifier is trained on the training data from all $T$ tasks jointly using cross-entropy loss.
The Adam \cite{kingma2014adam} optimizer is used, and the batch size is set as 128.
We note that the training data from all tasks is mixed for optimization.
Otherwise, the probing performance is degraded significantly.

\textbf{Cosine Linear Probing} is the same as linear probing except that the cosine similarity is adopted for calculating logits.
We use the same training process as the linear probing classifier to train the cosine linear probing classifier.
\citet{hou2019learning} propose to use cosine linear layers for IL to avoid prediction bias towards new classes.
However, we find that using cosine linear layers does not improve probing performance.

\textbf{Prototype Probing} requires no training of probing classifiers.
It calculates the class feature centre for each class using all training data.
It makes predictions of a test sample according to its Euclidean distances to all class centres.
The prototype probing classifier can be regarded as a linear classifier with a weight matrix specified as all class centres.

\textbf{Cosine Prototype Probing} is the same as prototype probing except that the cosine similarity is adopted for calculating logits.
The idea of using prototypes is widely adopted in IL \cite{ma-etal-2023-learning,cui-etal-2021-refining,han-etal-2020-continual}.
However, we reveal that using prototypes for classification may not be the best option.

\subsection{Probing Performance with Different Backbones}
\label{sec:appendix_probing_study_four_metrics_result}
We provide the probing performance of SEQ on class-incremental intent classification in Figure \ref{fig:probing_CIL_IC_GEN} and \ref{fig:probing_CIL_IC_DIS}, relation extraction in Figure \ref{fig:probing_CIL_RE_GEN} and \ref{fig:probing_CIL_RE_DIS}, text classification in Figure \ref{fig:probing_CIL_TC_GEN} and \ref{fig:probing_CIL_TC_DIS}, and named entity recognition in Figure \ref{fig:probing_CIL_NER_I2B2_DIS} and \ref{fig:probing_CIL_NER_Ontonotes5_DIS}.
We summarize the findings as follows:
\begin{itemize}
    \item The linear probing performance is significantly higher than the other three metrics across backbones, tasks, and datasets.
    \item For all probing metrics, the probing performance always increases when learning the first task.
    \item For generative backbones, the probing performance has not decreased or even increased since the second task. 
    It indicates that the smaller backbone can adapt to downstream tasks by SEQ. 
   The larger backbone can adapt to downstream tasks without training and maintain the knowledge during SEQ.
    \item For discriminant backbones, using the linear classifier maintains the probing performance, while the cosine linear classifier degrades the probing performance significantly.
    There is no significant difference between bert-base-cased and bert-large-cased in the probing performance.
\end{itemize}

\begin{figure*}[!t]
    \centering
    \subfloat[Feature Sim]{
        \includegraphics[width=0.24\linewidth]{fig/feature_embed_hist/pythia410m-feature-dist.pdf}
    }
    \subfloat[Feat-WordEmbed Sim]{
        \includegraphics[width=0.24\linewidth]{fig/feature_embed_hist/pythia410m-feature-wordembed-dist.pdf}
    }
    \subfloat[Feat-ClassEmbed Sim]{
        \includegraphics[width=0.24\linewidth]{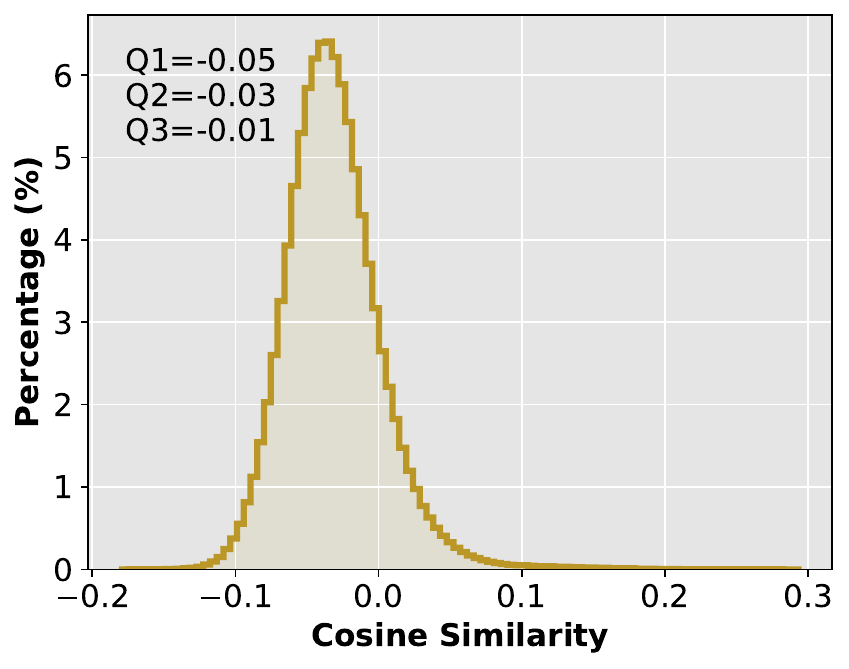}
    }
    \subfloat[Feat-Prototype Sim]{
        \includegraphics[width=0.24\linewidth]{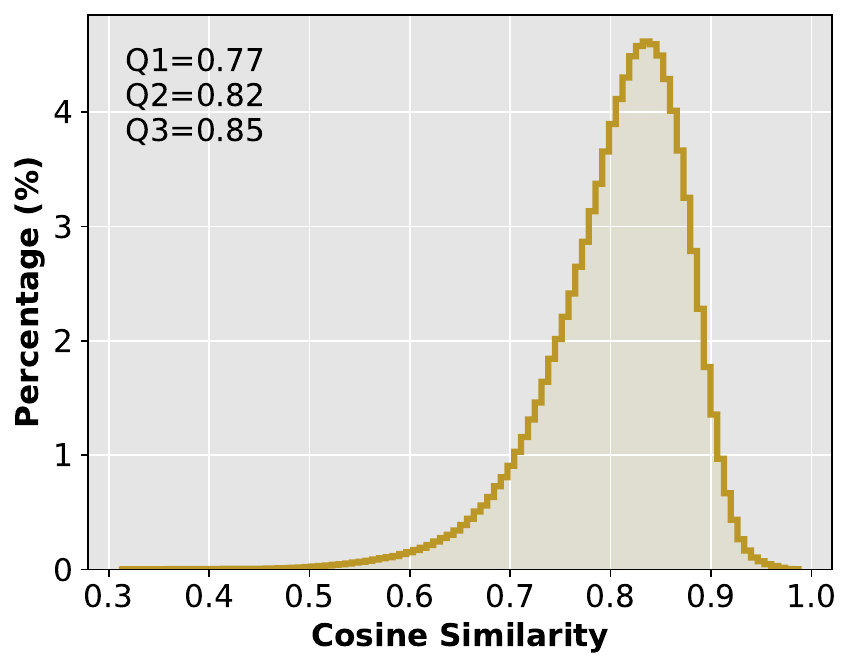}
    }

    \subfloat[Feature Norm]{
        \includegraphics[width=0.24\linewidth]{fig/feature_embed_hist/pythia410m-feature-norm.pdf}
    }
    \subfloat[WordEmbed Norm]{
        \includegraphics[width=0.24\linewidth]{fig/feature_embed_hist/pythia410m-wordembed-norm.pdf}
    }
    \subfloat[ClassEmbed Norm]{
        \includegraphics[width=0.24\linewidth]{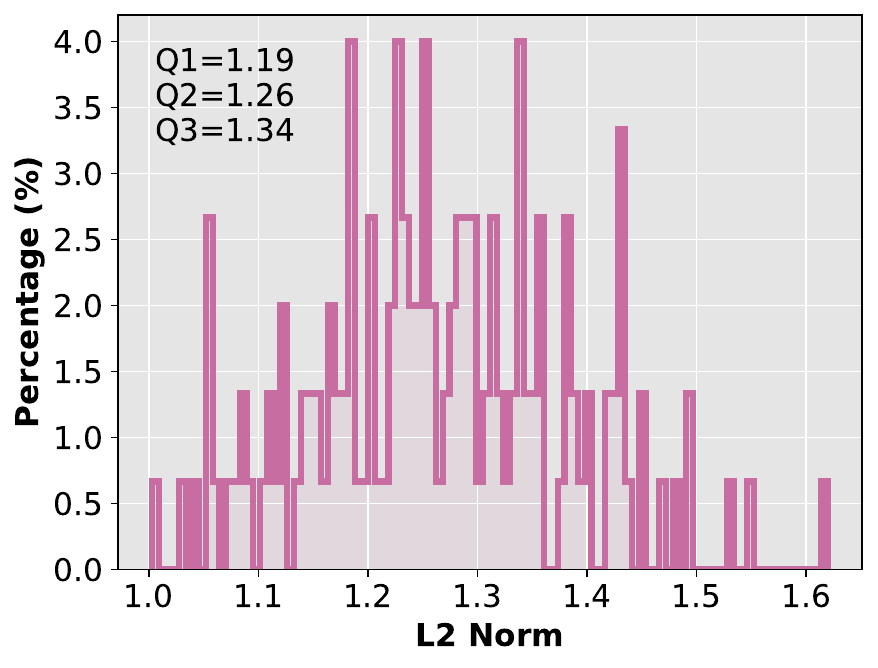}
    }
    \subfloat[Prototype Norm]{
        \includegraphics[width=0.24\linewidth]{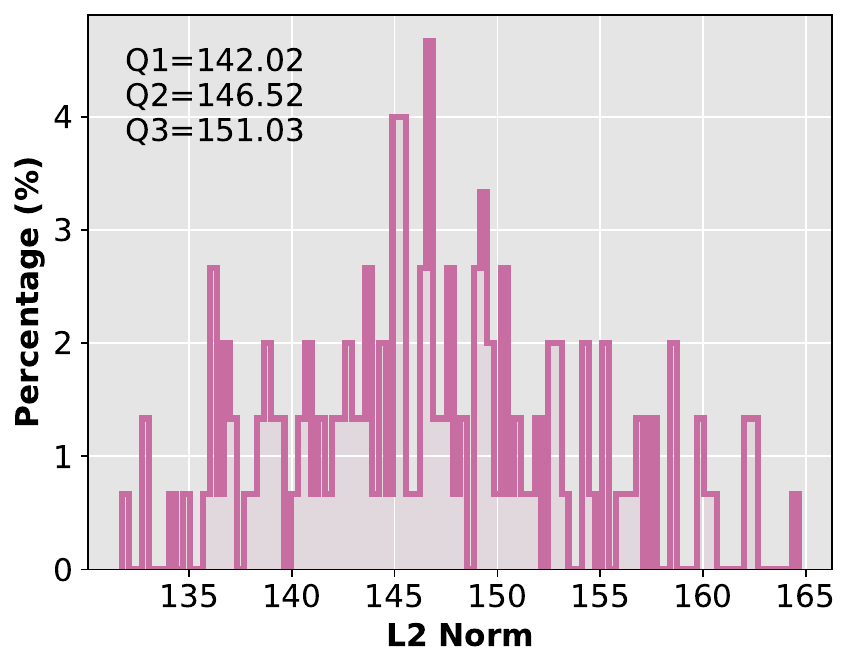}
    }
    
    \caption{The histogram of features and different embeddings of Pythia-410m. The features are calculated on the training set of CLINC150, and the output word embeddings are loaded from pre-trained weights. The class embeddings refer to the weight in the probing classifier on CLINC150. The class prototypes refer to the class feature centres on the training set of CLINC150.}
    \label{fig:metrics_illustration_full_pythia410m}
\end{figure*}

\begin{figure*}[!t]
    \centering
    \subfloat[Feature Sim]{
        \includegraphics[width=0.24\linewidth]{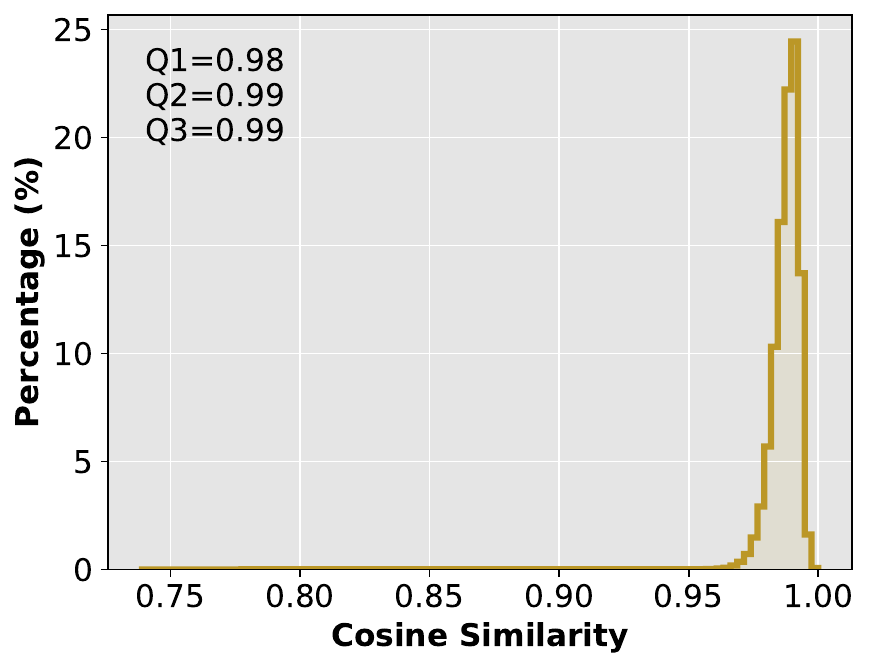}
    }
    \subfloat[Feat-WordEmbed Sim]{
        \includegraphics[width=0.24\linewidth]{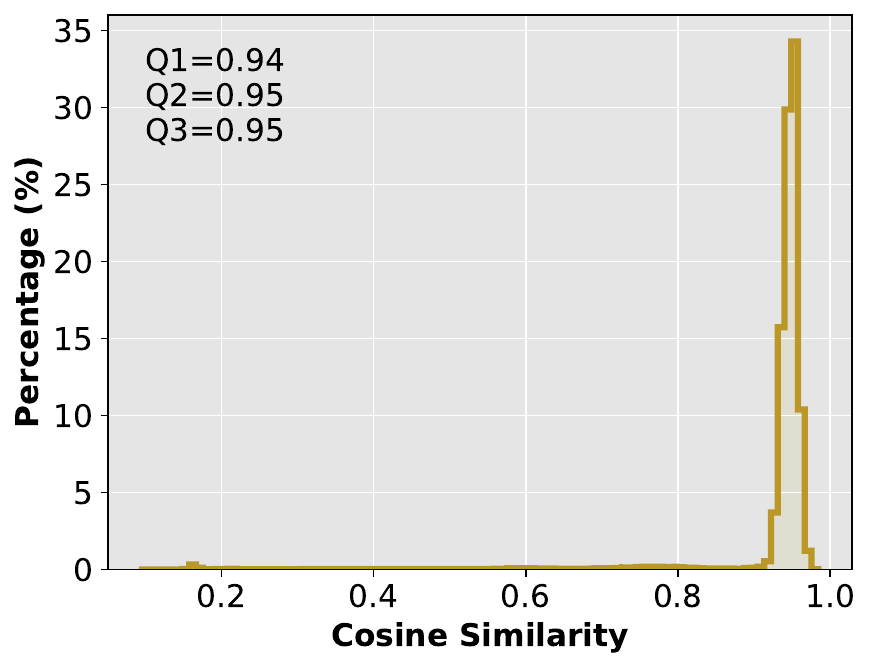}
    }
    \subfloat[Feat-ClassEmbed Sim]{
        \includegraphics[width=0.24\linewidth]{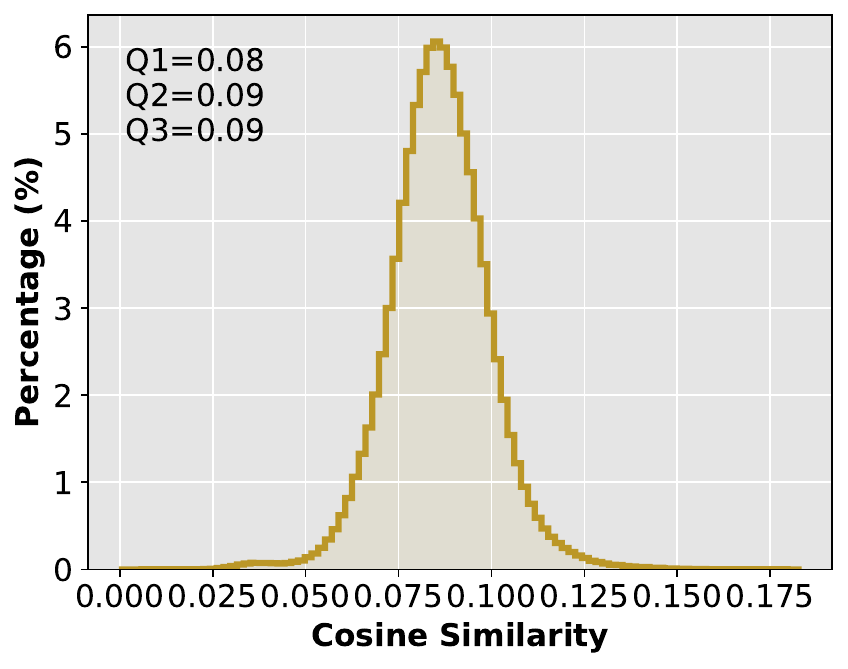}
    }
    \subfloat[Feat-Prototype Sim]{
        \includegraphics[width=0.24\linewidth]{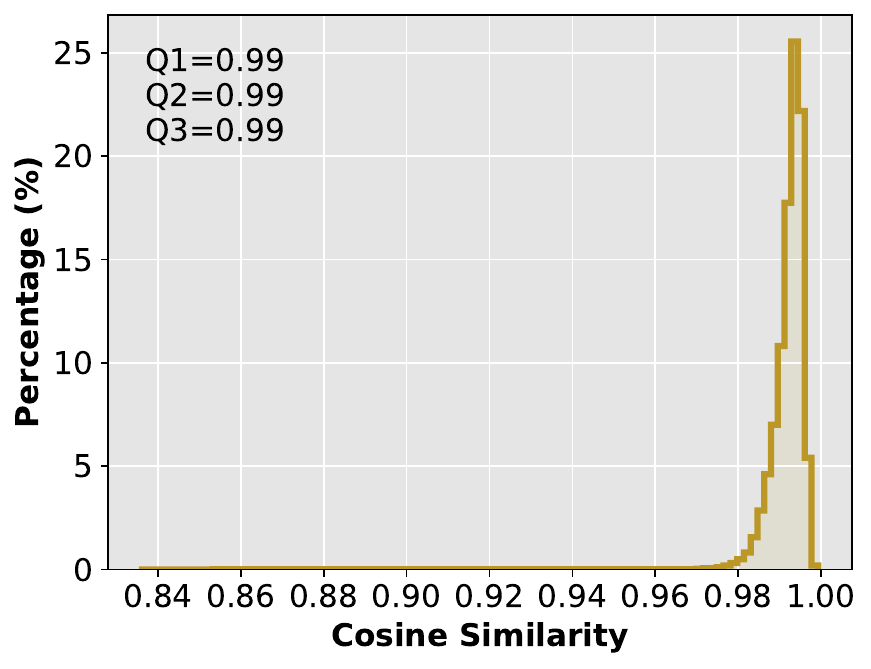}
    }

    \subfloat[Feature Norm]{
        \includegraphics[width=0.24\linewidth]{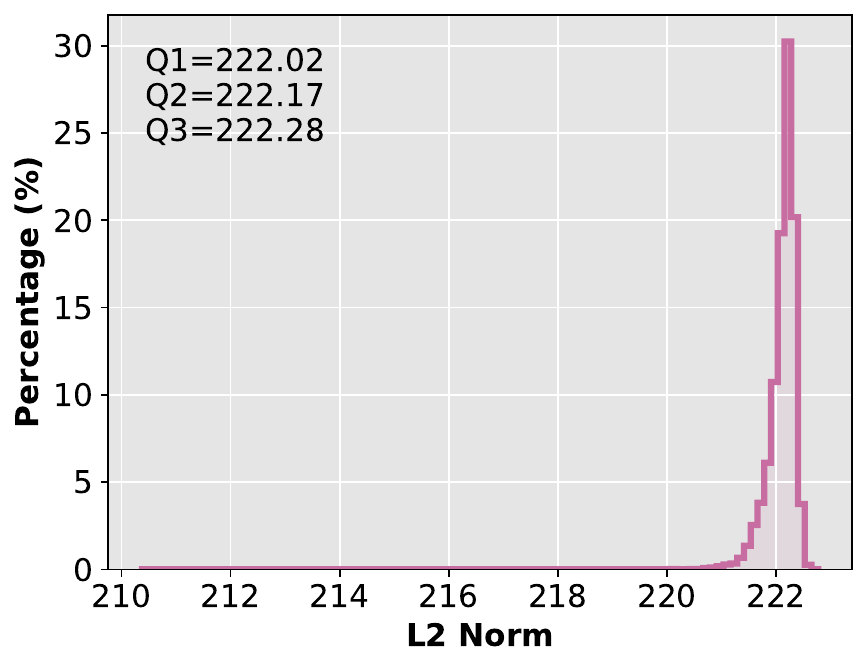}
    }
    \subfloat[WordEmbed Norm]{
        \includegraphics[width=0.24\linewidth]{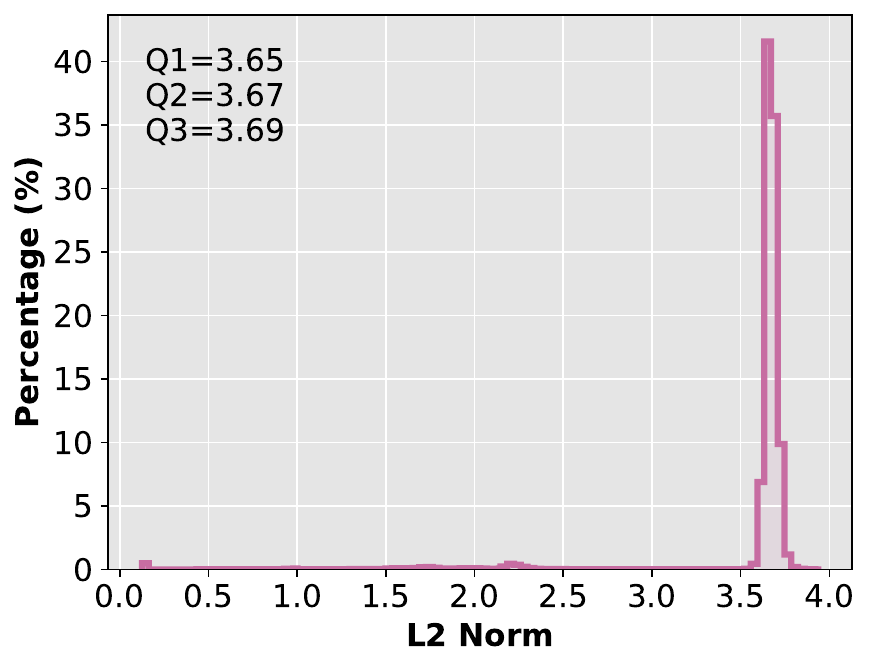}
    }
    \subfloat[ClassEmbed Norm]{
        \includegraphics[width=0.24\linewidth]{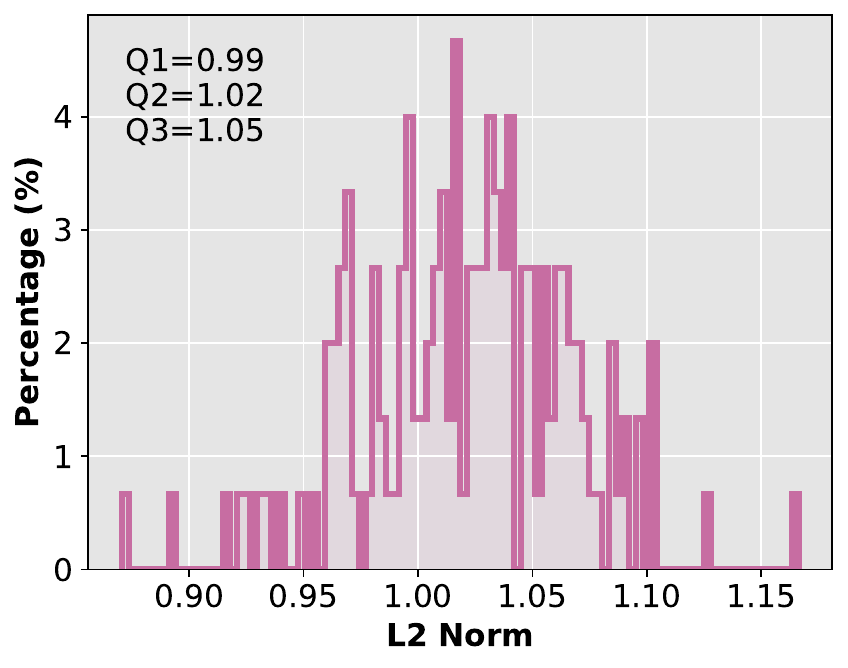}
    }
    \subfloat[Prototype Norm]{
        \includegraphics[width=0.24\linewidth]{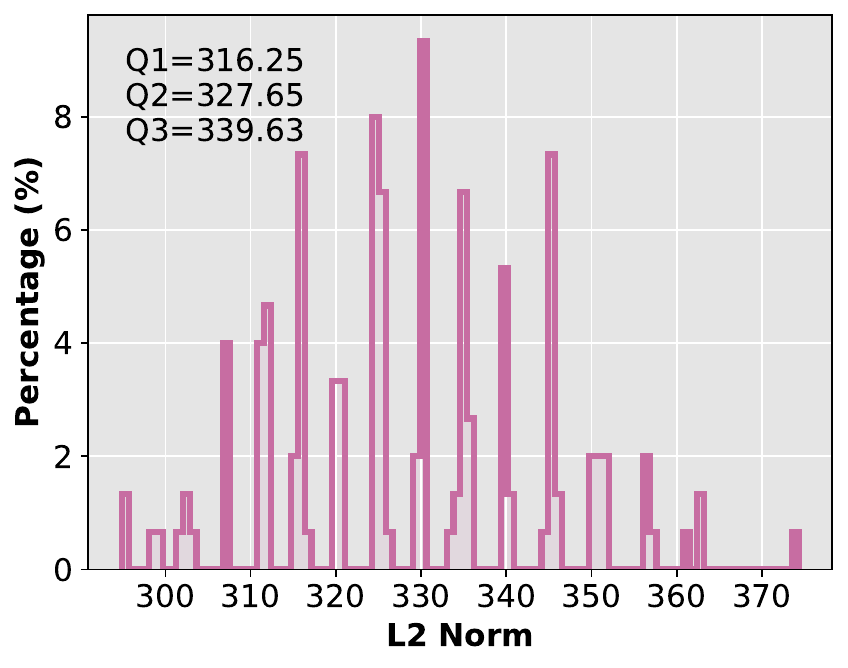}
    }
    
    \caption{The histogram of features and different embeddings of Pythia-160m. Other settings are the same as Figure \ref{fig:metrics_illustration_full_pythia410m}}
    \label{fig:metrics_illustration_full_pythia160m}
\end{figure*}

\begin{figure*}[!t]
    \centering
    \subfloat[Feature Sim]{
        \includegraphics[width=0.24\linewidth]{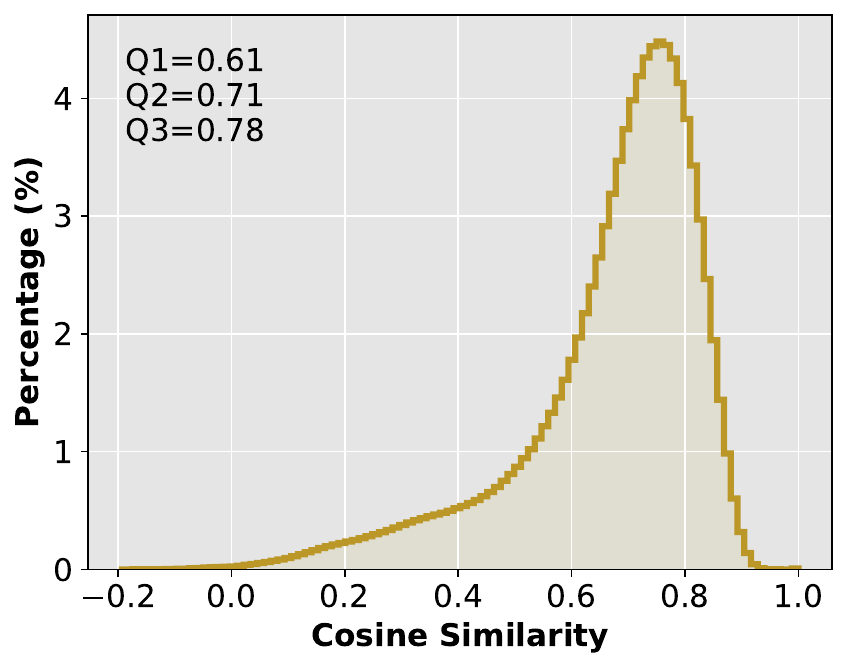}
    }
    \subfloat[Feat-WordEmbed Sim]{
        \includegraphics[width=0.24\linewidth]{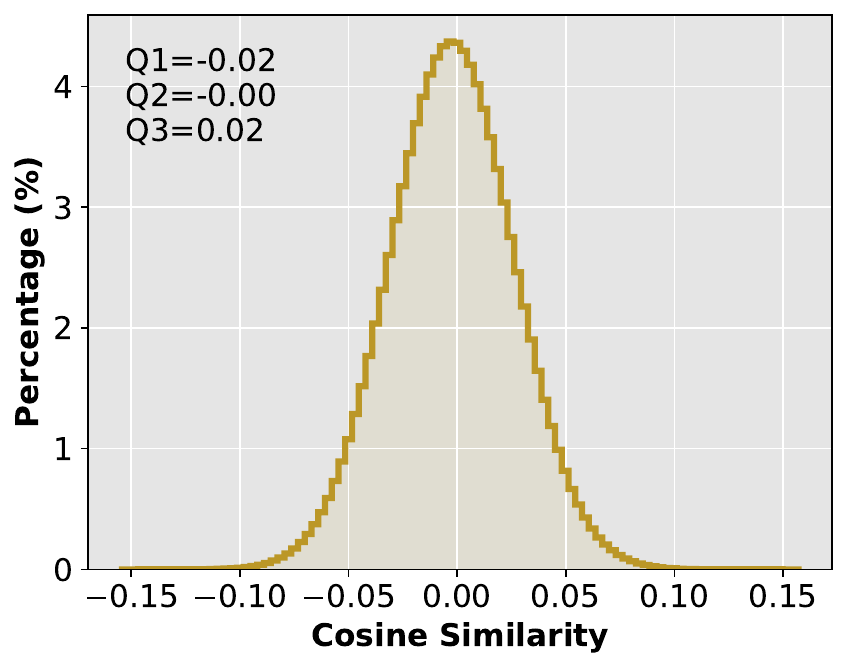}
    }
    \subfloat[Feat-ClassEmbed Sim]{
        \includegraphics[width=0.24\linewidth]{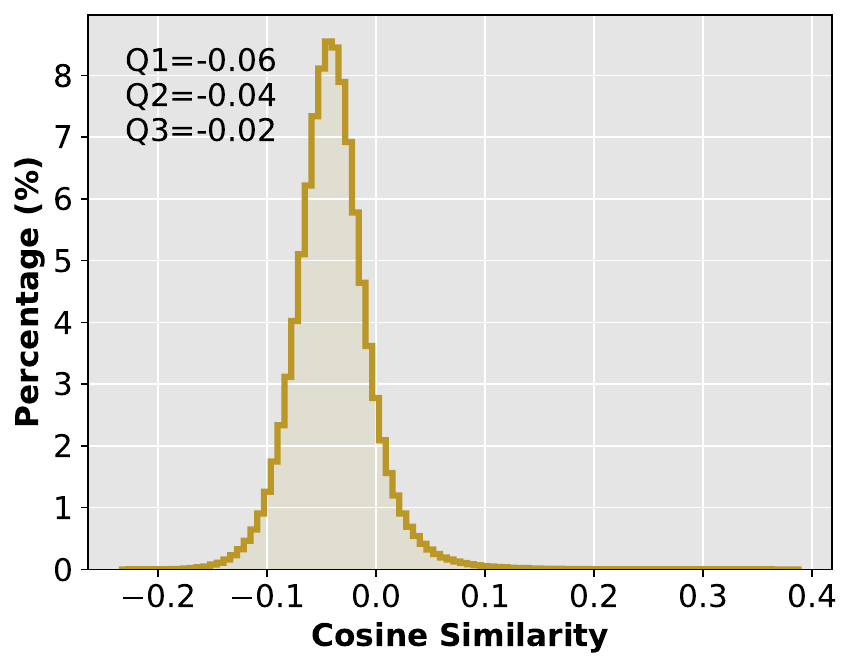}
    }
    \subfloat[Feat-Prototype Sim]{
        \includegraphics[width=0.24\linewidth]{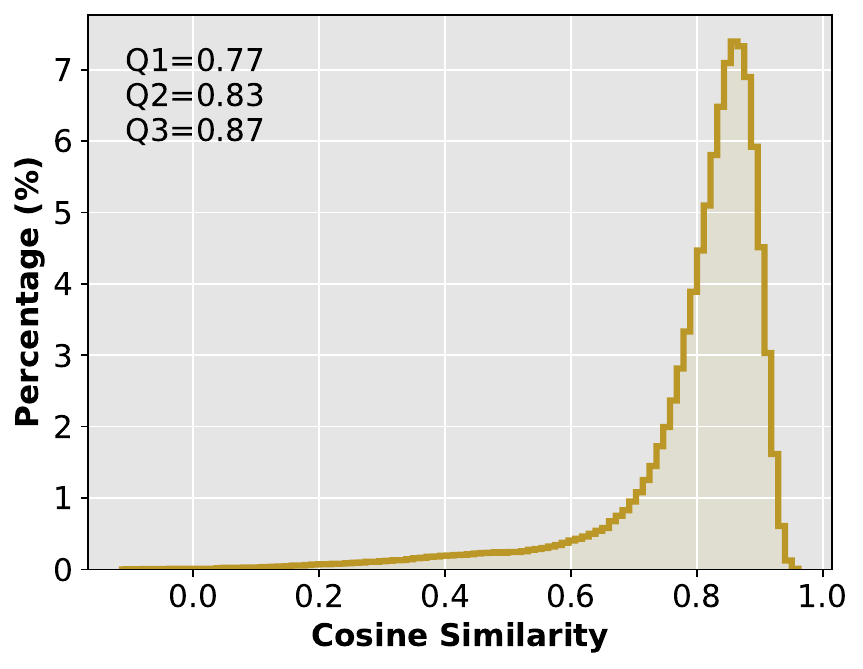}
    }

    \subfloat[Feature Norm]{
        \includegraphics[width=0.24\linewidth]{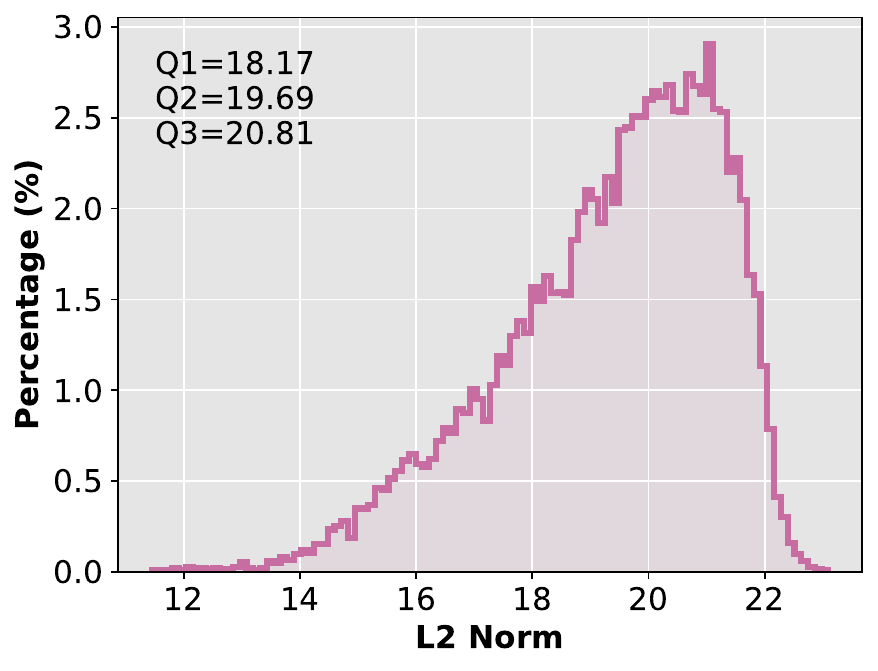}
    }
    \subfloat[WordEmbed Norm]{
        \includegraphics[width=0.24\linewidth]{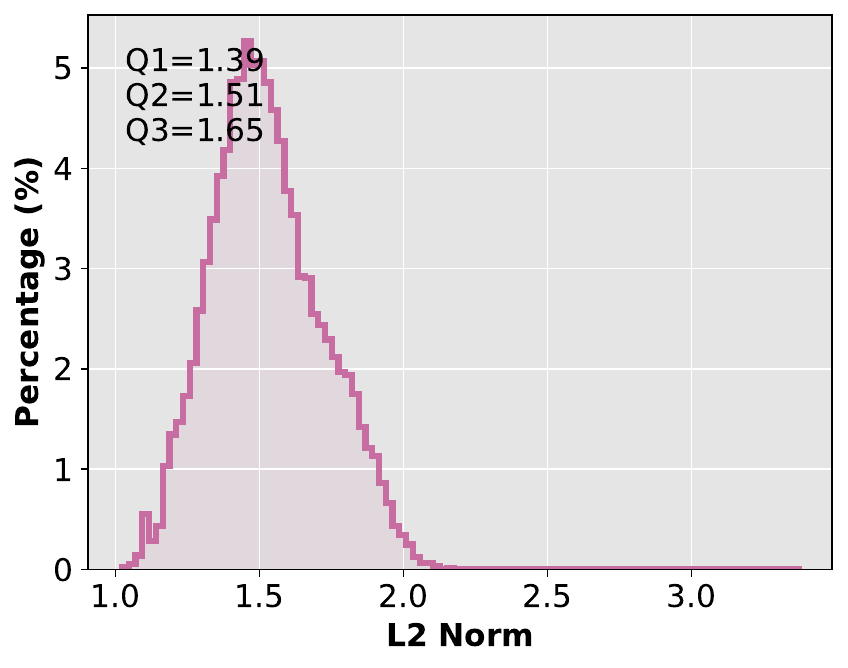}
    }
    \subfloat[ClassEmbed Norm]{
        \includegraphics[width=0.24\linewidth]{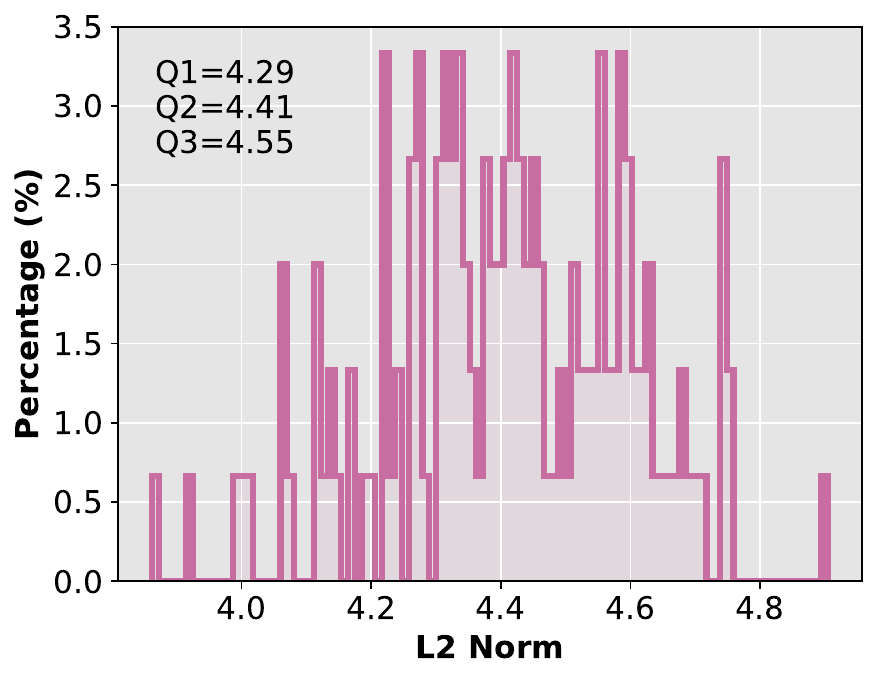}
    }
    \subfloat[Prototype Norm]{
        \includegraphics[width=0.24\linewidth]{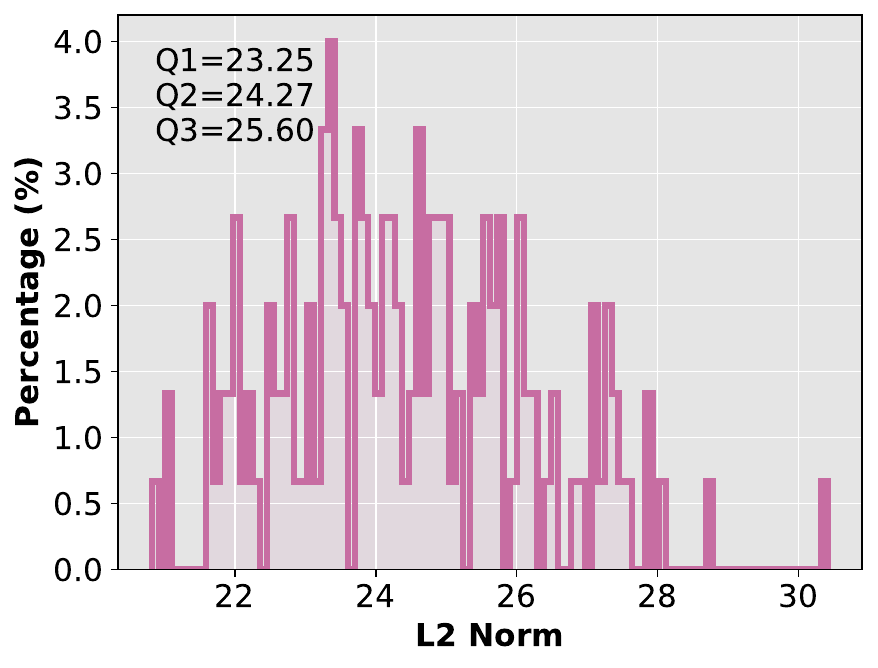}
    }
    
    \caption{The histogram of features and different embeddings of bert-large-cased. Other settings are the same as Figure \ref{fig:metrics_illustration_full_pythia410m}.}
    \label{fig:metrics_illustration_full_bert-large-cased}
\end{figure*}

\begin{figure*}[!t]
    \centering
    \subfloat[Feature Sim]{
        \includegraphics[width=0.24\linewidth]{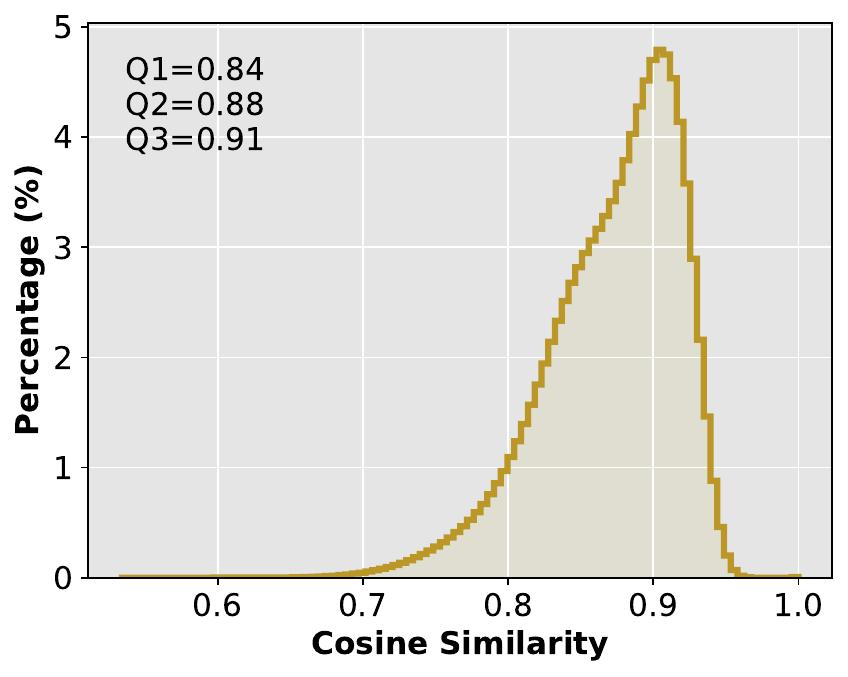}
    }
    \subfloat[Feat-WordEmbed Sim]{
        \includegraphics[width=0.24\linewidth]{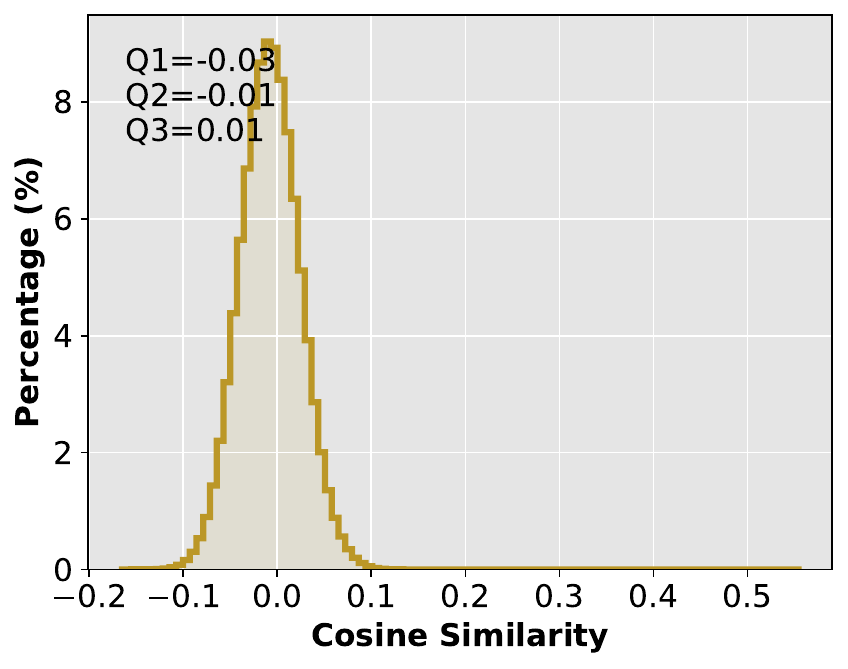}
    }
    \subfloat[Feat-ClassEmbed Sim]{
        \includegraphics[width=0.24\linewidth]{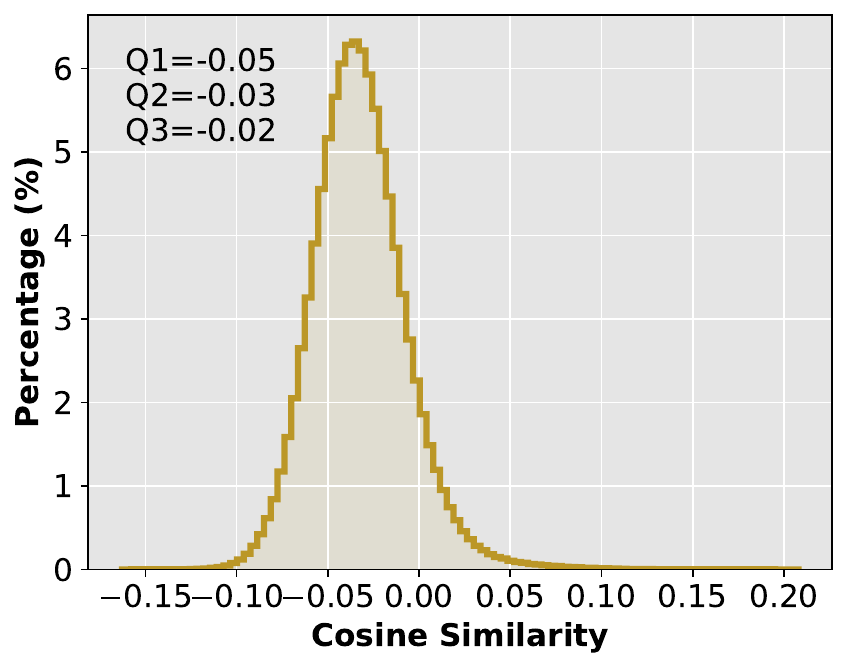}
    }
    \subfloat[Feat-Prototype Sim]{
        \includegraphics[width=0.24\linewidth]{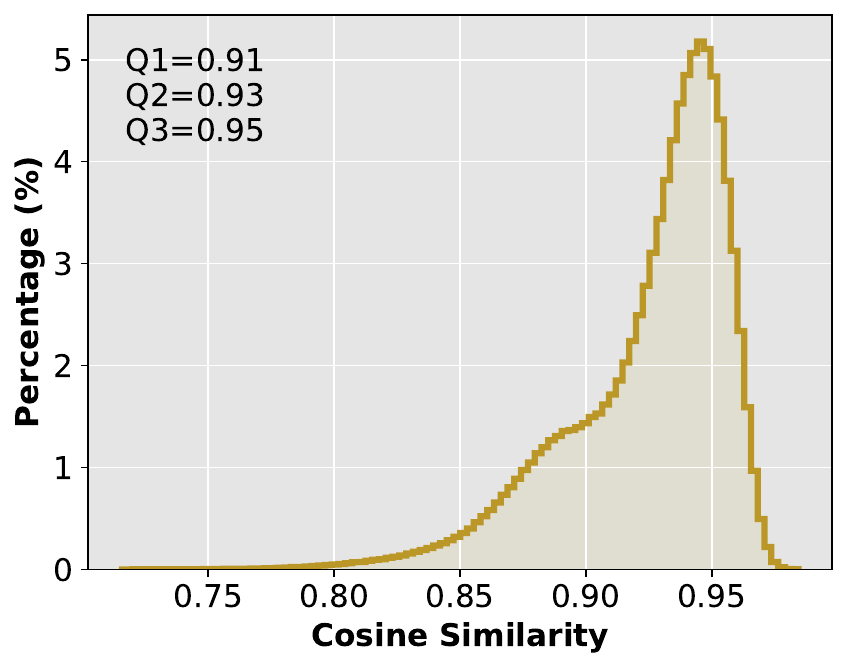}
    }

    \subfloat[Feature Norm]{
        \includegraphics[width=0.24\linewidth]{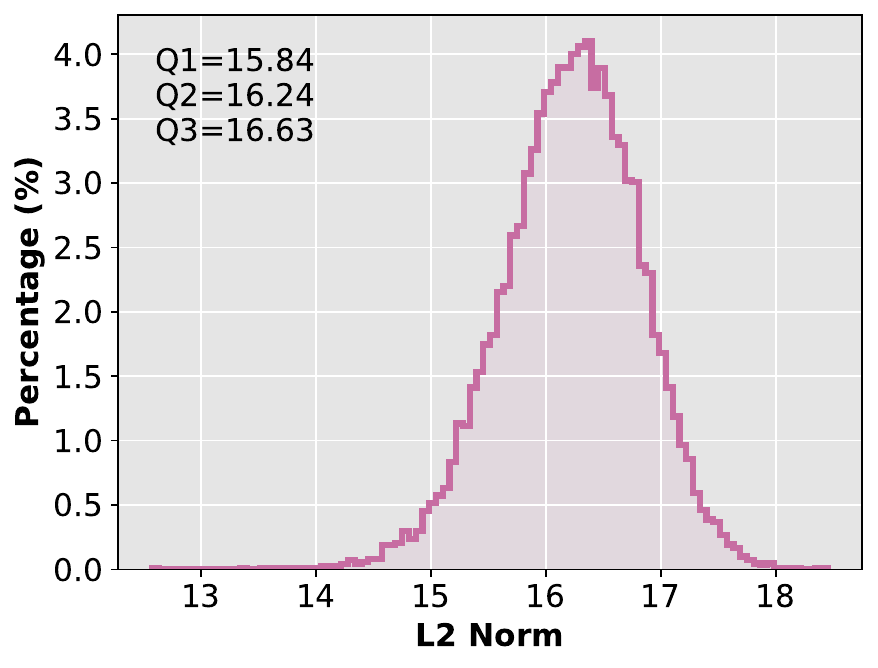}
    }
    \subfloat[WordEmbed Norm]{
        \includegraphics[width=0.24\linewidth]{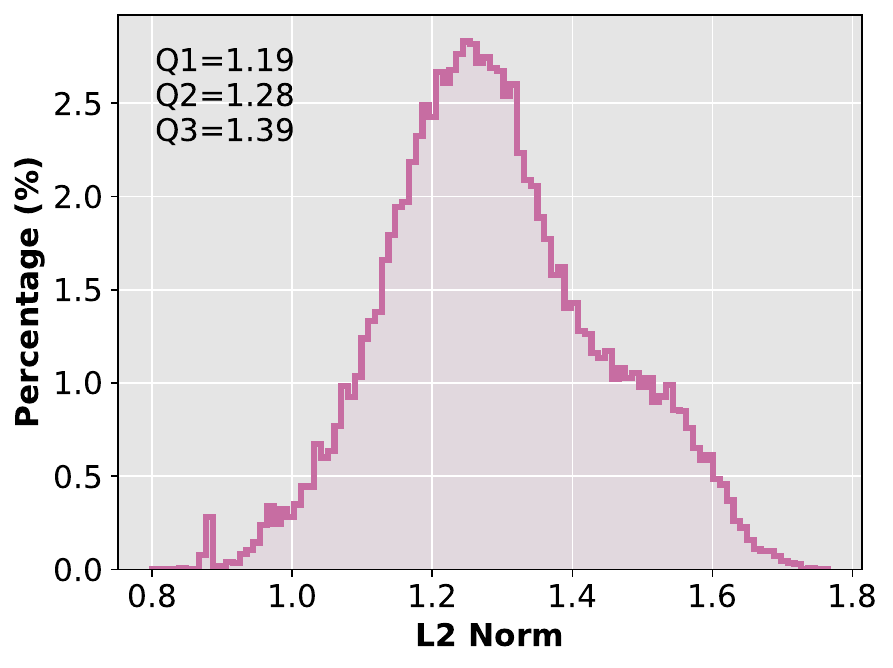}
    }
    \subfloat[ClassEmbed Norm]{
        \includegraphics[width=0.24\linewidth]{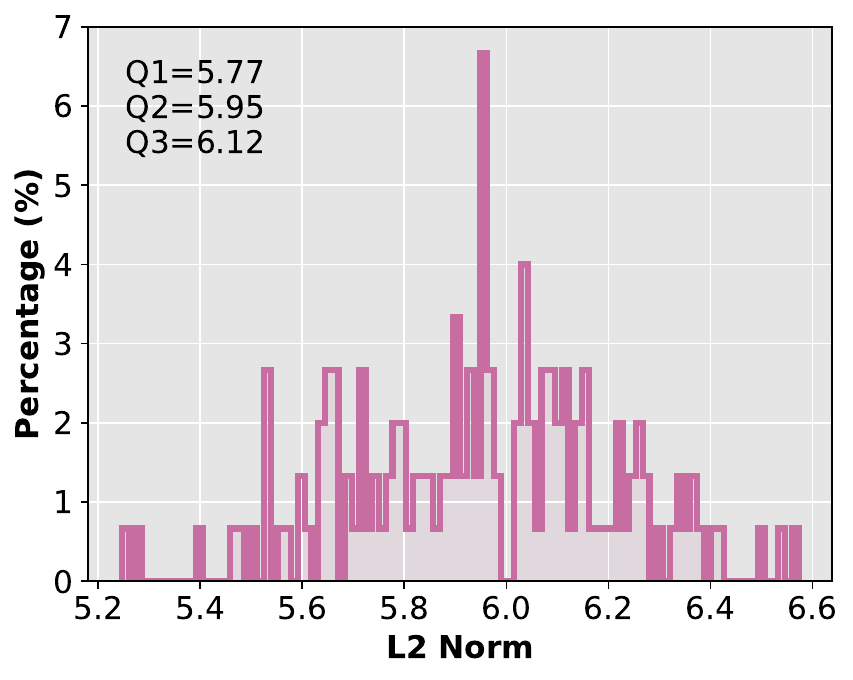}
    }
    \subfloat[Prototype Norm]{
        \includegraphics[width=0.24\linewidth]{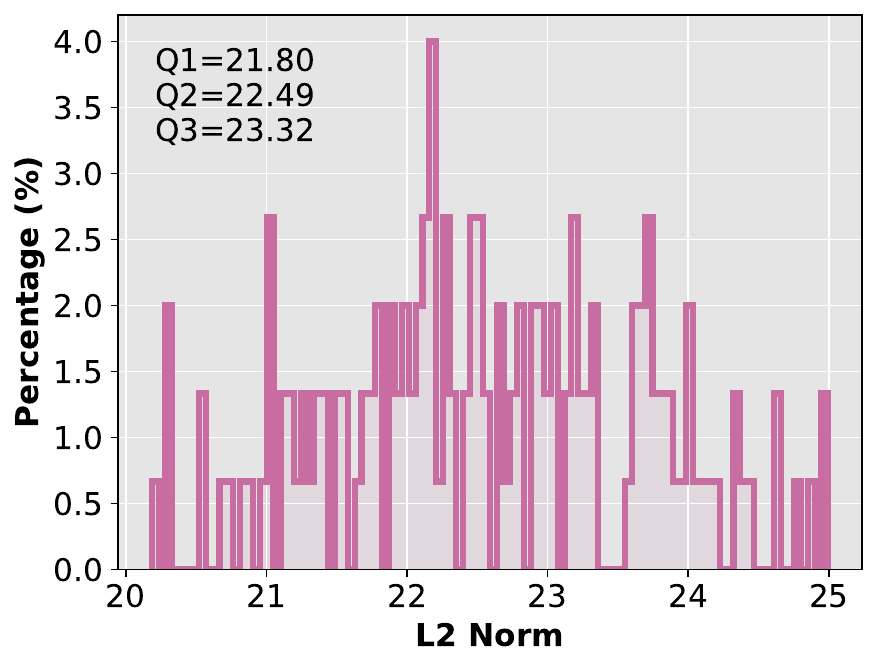}
    }
    
    \caption{The histogram of features and different embeddings of bert-base-cased. Other settings are the same as Figure \ref{fig:metrics_illustration_full_pythia410m}.}
    \label{fig:metrics_illustration_full_bert-base-cased}
\end{figure*}

\subsection{A Closer Look at Features, Word Embeddings, and Class Embeddings}
\label{sec:appendix_probing_study_four_metrics_analysis}
To investigate the significant difference between the four probing metrics, we plot the histogram of the features, the (output) word embeddings, and the class embeddings in (cosine) linear classifiers.
In this research, the features of PLMs refer to the last hidden states of the last word in generative backbones and the [CLS] token in discriminant backbones.
The class prototype refers to the class feature centres.
The class embeddings refer to the row vectors of the weight matrix in linear layers.
For example, for a linear layer whose input and output dimensions are 768 and 20, respectively, its weight matrix has the shape $20 \times 768$.
Then, each row vector corresponds to a certain category, and its shape is $1 \times 768$.
When using linear classifiers for prediction, the logits of a certain category are computed as the dot product between the feature and the class embeddings of that category.

During pre-training, the logits over vocabulary are computed as the dot product between features and word embeddings. 
In the probing study, the logits over categories are computed as the dot product between features and class embeddings. 
Therefore, we need to figure out the relationship between features, word embeddings, and class embeddings.
We note that there is a dense layer (i.e., linear layer) between backbones and linear classifiers in BERT, and we ignore it for simplicity.

The histogram of the cosine similarity and L2 norm is provided in Figure \ref{fig:metrics_illustration_full_pythia410m} (Pythia-410m), \ref{fig:metrics_illustration_full_pythia160m} (Pythia-160m), \ref{fig:metrics_illustration_full_bert-large-cased} (bert-large-cased), \ref{fig:metrics_illustration_full_bert-base-cased} (bert-base-cased).
The dataset is CLINC150.
We note that the PLMs are loaded directly without fine-tuning.
The features are computed on the whole training set of CLINC150.
The word embeddings are loaded directly from PLMs.
The class embeddings of the linear probing classifiers are obtained by training probing classifiers.

From the result, we have the following findings:
\begin{itemize}
    \item The features of PLMs have high cosine similarity, indicating that they fall in a cone space.
    \item The features are almost orthogonal to the word embeddings in all backbones except for Pythia-160m.
    \item The features are almost orthogonal to the class embeddings in all backbones.
    \item The features have high cosine similarity with the prototypes.
    \item The L2 norm of word embeddings, class embeddings have large discrepancy for all backbones except for Pythia-160m.
\end{itemize}
These findings explain why the linear classifier is the best option to utilize the backbone's knowledge.
Specifically, the word embedding layer is also a linear classifier for pre-training, and thus, the features are most discriminative when multiplied with the class embeddings of linear classifiers.
Furthermore, the discrepancy of the L2 norm of word and class embeddings suggests that the norm contains the prior knowledge obtained from pre-training.
We illustrate the four probing metrics in Figure \ref{fig:metrics_illustration}.

The word embeddings in Pythia-160m have high cosine similarity with features.
We speculate the reason is that the parameters are not enough for generalization with causal language modelling loss.

\begin{figure*}[!t]
    \centering
    \subfloat[TC+Before SEQ]{
        \includegraphics[width=0.45\linewidth]{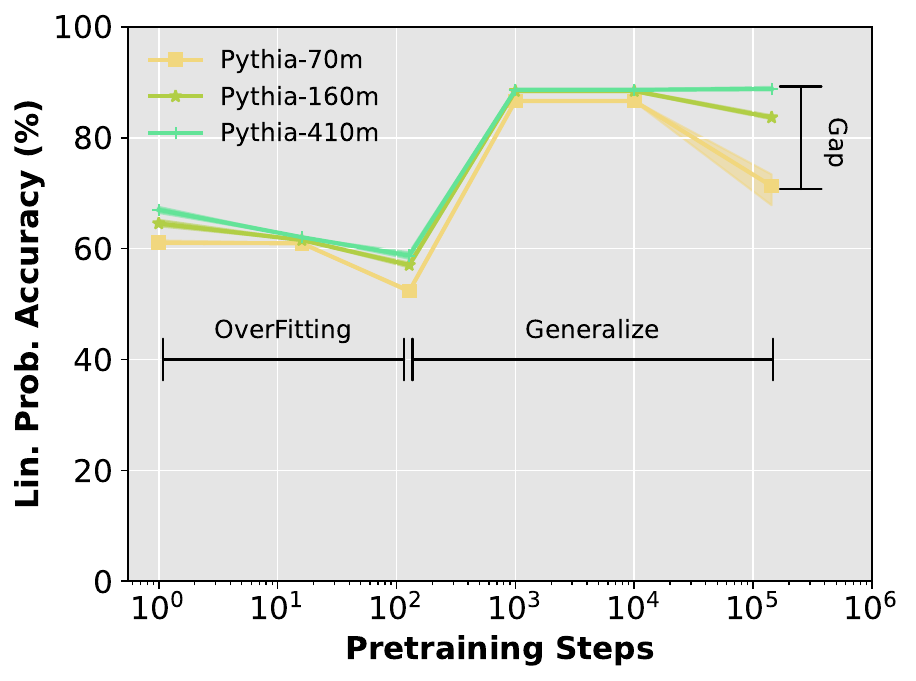}
    }
    \subfloat[TC+After SEQ]{
        \includegraphics[width=0.45\linewidth]{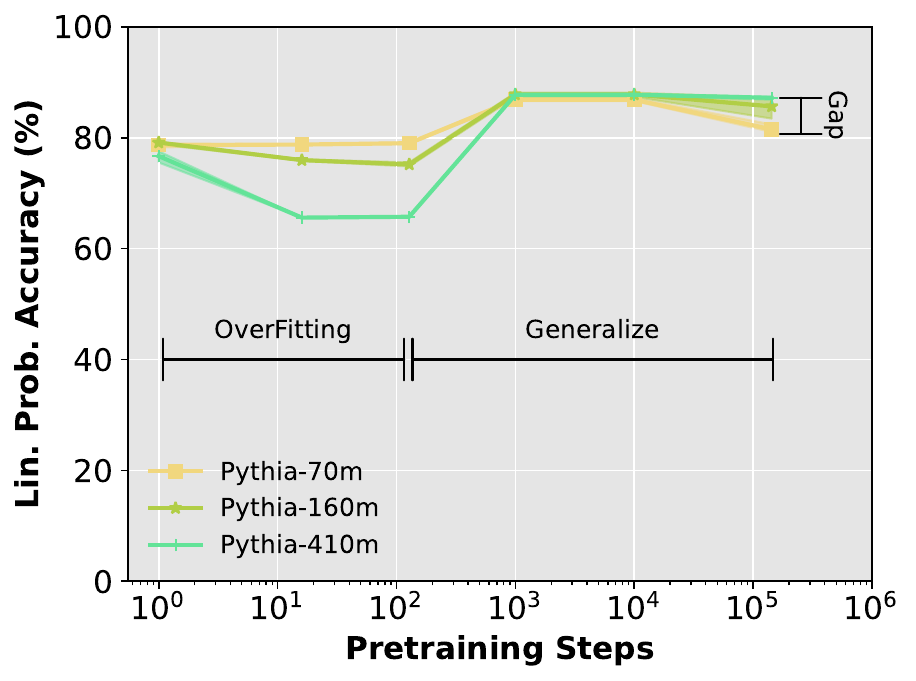}
    }
    
    \caption{The linear probing performance on PLMs with different pre-training steps. (a) and (b) are evaluated before and after incremental learning using SEQ. ``TC'' represents that the model is evaluated on the Class-Incremental Text Classification.}
    \label{fig:probing_pretraining_tc}
\end{figure*}

\begin{figure*}[!t]
    \centering
    \subfloat[Before SEQ w/o pre-training]{
        \includegraphics[width=0.24\linewidth]{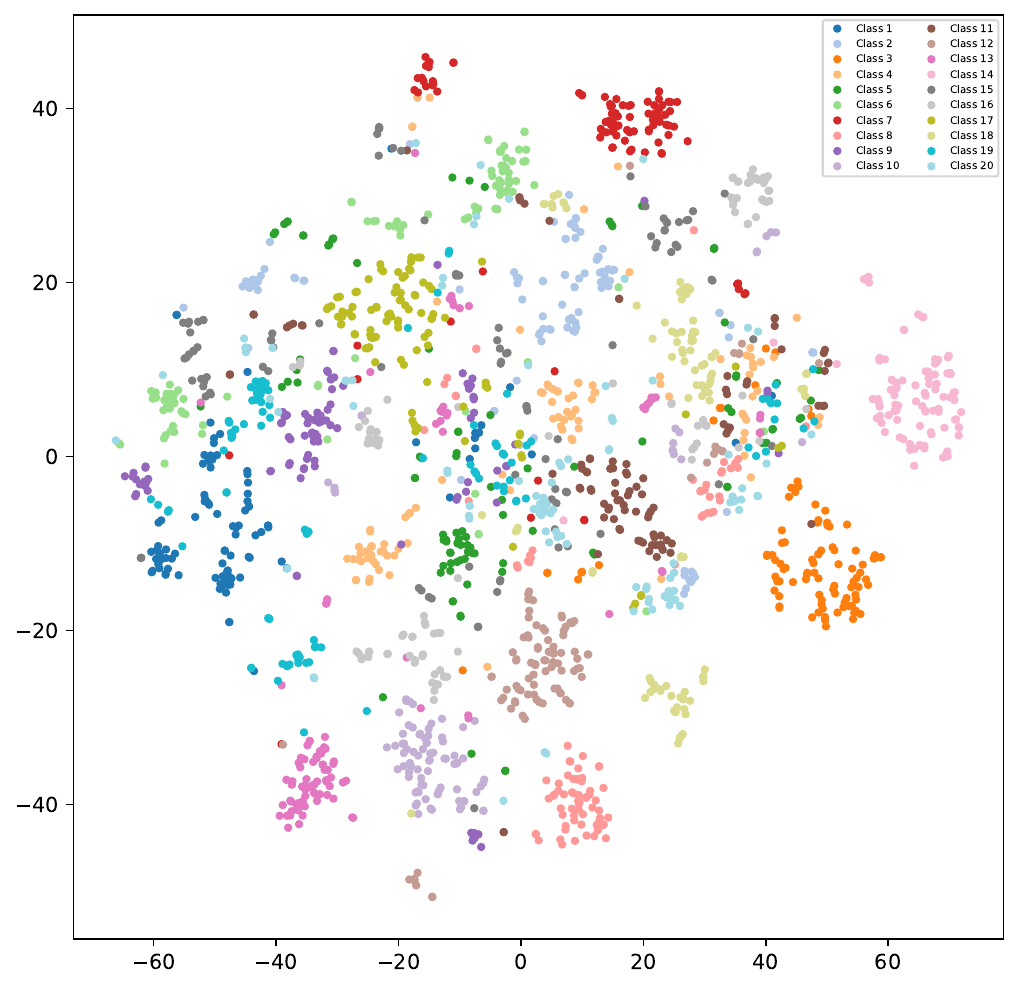}
        \label{fig:tsne_intent_a}
    }
    \subfloat[After SEQ w/o pre-training]{
        \includegraphics[width=0.24\linewidth]{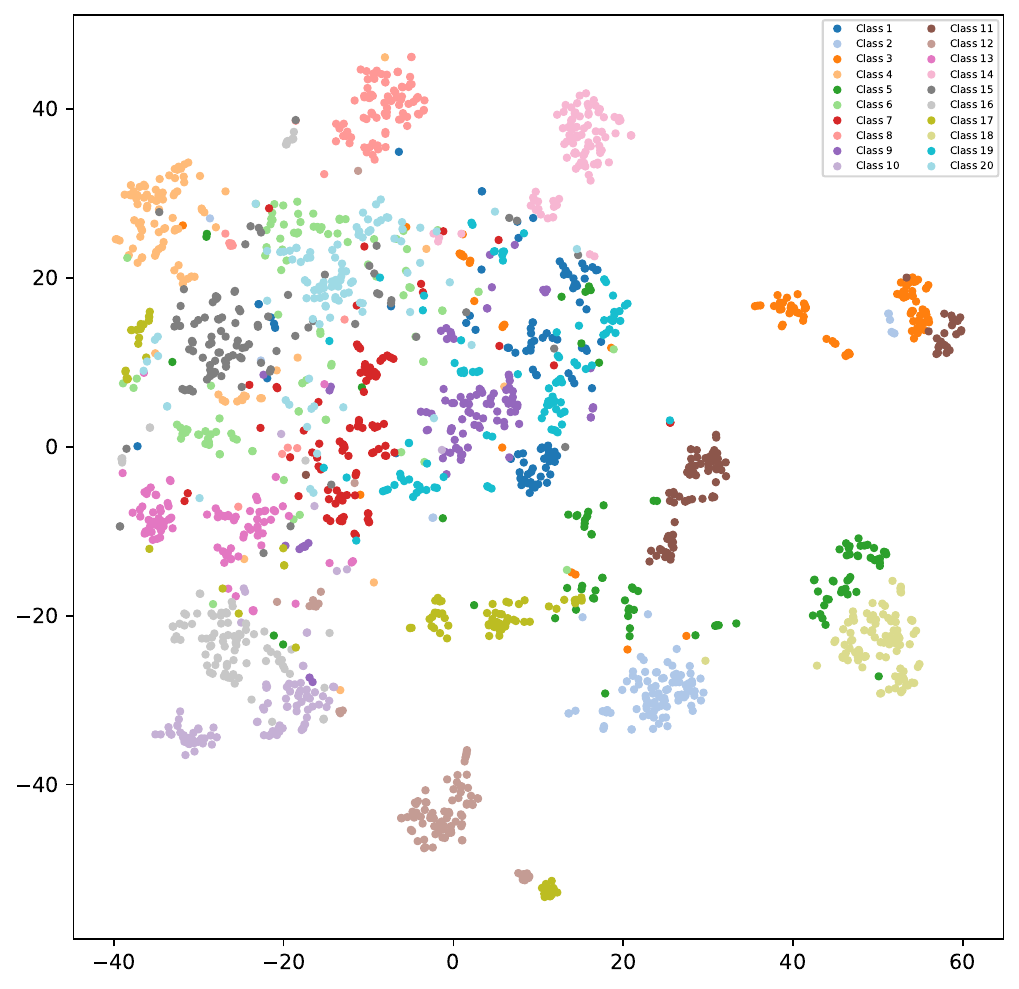}
        \label{fig:tsne_intent_b}
    }
    \subfloat[Before SEQ w/ pre-training]{
        \includegraphics[width=0.24\linewidth]{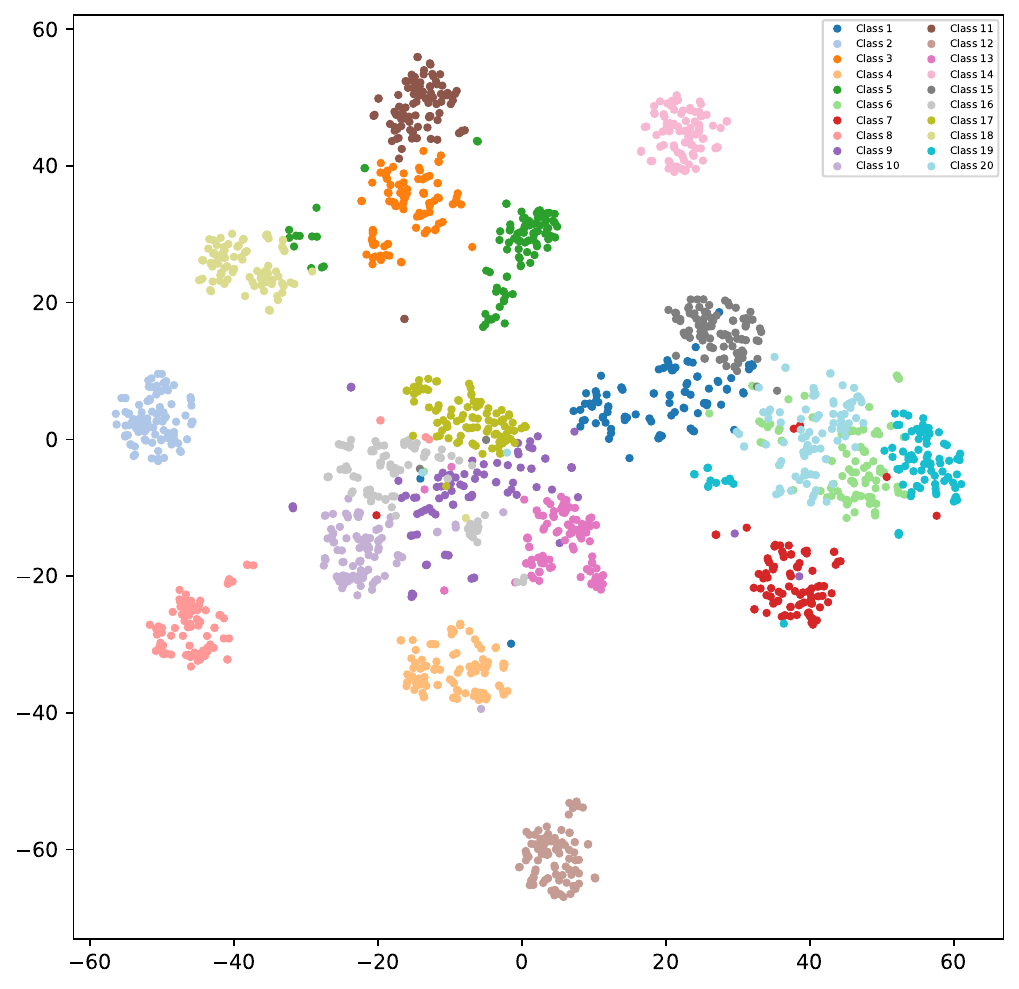}
        \label{fig:tsne_intent_c}
    }
    \subfloat[After SEQ w/ pre-training]{
        \includegraphics[width=0.24\linewidth]{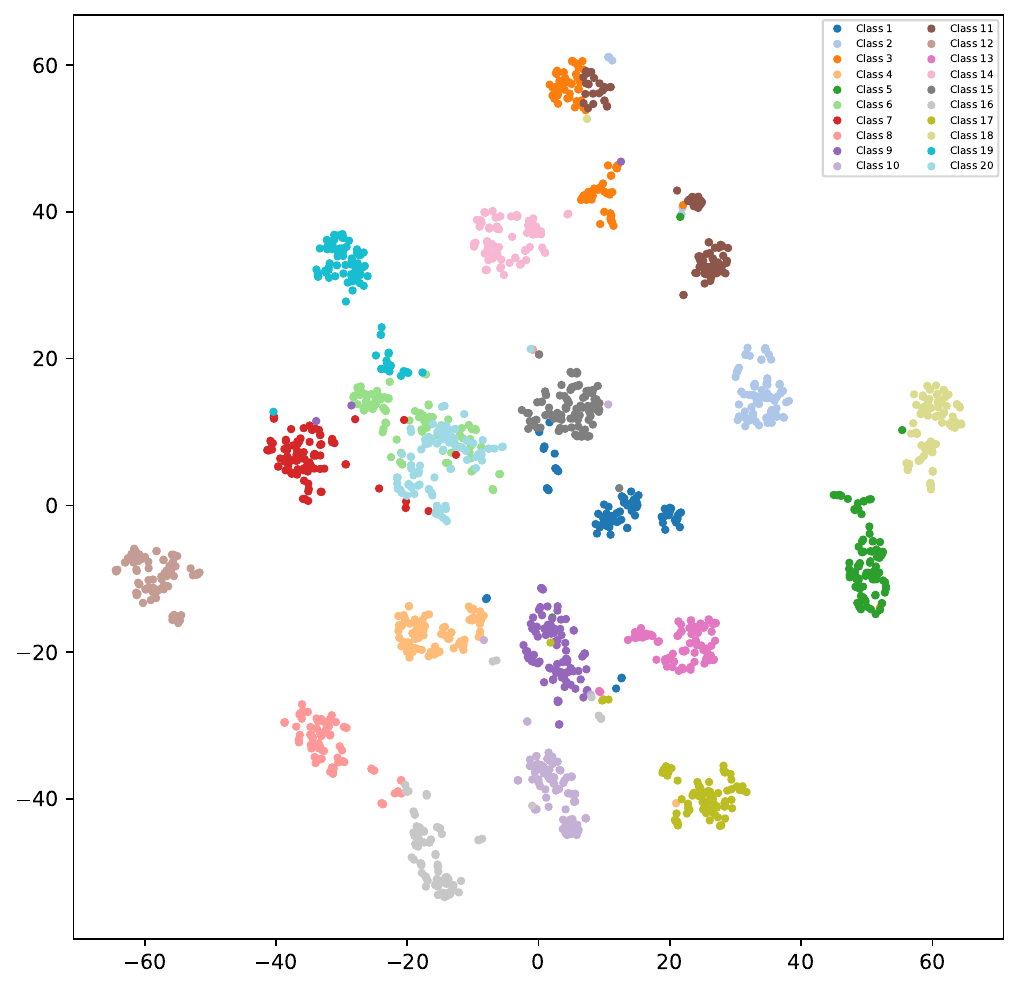}
        \label{fig:tsne_intent_d}
    }
    
    \caption{The t-SNE visualization of features on CLINC150. The backbone model is Pythia-410m. Only the first 20 classes are visualized for clarity. (a)(b): the backbone model is randomly initialized without pre-training; (c)(d): the backbone model is pre-trained.}
    \label{fig:tsne_intent}
\end{figure*}

\begin{figure*}[!t]
    \centering
    \subfloat[Before SEQ w/o pre-training]{
        \includegraphics[width=0.24\linewidth]{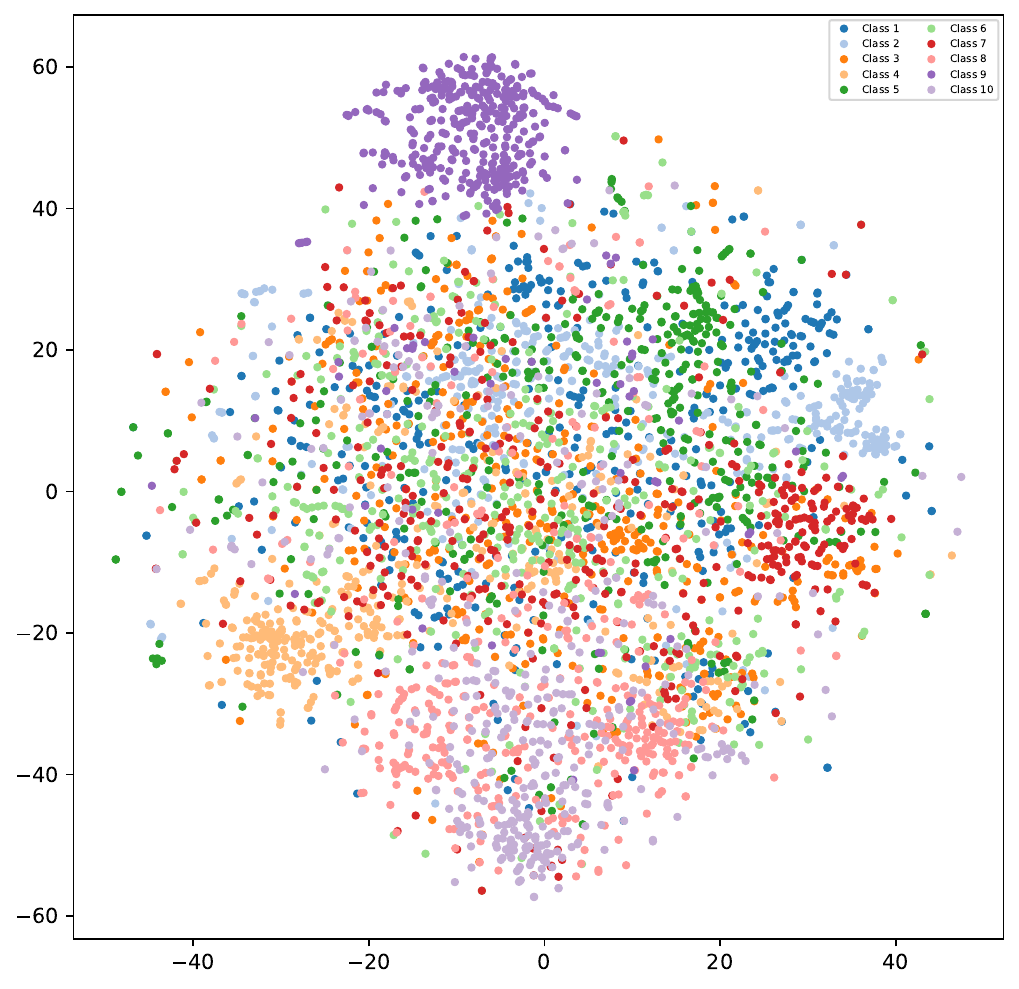}
        \label{fig:tsne_RE_a}
    }
    \subfloat[After SEQ w/o pre-training]{
        \includegraphics[width=0.24\linewidth]{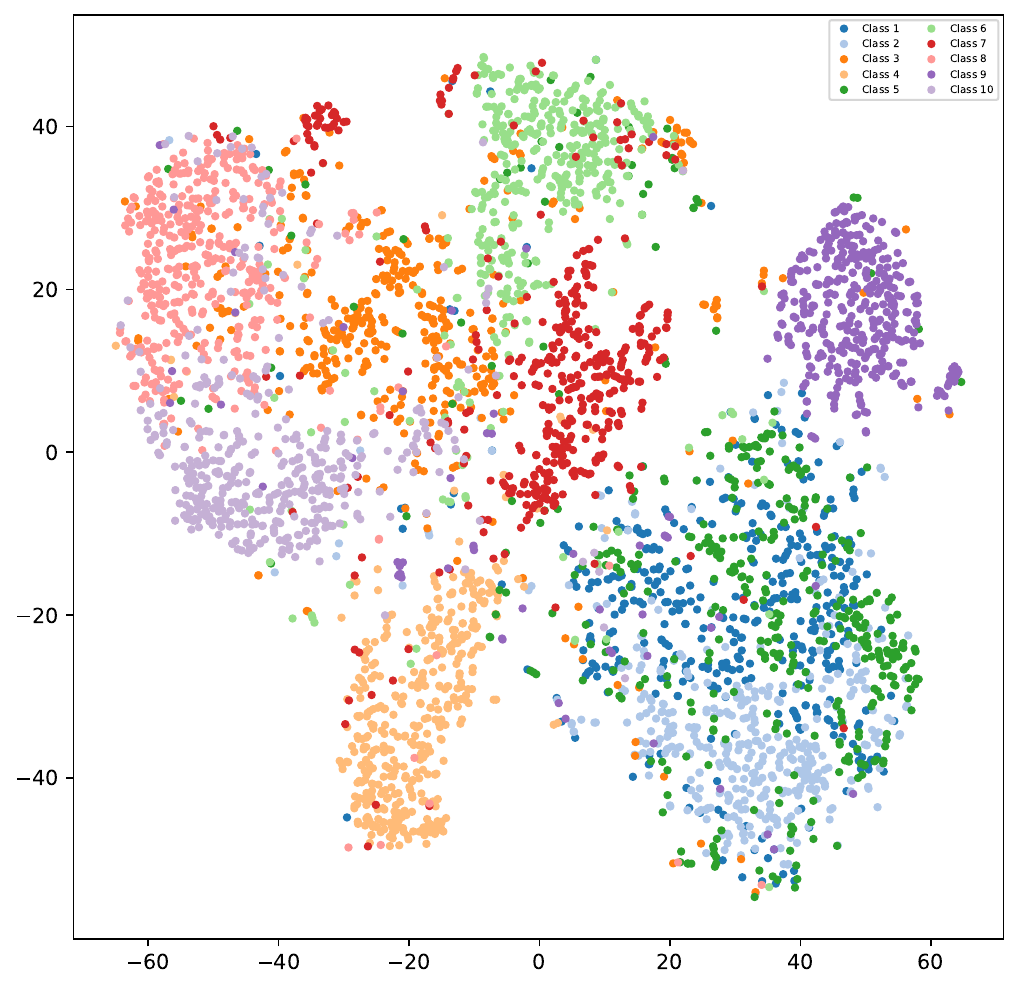}
        \label{fig:tsne_RE_b}
    }
    \subfloat[Before SEQ w/ pre-training]{
        \includegraphics[width=0.24\linewidth]{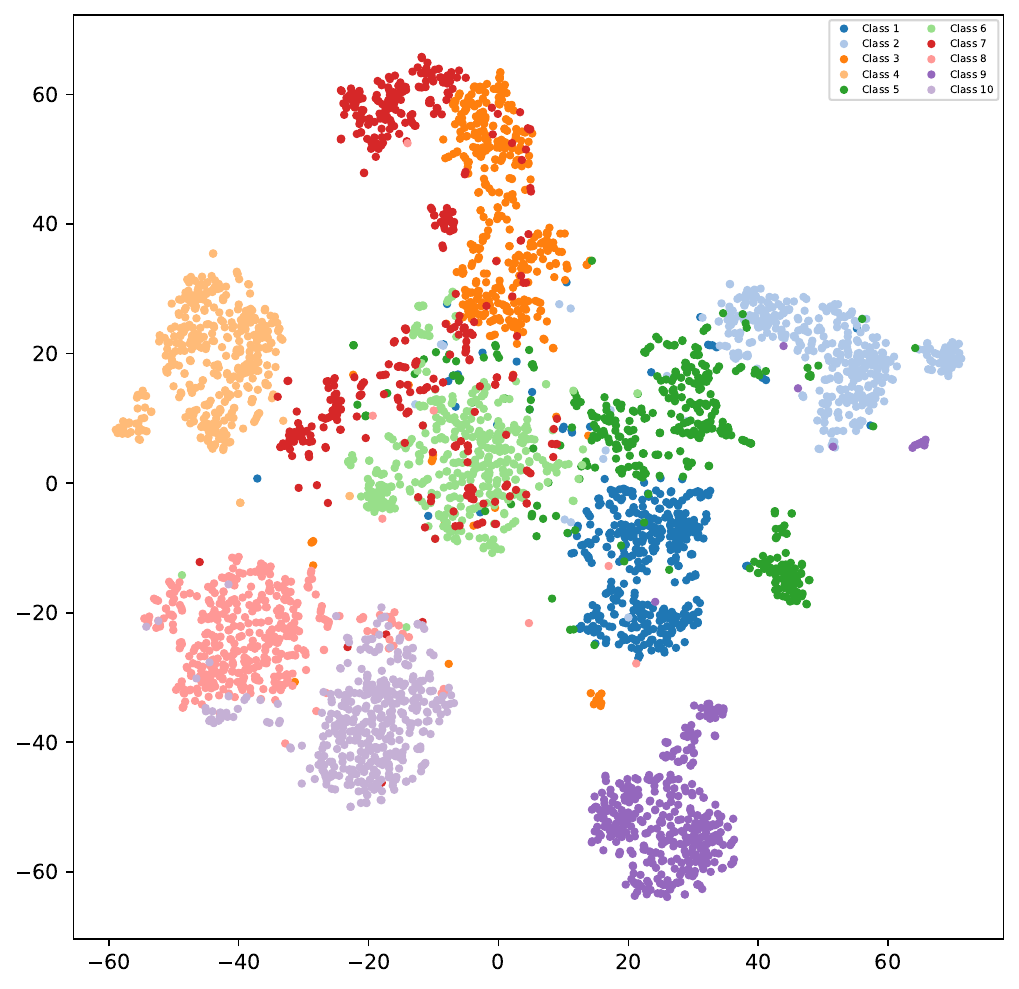}
        \label{fig:tsne_RE_c}
    }
    \subfloat[After SEQ w/ pre-training]{
        \includegraphics[width=0.24\linewidth]{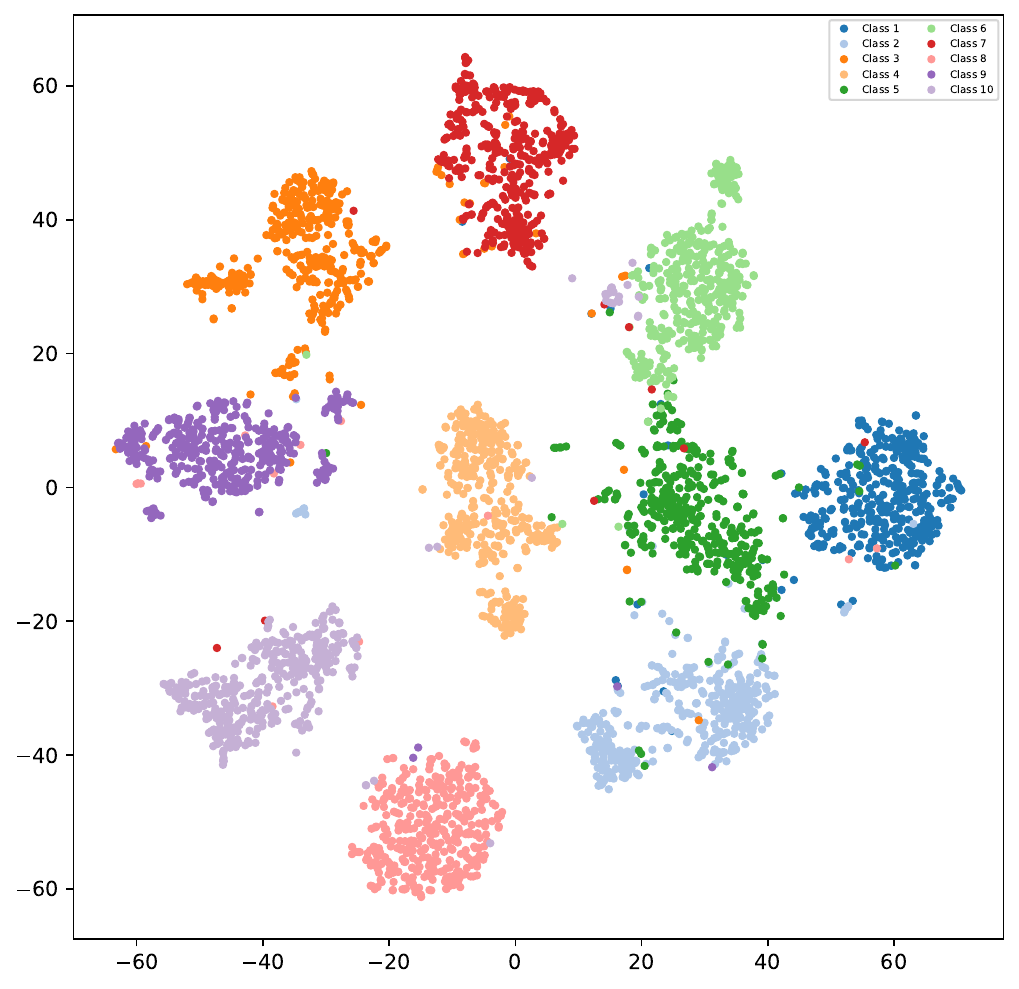}
        \label{fig:tsne_RE_d}
    }
    
    \caption{The t-SNE visualization of features on FewRel. The backbone model is Pythia-410m. Only the first 10 classes are visualized for clarity. (a)(b): the backbone model is randomly initialized without pre-training; (c)(d) the backbone model is pre-trained.}
    \label{fig:tsne_RE}
\end{figure*}

\begin{figure*}[!t]
    \centering
    \subfloat[Observed Classifier]{
        \includegraphics[width=0.24\linewidth]{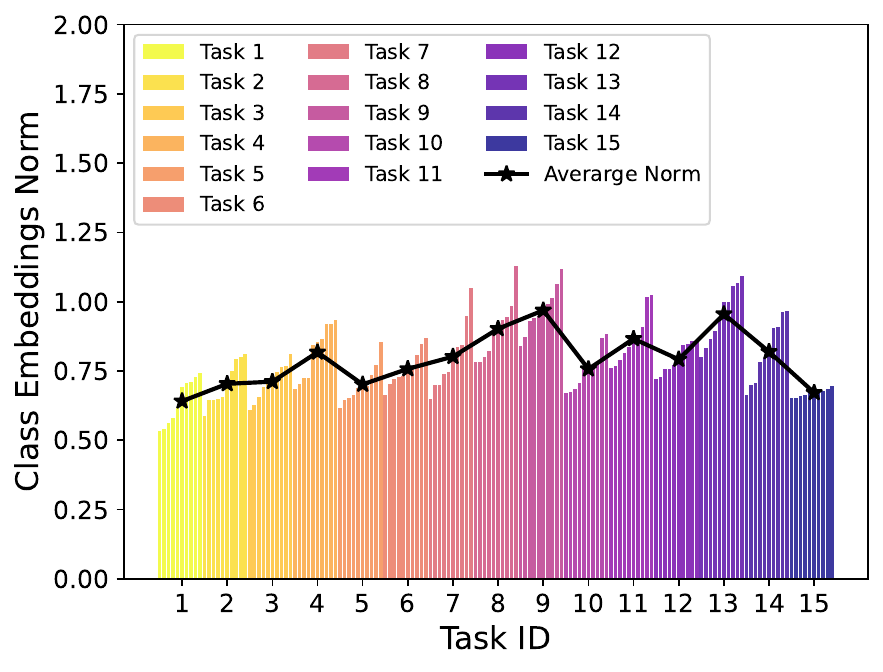}
    }
    \subfloat[Probing Classifier]{
        \includegraphics[width=0.24\linewidth]{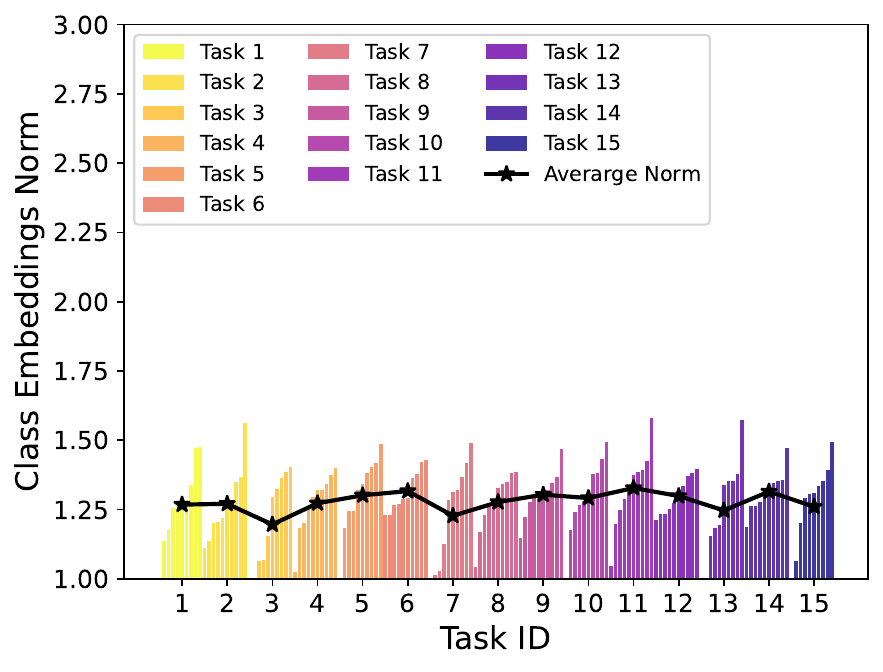}
    }
    \subfloat[Observed Classifier]{
        \includegraphics[width=0.24\linewidth]{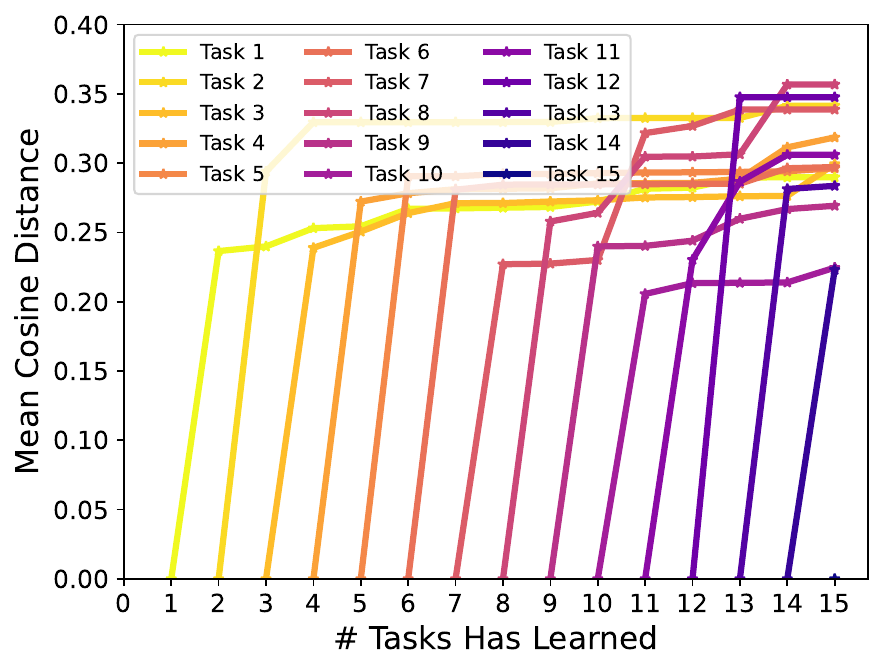}
    }
    \subfloat[Probing Classifier]{
        \includegraphics[width=0.24\linewidth]{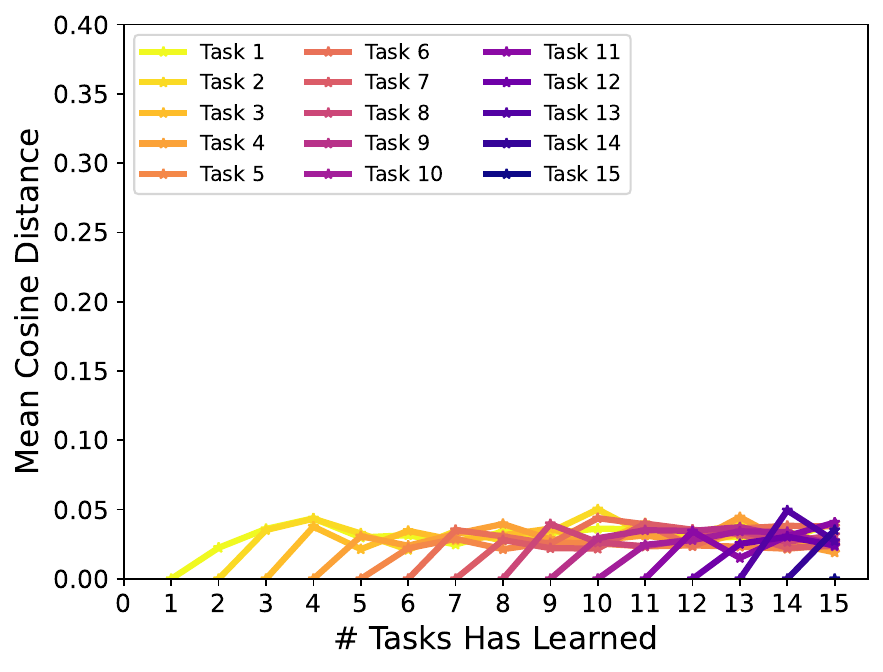}
    }
    
    \caption{Comparison between the observed and probing classifiers after SEQ on class-incremental intent classification. The backbone is Pythia-410 m and \textit{frozen} during IL. (a)(b) show the average norm of the class embeddings of each task; (c)(d) show the average moving distance of the class embeddings of each task.}
    \label{fig:comparison_class_norm_moving_distance_gen_frozen}
\end{figure*}

\begin{figure*}[!t]
    \centering
    \subfloat[Observed Classifier]{
        \includegraphics[width=0.24\linewidth]{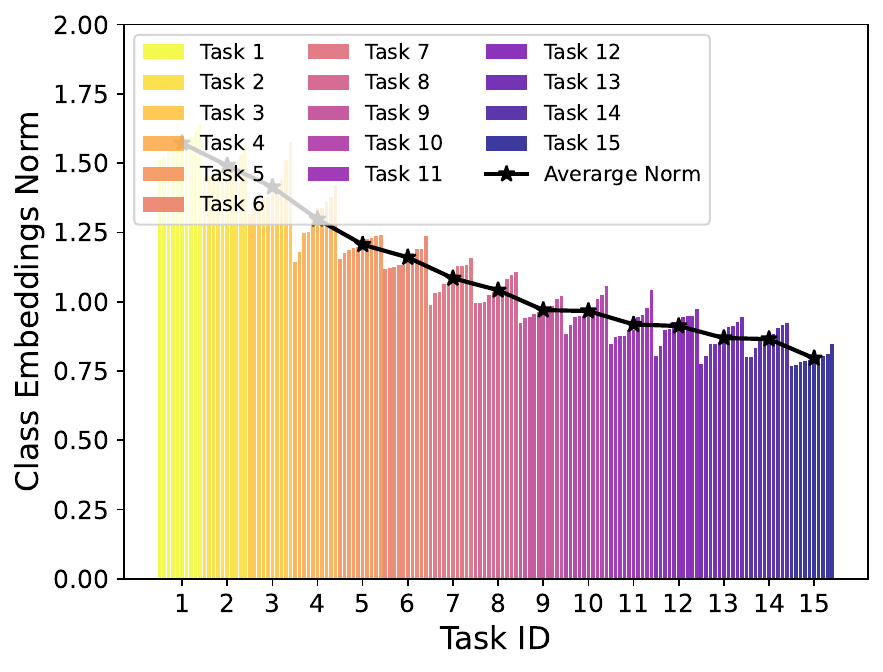}
    }
    \subfloat[Probing Classifier]{
        \includegraphics[width=0.24\linewidth]{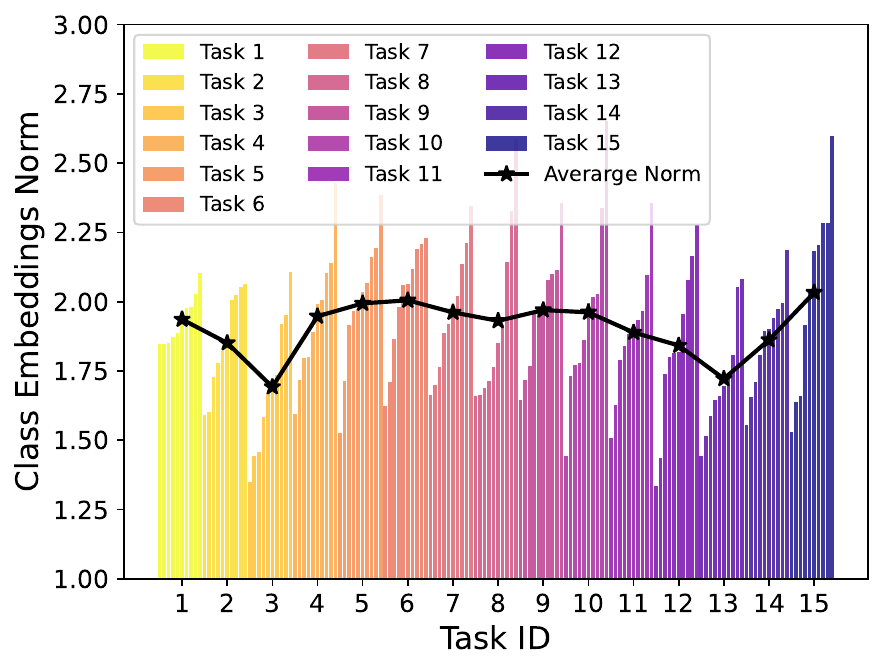}
    }
    \subfloat[Observed Classifier]{
        \includegraphics[width=0.24\linewidth]{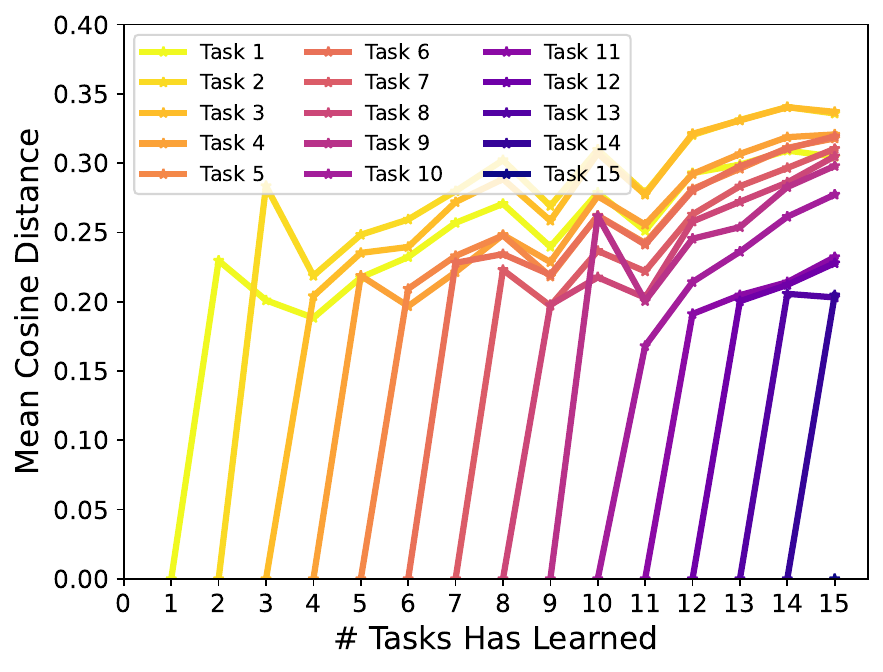}
    }
    \subfloat[Probing Classifier]{
        \includegraphics[width=0.24\linewidth]{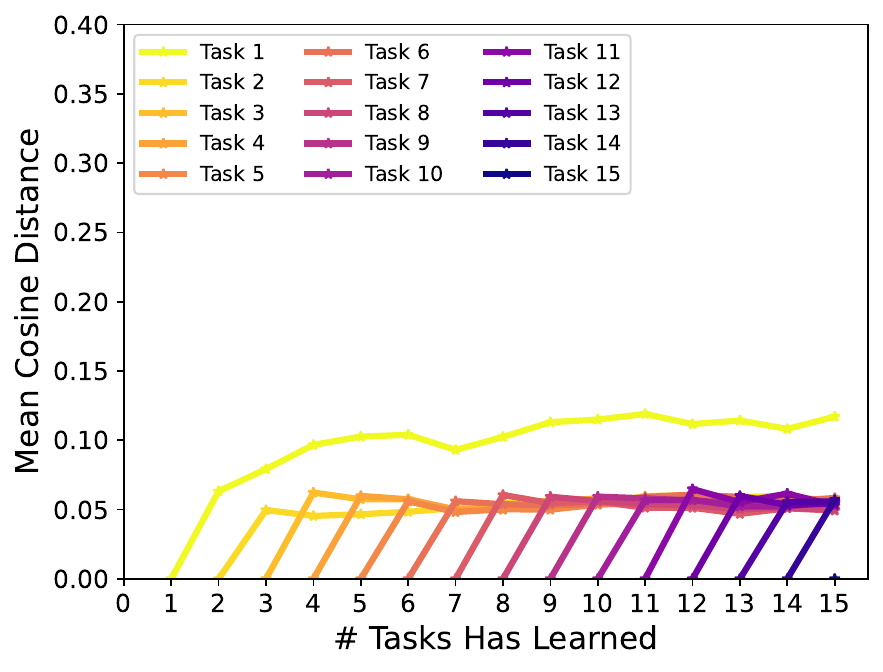}
    }
    
    \caption{Comparison between the observed and probing classifiers after SEQ on class-incremental intent classification. The backbone is bert-large-cased. (a)(b) show the average norm of the class embeddings of each task; (c)(d) show the average moving distance of the class embeddings of each task.}
    \label{fig:comparison_class_norm_moving_distance_dis}
\end{figure*}

\begin{figure*}[!t]
    \centering
    \subfloat[Observed Classifier]{
        \includegraphics[width=0.24\linewidth]{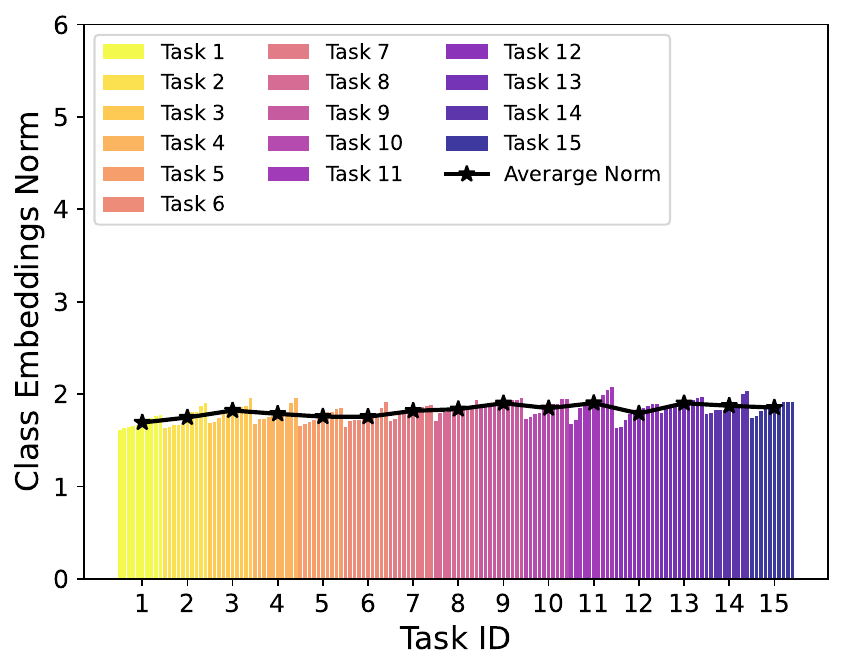}
    }
    \subfloat[Probing Classifier]{
        \includegraphics[width=0.24\linewidth]{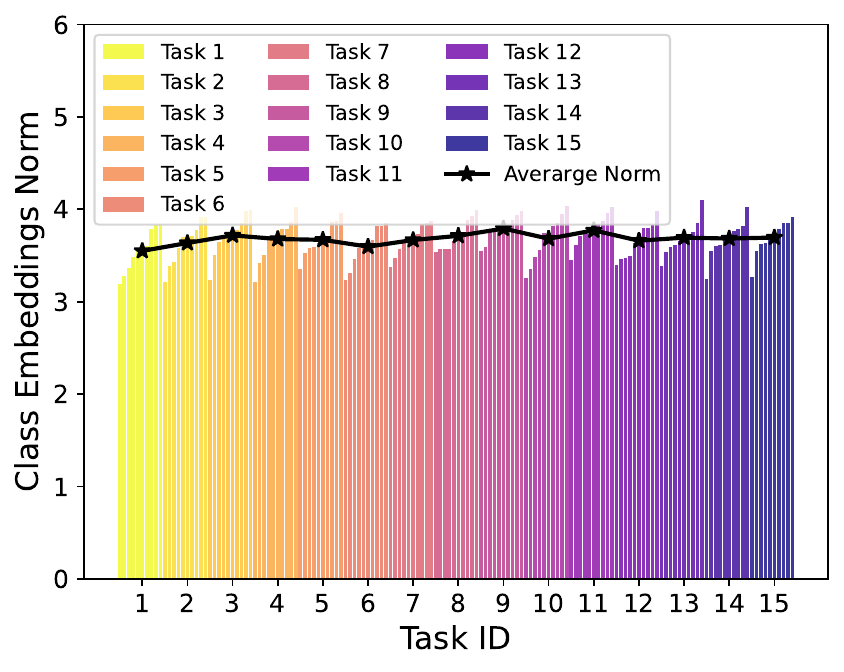}
    }
    \subfloat[Observed Classifier]{
        \includegraphics[width=0.24\linewidth]{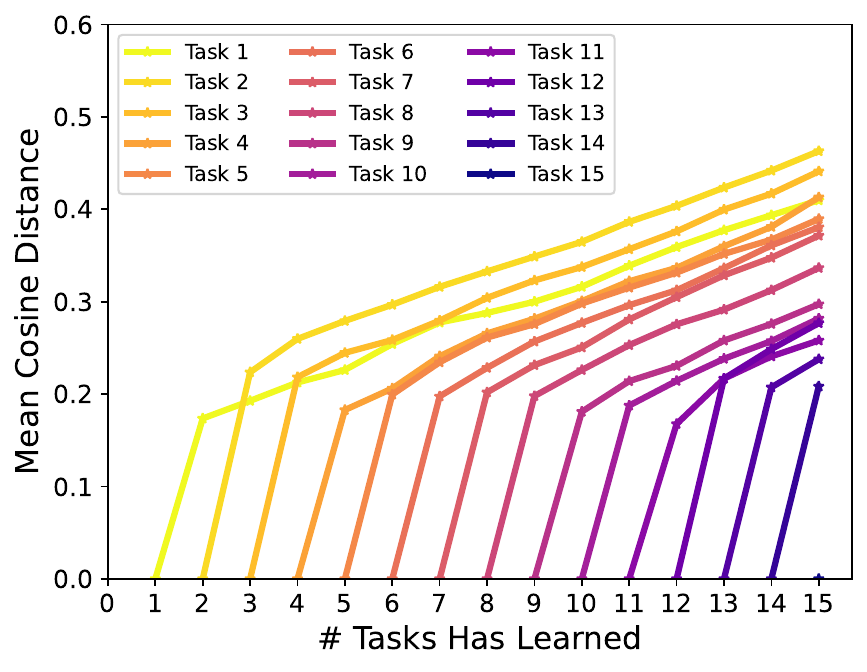}
    }
    \subfloat[Probing Classifier]{
        \includegraphics[width=0.24\linewidth]{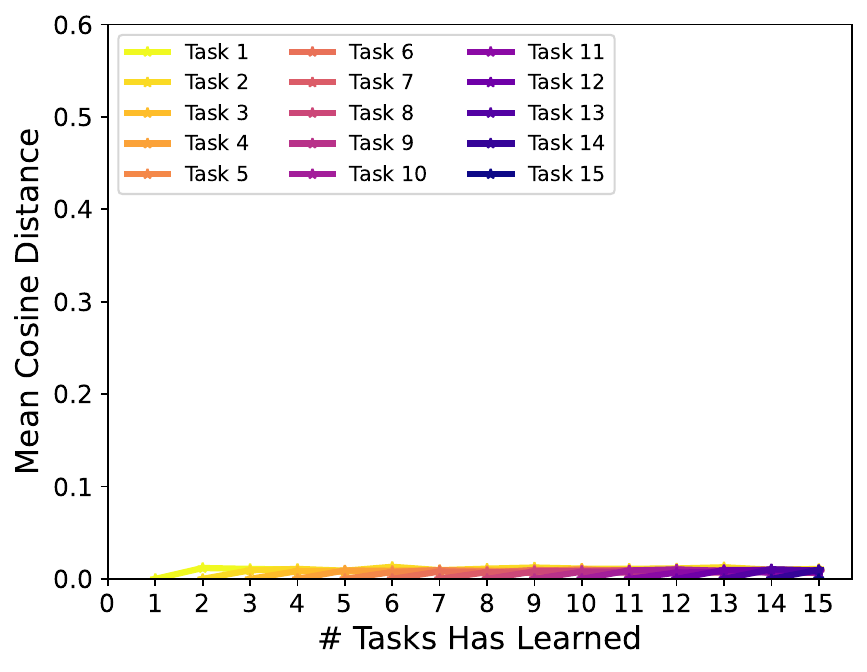}
    }
    
    \caption{Comparison between the observed and probing classifiers after SEQ on class-incremental intent classification. The backbone is bert-large-cased and \textit{frozen} during IL. (a)(b) show the average norm of the class embeddings of each task; (c)(d) show the average moving distance of the class embeddings of each task.}
    \label{fig:comparison_class_norm_moving_distance_dis_frozen}
\end{figure*}

\section{The Role of Pre-training}
\label{sec:appendix_role_of_pretraining}

The result of text classification is shown in Figure \ref{fig:probing_pretraining_tc}.
It shows a similar trend as intent classification and relation extraction in Figure \ref{fig:probing_pretraining}.

We further analyze why even a randomly-initialized model (step 0) achieves high probing performance.
We use t-SNE \cite{van2008visualizing} to visualize the features of randomly initialized models and PLMs in Figure \ref{fig:tsne_intent} and \ref{fig:tsne_RE}.

Figure \ref{fig:tsne_intent_a} shows that randomly initialized models extract discriminative features with the Transformer architecture, which is overlooked in previous IL studies.
With SEQ or pre-training, the representation ability is enhanced, and the category boundaries become clearer.
For the harder dataset, FewRel, the randomly-initialized model also learns the downstream knowledge through SEQ (Figure \ref{fig:tsne_RE_a} vs \ref{fig:tsne_RE_b}).
It explains why the performance is improved by SEQ without pre-training.

In summary, we highlight that both pre-training and the architecture of Transformers are the key factors of the anti-forgetting ability of PLMs.
Most existing studies \cite{scialom-etal-2022-fine,peng2023semiparametric} only attribute it to the pre-training stage. 
Our findings are consistent with the recent advance in the incremental learning dynamics of Transformers \cite{tarzanagh2023transformers}.
\citet{tarzanagh2023transformers} discover that the rank of attention head weights gradually increases during training.
We leave a deeper analysis of the incremental learning dynamics of the Transformers architecture to the future.

\begin{figure*}[!t]
    \centering
    \subfloat[Initial Learned]{
        \includegraphics[width=0.22\linewidth]{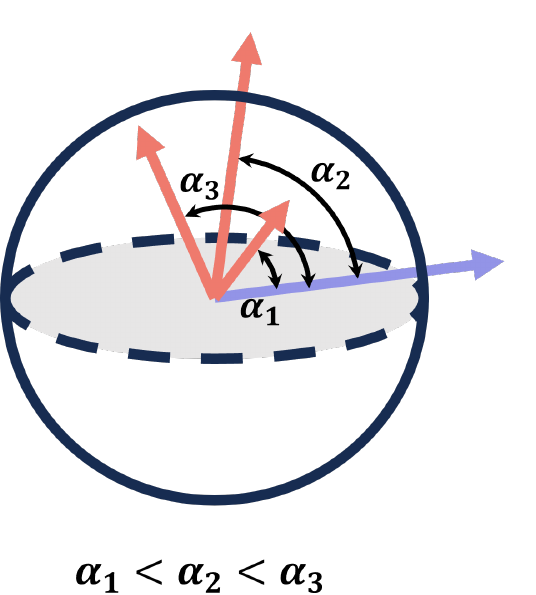}
    }
    \subfloat[Forgetting]{
        \includegraphics[width=0.20\linewidth]{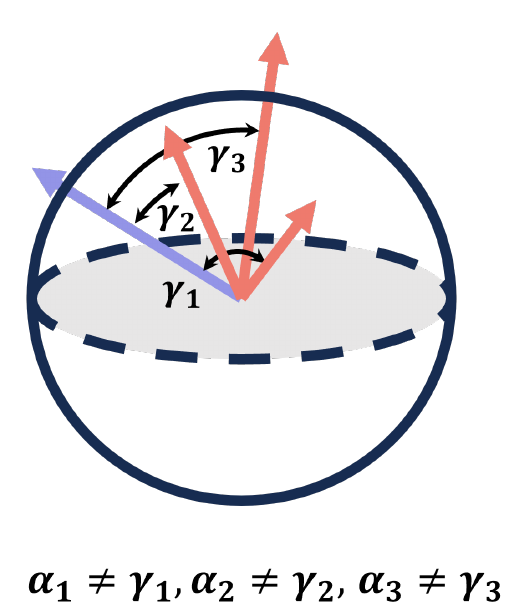}
    }
    \subfloat[No Forgetting]{
        \includegraphics[width=0.20\linewidth]{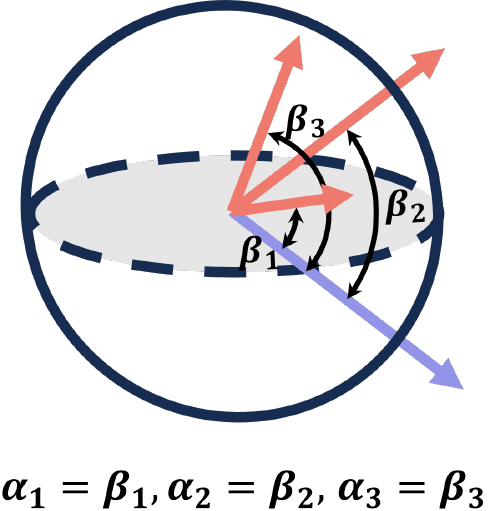}
    }
    \hspace{1cm}
    \includegraphics[width=0.20\linewidth]{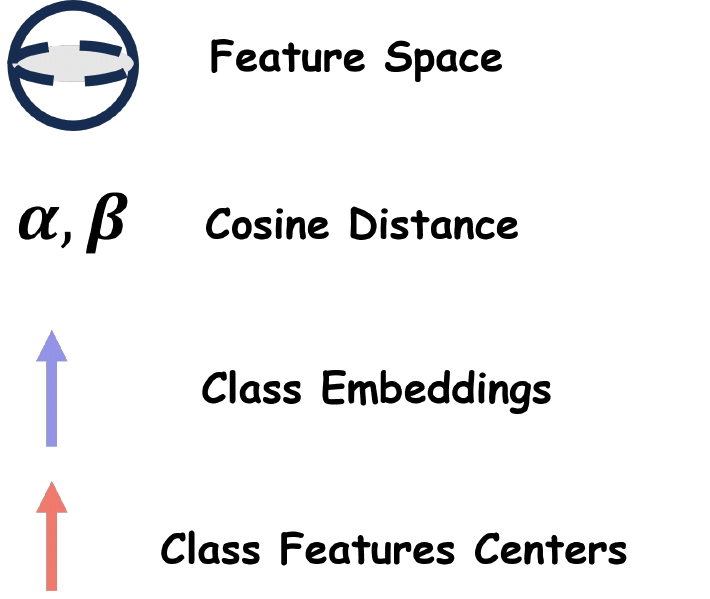}
    
    \caption{The illustration of the moving distance of class embeddings. (a) shows the cosine similarity between class embeddings and class feature centres after a new task is learned; (b) shows that forgetting happens when the relative cosine similarity is changed; (c) shows that forgetting will not happen when the relative cosine similarity is maintained.}
    \label{fig:moving_distance_illustration}
\end{figure*}

\begin{figure*}[!t]
    \centering
    \subfloat[Obs. Cls. after Task 1]{
        \includegraphics[width=0.24\linewidth]{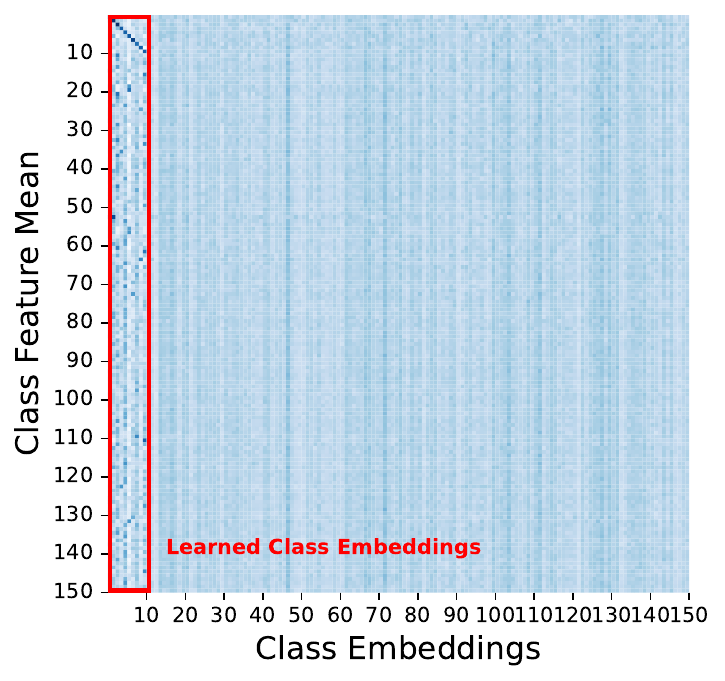}
    }
    \subfloat[Obs. Cls. after Task 5]{
        \includegraphics[width=0.24\linewidth]{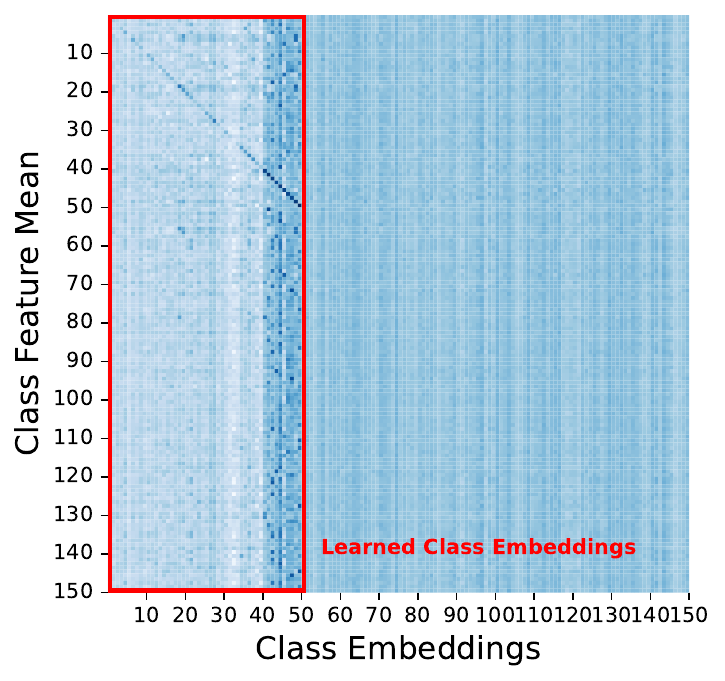}
    }
    \subfloat[Obs. Cls. after Task 10]{
        \includegraphics[width=0.24\linewidth]{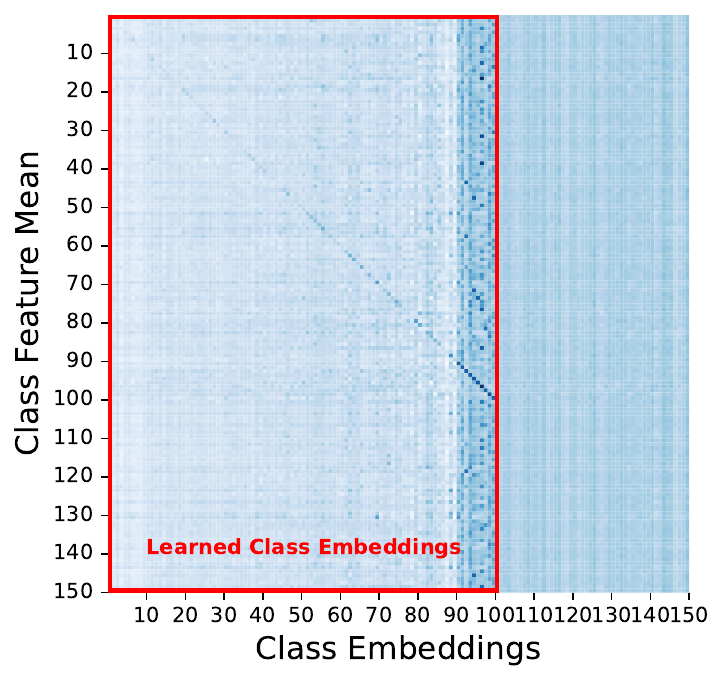}
    }
    \subfloat[Obs. Cls. after Task 15]{
        \includegraphics[width=0.24\linewidth]{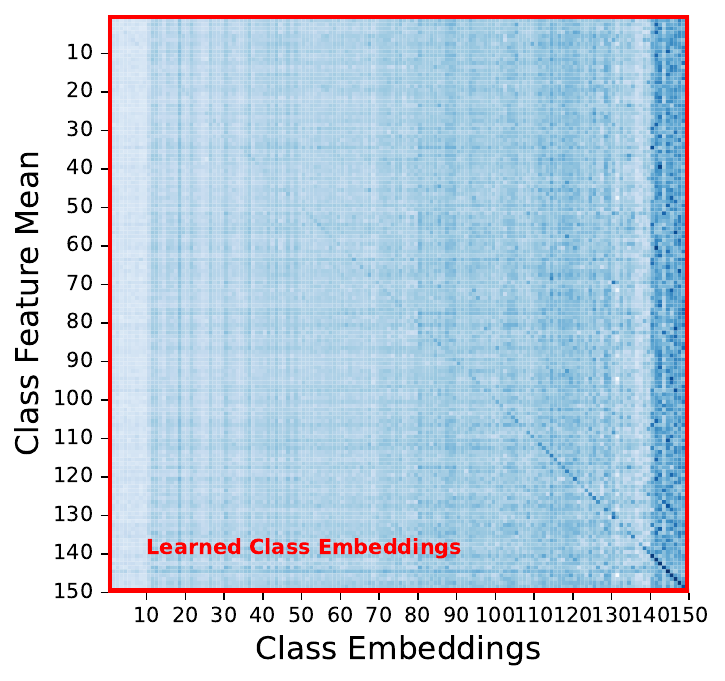}
    }

    \subfloat[Prob. Cls. after Task 1]{
        \includegraphics[width=0.24\linewidth]{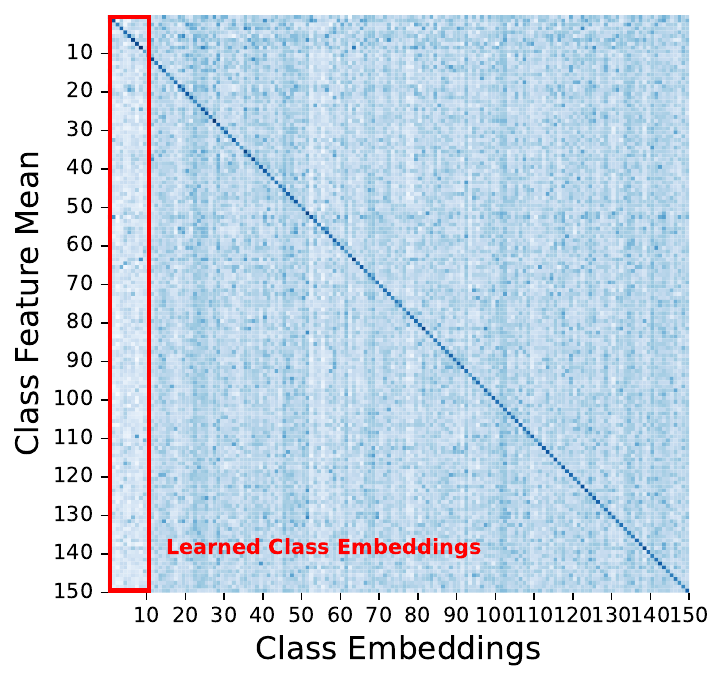}
    }
    \subfloat[Prob. Cls. after Task 5]{
        \includegraphics[width=0.24\linewidth]{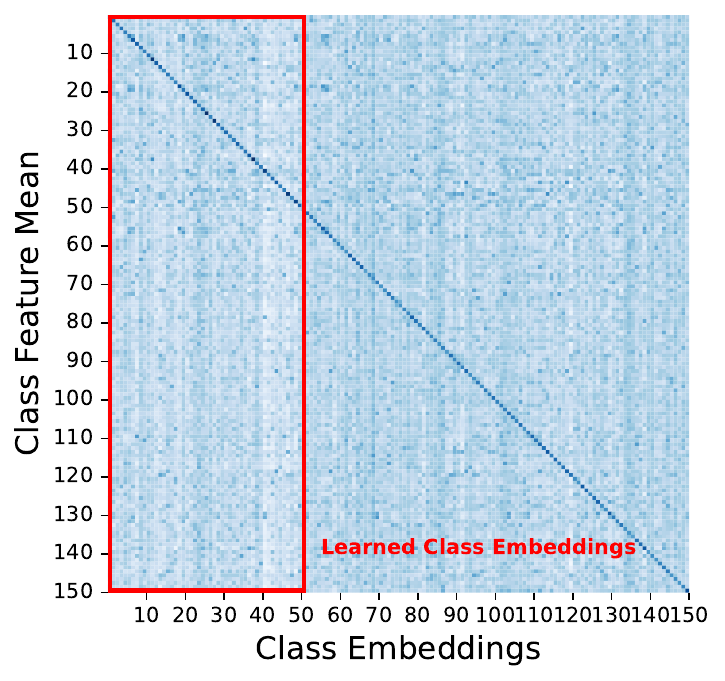}
    }
    \subfloat[Prob. Cls. after Task 10]{
        \includegraphics[width=0.24\linewidth]{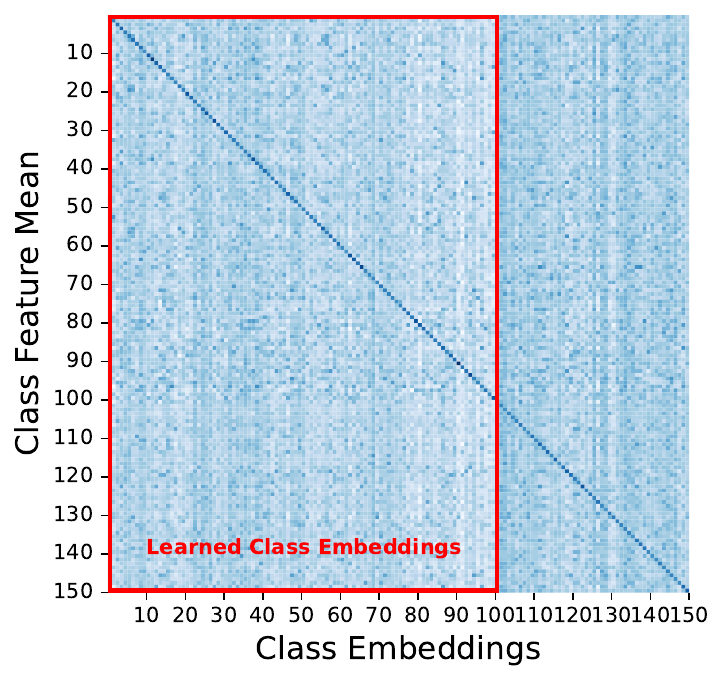}
    }
    \subfloat[Prob. Cls. after Task 15]{
        \includegraphics[width=0.24\linewidth]{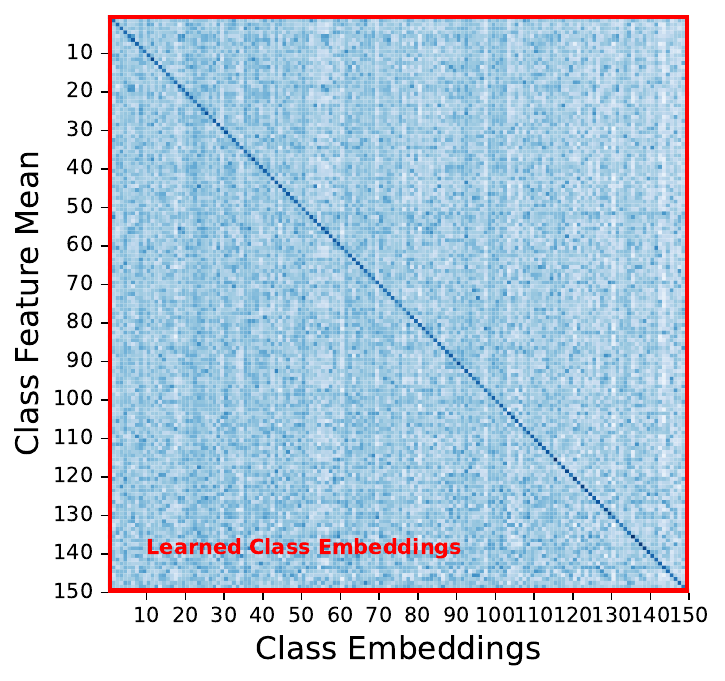}
    }
    \caption{Comparison between the cosine similarity matrix between the linear classifiers (Obs. Cls.) and probing classifiers (Prob. Cls.). The backbone is Pythia-410m. The model is trained on class-incremental intent classification and is evaluated after learning 1,5,10,15 tasks. The result shows that the class embeddings in observed classifiers change significantly compared with probing classifiers. }
    \label{fig:cos_distance_matrix_examples}
\end{figure*}

\section{The Forgetting in Classifiers}
\label{sec:appendix_forgetting_in_classifiers}

\subsection{Overview of the Forgetting in Classifiers}
Recall that we train probing classifiers on top of PLMs and achieve superior performance during SEQ.
In contrast, the observed performance has dropped consistently since the second task.
In previous studies \cite{wu2019large,hou2019learning}, they attributed the reason for catastrophic forgetting to the bias prediction between new and old categories.
Indeed, a model trained with SEQ consistently achieves high performance on the new task while the accuracy on old tasks becomes nearly zero.
In other words, the model predicts larger logits on new classes and smaller logits on old classes.
Then, we investigate the dynamics of L2 norm and cosine similarity between features and class embeddings during SEQ.

\subsection{The Bias in the L2 Norm of Class Embeddings}
We summarize the results of SEQ when the backbone is Pythia-410m in Figure \ref{fig:comparison_class_norm_moving_distance}, frozen Pythia-410m in Figure \ref{fig:comparison_class_norm_moving_distance_gen_frozen}, bert-large-cased in Figure \ref{fig:comparison_class_norm_moving_distance_dis}, and frozen bert-large-cased in Figure \ref{fig:comparison_class_norm_moving_distance_dis_frozen}.
The class embeddings in each task are sorted according to the norm for clarity.
The results show no apparent tendency in the norm of the class embeddings in both observed classifiers and probing classifiers.
The norm of the class embeddings of newer classes even decreases when the backbone is bert-large-cased.
Therefore, the norm is not the critical factor in the bias prediction towards new classes in SEQ.

\begin{figure*}[!t]
    \centering
    \subfloat[S1: warm-up and freeze PLM]{
        \includegraphics[width=0.49\linewidth]{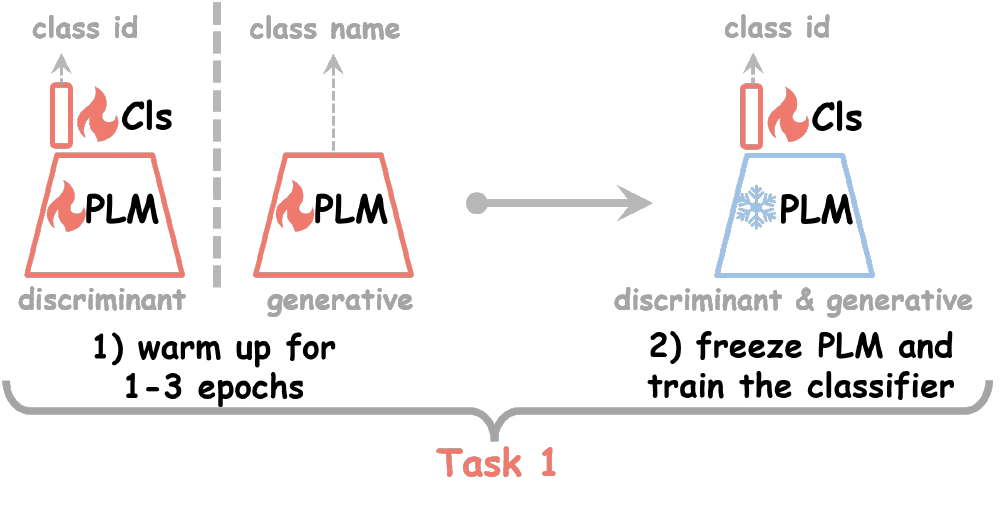}
    }
    \subfloat[S2: freeze old classifiers]{
        \includegraphics[width=0.49\linewidth]{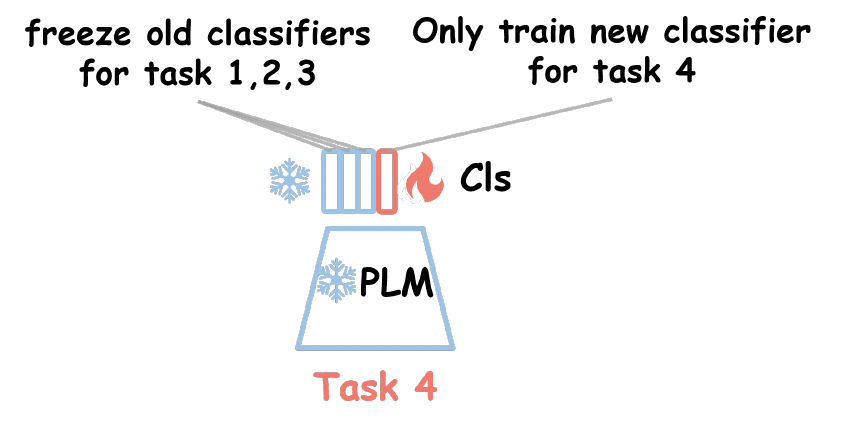}
    }

    \subfloat[S3: use proper classifiers]{
        \includegraphics[width=0.49\linewidth]{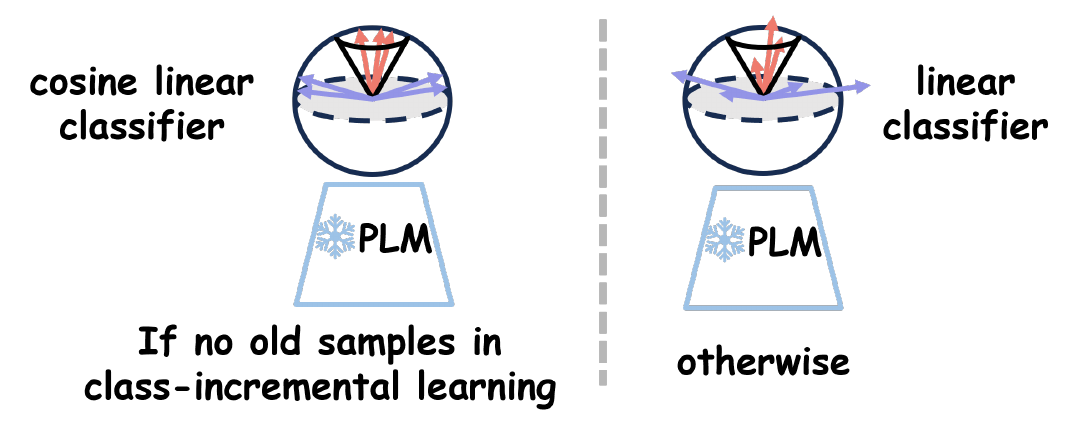}
    }
    \subfloat[S4, optional: pre-allocate future classifiers]{
        \includegraphics[width=0.49\linewidth]{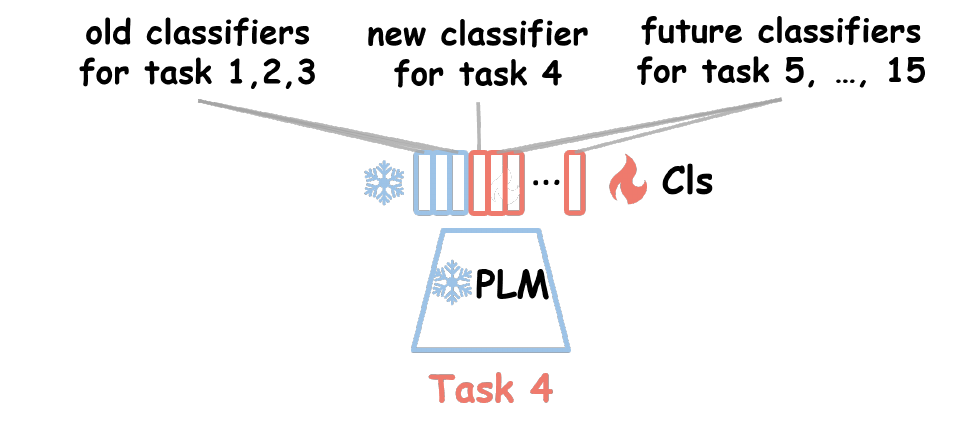}
    }
    
    \caption{An illustration of the proposed SEQ*. }
    \label{fig:illustration_seqstar}
\end{figure*}

\subsection{The Bias in the Cosine Similarity between Features and Class Embeddings}
First, we define the \textit{moving distance} to measure how the cosine similarity between features and class embeddings changes during SEQ. 
After a new task is learned, the relative position between features and class embeddings is optimal for this task.
Intuitively, the forgetting happens only when the relative position changes.
Ideally, if all pairs of features and class embeddings remain at the same angle, forgetting will not occur.
In practice, we measure the cosine similarity between class embeddings and class feature centres instead of all features.
An illustration is provided in Figure \ref{fig:moving_distance_illustration}.

Formally, we define the moving distance of class embeddings of the $i$-th task at the $i+k$-th task as follows:
\begin{equation}
    MD_{i+k}^{i} = \frac{1}{mn}\sum_{m=1}^{m=|\mathcal{Y}_{all}|}\sum_{n=1}^{n=|\mathcal{Y}_{t}|}|\mathbf{C}_{i+k}^{i}[m,n] - \mathbf{C}_{i}^{i}[m,n]|
\end{equation}
The $\mathcal{Y}_{all}$ and $\mathcal{Y}_{t}$ represent the label set of all $T$ tasks and the $t$-th task respectively.
$\mathbf{C}$ represent the cosine similarity matrix between all pairs of class embeddings and class feature centres.
$\mathbf{C}_{i+k}^{i}$ represents the cosine similarity matrix between all pairs of class embeddings from task $i$ and all class feature centres, and it is measured at task $i+k$. 
$\mathbf{C}_{i+k}^{i}[m,n]$ is the entry of the cosine similarity matrix $\mathbf{C}_{i+k}^{i}$ at position $[m,n]$.
This entry represents the cosine similarity between the $n$-th class embeddings of task $t$ and the $m$-th class embeddings of all $T$ tasks. 

We summarize the moving distance of SEQ when the backbone is Pythia-410m in Figure \ref{fig:comparison_class_norm_moving_distance}, frozen Pythia-410m in Figure \ref{fig:comparison_class_norm_moving_distance_gen_frozen}, bert-large-cased in Figure \ref{fig:comparison_class_norm_moving_distance_dis}, and frozen bert-large-cased in Figure \ref{fig:comparison_class_norm_moving_distance_dis_frozen}.
The results clearly show that the old class embeddings in observed classifiers have drifted out of position since they were learned.
Besides, the moving distance becomes larger when the backbone is frozen.
In contrast, the old class embeddings in the probing classifier do not deviate much from the initial position.

We provide instances of the cosine similarity matrix in Figure \ref{fig:cos_distance_matrix_examples}.
The cosine similarity matrix of observed classifiers changes significantly when new tasks are learned.
The forgetting is more apparent when we focus on the diagonal.

In summary, we reveal that the change of relative position between class embeddings and features leads to the forgetting of classifiers.
Additionally, the old knowledge is preserved if the relative position is maintained, as in probing classifiers.

\section{SEQ*: Boosting Performance of SEQ}
\label{sec:appendix_strategies_for_SEQ}
We illustrate the proposed four strategies in Figure \ref{fig:illustration_seqstar}.
\textbf{(S1) Freeze the PLMs after warm-up.} 
For generative backbones, we train models with causal language modelling loss to generate the ground-truth class name in the warm-up stage.
For discriminant backbones, we train models with classifiers to predict the class ID with cross-entropy loss. 
We warm up 1 epoch for discriminant backbones and 3 epochs for generative backbones.
After the warm-up stage at the first task, all parameters of backbones are frozen and will not be updated in the subsequent incremental tasks.

\textbf{(S2) Freeze the old classifiers.}
When the model finishes training on the current task, we freeze the classifier of the current task.
When learning more tasks, we only update the new classifier while the old classifiers are frozen.

\textbf{(S3) Use cosine linear classifiers only when no old data is available in the CIL scenario. Otherwise, use linear classifiers.}
We use linear classifiers even without old samples for named entity classification because the non-entity tokens (i.e., ``Other'' tokens) can be regarded as old samples from previous tasks \cite{zheng-etal-2022-distilling}.

\textbf{(S4, optional) Pre-allocate future classifiers.} 
For example, when learning the 10 classes in the first task of CLINC150, the future classifiers from task 2 to task 15 are pre-allocated.
The future classifiers are trained in advance in the first task through the softmax layer even when there are no instances from task 2 to task 15.
This strategy is especially effective for generative backbones.
The reason may be that it enhances the forward compatibility of classifiers, and thus, new classifiers are easier to adapt to new classes when old classifiers are frozen.

\section{Introduction of Baselines}
\label{sec:appendix_introduction_training_details_baselins}
Except for Topic3Datasets, we train all baselines for 5 epochs for each incremental task.
On Topic3Datasets, we train all baselines for 3 epochs for each incremental task.
The learning rate of backbones and classifiers are $1\times 10^{-5}$ and $1\times 10^{-3}$ respectively.
We use AdamW \cite{loshchilov2018fixing} optimizer.
We use RTX 3090 GPUs for our experiments.
Each experiment is repeated three times, and the average and standard deviations are reported.
We train SEQ* and all baselines with exactly the same training settings for fair comparison.
We search the best hyper-parameters for each compared method and use the same hyper-parameters across backbones and datasets.

The introduction of SOTA methods is as follows:
\begin{itemize}
    \item SEQ: SEQ refers to sequential fine-tuning. For generative backbones, SEQ trains models to output the class name with causal language modelling loss. For discriminant backbones, SEQ trains models with classifiers to predict the correct class ID. 
    \item LAMOL \cite{sun2019lamol}: LAMOL trains a generative model with question-answering and generation targets and generates pseudo samples before learning each new task. The weight of the generation target $\lambda=0.25$. The ratio of pseudo samples with respect to the training data of new task $\gamma=0.20$. The top-k sampling for generating pseudo samples $K=20$. There are two variations of LAMOL, including LAMOL\_t and LAMOL\_g. The difference is whether or not to use task-specific tokens for generation.
    \item L2KD \cite{chuang-etal-2020-lifelong}: L2KD proposes to add knowledge distillation targets based on LAMOL. We implement the word-level (Word-KD) variation since it performs best on text classification tasks.
    \item LAMOL\_KD: LAMOL\_KD further utilizes knowledge distillation based on LAMOL\_t. Unlike L2KD, the teacher model in LAMOL\_KD is trained on all previous tasks. The new data is for learning both LOAMOL targets, and the pseudo data is for word-level knowledge distillation as regularization terms. LAMOL\_KD is an extension to LAMOL.
    \item PCLL \cite{zhao-etal-2022-prompt}: PCLL is build upon LAMOL. Furthermore, PCLL utilizes the targets of varational autoencoders and word-level knowledge distillation to train generative models. 
    \item LFPT5 \cite{qin2021lfpt5}: LFPT5 further utilizes knowledge distillation based on LAMOL. Besides, LFPT5 utilizes prompt tuning instead of fine-tuning the whole model. The number of tokens for prompt tuning is 10.
    \item AdapterCL \cite{madotto-etal-2021-continual}: AdapterCL learns one adapter \cite{houlsby2019parameter} for each task. During inference in CIL, the model predicts the task ID according to causal language modelling loss, which requires $T$ forward passes for each instance.
    \item LoRA \cite{hu2021lora}: LoRA learns one LoRA adapter for each task. We only compare LoRA in the TIL setting. We set the rank $r=4$ and the scaling parameter $\alpha=8$. We use the implementation of LoRA in the PEFT library \cite{peft}. 
    \item ProgPrompt \cite{razdaibiedina2023progressive}: Progressive Prompt learns soft prompt for each task progressively. Following \cite{razdaibiedina2023progressive}, we Use a residual two-layer MLP to encode soft prompt, and the number of soft prompt tokens for each task is 5. ProgPrompt can only be used for the TIL setting because the task ID is required during inference.
    \item ER: Experience replay stores representative old samples and jointly optimizes both new and old samples when learning new tasks.
    \item DER++ \cite{buzzega2020dark}: DER++ is based on data replay. Besides, DER++ adds an MSE loss to regularize the logits of old samples between teacher and student.
    \item CLSER \cite{arani2022learning}: CLSER is based on DER++. 
    Besides, CLSER additionally stores two models (i.e., fast model and slow model) for selecting teacher logits when computing the MSE loss.
    \item SpanKL \cite{zhang2023spankl}: SpanKL converts the named entity recognition task from entity-level annotation to Span-level annotation, using a binary cross-entropy loss function and a distillation loss, with the distillation loss coefficient $\lambda$ is 1.
    \item OCILNER \cite{ma-etal-2023-learning}: OCILNER utilizes contrastive learning to adaptively detect entity clusters and utilizes two distance-based methods to label non-entity tokens. The threshold for labelling Non-entity is the median class similarity of all classes in the memory.
    \item ExtendNER \cite{monaikul2021continual}: ExtendNER utilizes knowledge distillation to review old entity types, and the distillation loss coefficient $\lambda$ is 2.
    \item DLD \cite{zhang2023decomposing}: DLD improves the knowledge distillation method in ExtendNER, dividing it into negative terms and positive terms for knowledge distillation, and the distillation loss coefficient $\lambda$ is 2.
    \item SelfTrain \cite{rosenberg2005semi, de2019continual}: SelfTrain utilizes the teacher model to directly label current data and train it together with samples of new entity types.
    \item RDP \cite{zhang2023task}: RDP utilizes the previous model to pseudo-label new data and increases self-entropy loss to improve the confidence of model prediction. The self-entropy loss coefficient $\lambda$ is 0.1.
    \item CPFD \cite{zhang2023continual}: CPFD designed a pooled feature distillation loss and proposed a confidence-based pseudo-label annotation method. The feature distillation loss coefficient $\lambda$ is 2.
    \item ICE \cite{liu-huang-2023-teamwork}: ICE includes two methods, ICE\_O and ICE\_PLO. These methods freeze the backbone model and the old classifiers. ICE\_O combine the non-entity logits with the new task logits to obtain the output probability during training. ICE\_PLO uses all previous logits and new logits during training. Unlike ICE, SEQ* additionally warm-up the PLMs and pre-allocating future classifiers. Furthermore, ICE is limited to class-incremental information extraction, while SEQ* can be applied to both sentence and word-level classification tasks. 
    \item CFNER \cite{zheng-etal-2022-distilling}: CFNER proposes a causal effect framework to alleviate the forgetting of old entity types and uses curriculum learning methods to reduce the impact of noisy labels. The number of matched tokens $K$ is 3. 
    
\end{itemize}

\section{Revisiting SOTA Methods}
\label{sec:appendix_full_results_of_SEQ*}

\subsection{CIL with Generative Backbones for Sentence-Level Tasks}
The results of SOTA methods and SEQ* with Pythia-410m, Pythia-160, gpt2-base, gpt2-large are provided in Table \ref{tab:sota_full_pythia410m_sentence}, \ref{tab:sota_full_pythia160m_sentence}, \ref{tab:sota_full_gpt2}.
We compare AdapterCL only on GPT2 because Adapter is not implemented for Pythia.
From the results, we have the following findings:

SEQ* shows competitive or even superior performance on all datasets. 
The warm-up strategy (S1) is effective when the gap between pre-training and downstream data is large (CLINC150, Banking77, FewRel, TACRED). 
In contrast, it is not so effective when the gap is small (Topic3Datasets).
Freezing old classifiers (S2) is more important than freezing backbones. 
The result also validates that the forgetting results from classifiers instead of backbone PLMs. 
Using cosine linear classifiers (S3) is crucial when the IL scenario is CIL, and no old samples are stored. 
Otherwise, models will be strongly biased towards new classes.
Pre-allocating future classifiers (S4) are very effective for generative backbones and brings considerable improvements. 
The reason may be that training future classifiers in advance reduces the overlapping between class embeddings.

The LAMOL-based methods achieve superior performance on the topic classification datasets. 
It indicates that generating pseudo samples is effective when the downstream data have close data distribution with the pre-training data. 
However, their performance is much worse than SEQ* when more tasks are learned (Clinc150), or the gap between pre-training and downstream data is larger (FewRel, TACRED). 
LFPT5 has the worst performance in our settings. 
LFPT5 was originally designed for T5 \cite{raffel2020exploring}, and we find that the decoder-only models do not learn to generate pseudo samples by just learning soft prompts. 

SEQ* fails to achieve superior performance on gpt2-base. 
We find the training loss hard to decrease when adding cosine classifiers on top of the gpt2-base. 
On gpt2-large, SEQ* outperforms AdapterCL significantly while updating much fewer parameters.

\begin{table*}[!t]
  \centering
  \caption{Comparison between SOTA methods and SEQ* on sentence-level classification tasks. The backbone is Pythia-410m. The IL scenario is CIL. No old samples are stored for all models. Other notation is the same as Table \ref{tab:sota_main_gen_pythia410m}.}
  \resizebox{0.8\linewidth}{!}{
    \begin{tabular}{lcccccccccc}
    \toprule
          & \multicolumn{2}{c}{\textbf{Topic3Datasets}} & \multicolumn{2}{c}{\textbf{CLINC150}} & \multicolumn{2}{c}{\textbf{Banking77}} & \multicolumn{2}{c}{\textbf{FewRel}} & \multicolumn{2}{c}{\textbf{TACRED}} \\
    \midrule
          & \boldmath{}\textbf{$\mathcal{A}_T$}\unboldmath{} & \boldmath{}\textbf{$\bar{\mathcal{A}}$}\unboldmath{} & \boldmath{}\textbf{$\mathcal{A}_T$}\unboldmath{} & \boldmath{}\textbf{$\bar{\mathcal{A}}$}\unboldmath{} & \boldmath{}\textbf{$\mathcal{A}_T$}\unboldmath{} & \boldmath{}\textbf{$\bar{\mathcal{A}}$}\unboldmath{} & \boldmath{}\textbf{$\mathcal{A}_T$}\unboldmath{} & \boldmath{}\textbf{$\bar{\mathcal{A}}$}\unboldmath{} & \boldmath{}\textbf{$\mathcal{A}_T$}\unboldmath{} & \boldmath{}\textbf{$\bar{\mathcal{A}}$}\unboldmath{} \\
    \midrule
    \textbf{LFPT5} & 16.78  & 39.23  & 3.48  & 19.99  & 7.98  & 19.87  & 5.52  & 9.05  & 7.60  & 15.90  \\
    \textbf{L2KD} & 58.89  & 72.82  & 22.48  & 51.23  & 47.47  & 75.11  & 37.08  & 57.29  & 20.86  & 41.15  \\
    \textbf{LAMOL\_KD} & 49.94  & 70.61  & 41.99  & 68.08  & 52.60  & 73.44  & 25.77  & 49.46  & 29.03  & 48.38  \\
    \textbf{LAMOL\_g} & \textbf{74.45 } & \underline{84.36} & 35.43  & 60.67  & 48.40  & 73.93  & 28.10  & 50.81  & 32.70  & \textbf{49.90 } \\
    \textbf{LAMOL\_t} & \underline{74.05} & \textbf{84.84 } & 43.37  & 67.77  & \underline{57.00} & \textbf{78.13 } & 28.44  & 51.19  & 28.81  & 48.41  \\
    \textbf{PCLL} & 58.83  & 74.18  & 47.09  & 71.72  & 45.33  & 72.25  & 31.00  & 51.93  & 24.50  & 51.59  \\
    \midrule
    \rowcolor{black!10}\textbf{SEQ (Lin)} & 19.66  & 45.55  & 12.26  & 28.33  & 14.88  & 34.17  & 13.43  & 35.78  & 12.64  & 32.06  \\
    \rowcolor{black!10}\textbf{SEQ (Cos)} & 16.89  & 41.68  & 5.97  & 18.27  & 11.10  & 15.62  & 11.40  & 17.76  & 10.08  & 16.75  \\
    \rowcolor{black!10}\textbf{SEQ (FixB+Cos)} & 17.13  & 42.38  & 6.08  & 18.31  & 10.32  & 12.72  & 7.45  & 19.59  & 9.30  & 15.73  \\
    \rowcolor{black!10}\textbf{SEQ (FixC+Cos)} & 50.96  & 55.69  & 64.28  & 56.54  & 44.93  & 36.40  & 33.48  & 34.74  & 28.90  & 26.61  \\
    \rowcolor{black!10}\textbf{SEQ (FixBC+Cos)} & 53.18  & 57.35  & 62.72  & 56.43  & 44.09  & 33.96  & 33.58  & 33.54  & 28.02  & 26.50  \\
    \rowcolor{black!10}\textbf{SEQ (W+FixBC+Lin)} & 33.41  & 54.47  & 19.06  & 33.90  & 17.79  & 40.93  & 13.68  & 36.14  & 13.65  & 33.78  \\
    \rowcolor{black!10}\textbf{SEQ (P+W+FixBC+Lin)} & 33.70  & 52.33  & 27.20  & 37.50  & 15.09  & 36.70  & 17.08  & 37.00  & 14.54  & 34.12  \\
    \midrule
    \rowcolor{black!20}\textbf{SEQ* (W+FixBC+Cos)} & 50.77  & 61.69  & \underline{75.96} & \underline{74.29} & 53.76  & 50.02  & \underline{46.12} & \underline{50.36} & \underline{36.55} & 33.51  \\
    \rowcolor{black!20}\textbf{SEQ* (P+W+FixBC+Cos)} & 70.56  & 83.69  & \textbf{84.51 } & \textbf{89.43 } & \textbf{67.12 } & \underline{75.54} & \textbf{61.99 } & \textbf{73.97 } & \textbf{44.34 } & \underline{48.52} \\
    \bottomrule
    \end{tabular}%
  \label{tab:sota_full_pythia410m_sentence}%
  }
\end{table*}%

\begin{table*}[!t]
  \centering
  \caption{Comparison between SOTA methods and SEQ* on sentence-level classification tasks. The backbone is Pythia-160m. The IL scenario is CIL. No old samples are stored for all models. Other notation is the same as Table \ref{tab:sota_main_gen_pythia410m}.}
  \resizebox{0.8\linewidth}{!}{
    \begin{tabular}{lcccccccccc}
    \toprule
          & \multicolumn{2}{c}{\textbf{Topic3Datasets}} & \multicolumn{2}{c}{\textbf{CLINC150}} & \multicolumn{2}{c}{\textbf{Banking77}} & \multicolumn{2}{c}{\textbf{FewRel}} & \multicolumn{2}{c}{\textbf{TACRED}} \\
    \midrule
          & \boldmath{}\textbf{$\mathcal{A}_T$}\unboldmath{} & \boldmath{}\textbf{$\bar{\mathcal{A}}$}\unboldmath{} & \boldmath{}\textbf{$\mathcal{A}_T$}\unboldmath{} & \boldmath{}\textbf{$\bar{\mathcal{A}}$}\unboldmath{} & \boldmath{}\textbf{$\mathcal{A}_T$}\unboldmath{} & \boldmath{}\textbf{$\bar{\mathcal{A}}$}\unboldmath{} & \boldmath{}\textbf{$\mathcal{A}_T$}\unboldmath{} & \boldmath{}\textbf{$\bar{\mathcal{A}}$}\unboldmath{} & \boldmath{}\textbf{$\mathcal{A}_T$}\unboldmath{} & \boldmath{}\textbf{$\bar{\mathcal{A}}$}\unboldmath{} \\
    \midrule
    \textbf{LFPT5} & 11.61  & 34.39  & 0.00  & 5.17  & 2.85  & 8.80  & 0.78  & 4.31  & 5.14  & 15.02  \\
    \textbf{L2KD} & 52.33  & 70.76  & 21.84  & 47.06  & 37.43  & 56.40  & 21.70  & 47.51  & 10.79  & 28.61  \\
    \textbf{LAMOL\_KD} & 45.66  & 65.92  & 34.05  & 56.15  & 40.39  & 61.96  & 22.35  & 42.96  & 18.08  & 37.07  \\
    \textbf{LAMOL\_g} & \underline{66.09} & \textbf{80.98 } & 32.35  & 54.50  & 44.78  & \textbf{66.15 } & 21.33  & 42.18  & 25.93  & 39.24  \\
    \textbf{LAMOL\_t} & \textbf{66.11 } & \underline{80.27} & 32.62  & 58.99  & \underline{45.93} & 64.08  & 21.09  & 44.27  & 19.75  & 40.21  \\
    \textbf{PCLL} & 44.43  & 66.43  & 39.89  & 64.26  & 41.25  & 63.07  & 26.79  & 48.76  & 21.14  & \textbf{42.53 } \\
    \midrule
    \rowcolor{black!10}\textbf{SEQ (Lin)} & 19.99  & 44.28  & 10.94  & 25.49  & 13.16  & 32.10  & 12.69  & 34.07  & 11.56  & 31.65  \\
    \rowcolor{black!10}\textbf{SEQ (Cos)} & 12.33  & 18.78  & 5.22  & 8.22  & 6.75  & 7.28  & 5.99  & 10.67  & 8.77  & 12.55  \\
    \rowcolor{black!10}\textbf{SEQ (FixB+Cos)} & 8.98  & 14.80  & 5.42  & 8.98  & 7.07  & 8.23  & 6.80  & 8.16  & 9.44  & 12.70  \\
    \rowcolor{black!10}\textbf{SEQ (FixC+Cos)} & 42.89  & 33.35  & 43.53  & 36.47  & 28.66  & 20.04  & 26.55  & 23.74  & 23.51  & 19.29  \\
    \rowcolor{black!10}\textbf{SEQ (FixBC+Cos)} & 44.57  & 32.20  & 43.11  & 37.67  & 27.01  & 22.47  & 26.80  & 22.87  & 23.96  & 20.29  \\
    \rowcolor{black!10}\textbf{SEQ (W+FixBC+Lin)} & 23.35  & 46.60  & 7.51  & 24.93  & 13.24  & 29.78  & 11.70  & 34.38  & 10.59  & 28.72  \\
    \rowcolor{black!10}\textbf{SEQ (P+W+FixBC+Lin)} & 18.45  & 45.66  & 10.02  & 23.56  & 13.73  & 31.39  & 13.54  & 33.05  & 10.90  & 29.25  \\
    \midrule
    \rowcolor{black!20}\textbf{SEQ* (W+FixBC+Cos)} & 42.30  & 47.28  & \underline{67.97} & \underline{62.37} & 45.80  & 37.40  & \underline{47.38} & \underline{50.54} & \underline{34.27} & 31.25  \\
    \rowcolor{black!20}\textbf{SEQ* (P+W+FixBC+Cos)} & 57.71  & 52.33  & \textbf{77.94 } & \textbf{83.73 } & \textbf{62.66 } & \underline{64.81} & \textbf{59.57 } & \textbf{72.23 } & \textbf{36.99 } & \underline{40.21} \\
    \bottomrule
    \end{tabular}%
    }
  \label{tab:sota_full_pythia160m_sentence}%
\end{table*}%

\begin{table*}[!t]
  \centering
  \caption{Comparison between SOTA methods and SEQ* on CLINC150. The backbone is gpt2-base and gpt2-large. The IL scenario is CIL. No old samples are stored for all models. \boldmath{}\textbf{$\mathcal{A}_{prob,0}$}\unboldmath{}: the linear probing performance before IL; \boldmath{}\textbf{$\mathcal{A}_{prob,T}$}\unboldmath{}: the linear probing performance after IL. Other notation is the same as Table \ref{tab:sota_main_gen_pythia410m}.}
  \resizebox{0.8\linewidth}{!}{
    \begin{tabular}{clccccc}
    \toprule
    \textbf{Backbone} & \multicolumn{1}{c}{\textbf{Method}} & \boldmath{}\textbf{$\mathcal{A}_T$}\unboldmath{} & \boldmath{}\textbf{$\bar{\mathcal{A}}$}\unboldmath{} & \boldmath{}\textbf{$\mathcal{A}_{prob,0}$}\unboldmath{} & \boldmath{}\textbf{$\mathcal{A}_{prob,T}$}\unboldmath{} & \textbf{\# New Params per Task } \\
    \midrule
    \multirow{4}[2]{*}{\textbf{gpt2-base}} & \textbf{AdapterCL} & 71.64 & 75.57 & 83.71 & 84.05 & 894K \\
          & \textbf{SEQ (Cos)} & 1.31  & 4.34  & 83.71 & 85.28 & 7.68K \\
          & \cellcolor{black!10}\textbf{SEQ* (W+FixBC+Cos)} & \cellcolor{black!10}4.55  & \cellcolor{black!10}9.22  & \cellcolor{black!10}83.71 & \cellcolor{black!10}86.84 & \cellcolor{black!10}7.68K \\
          & \cellcolor{black!10}\textbf{SEQ* (P+W+FixBC+Cos)} & \cellcolor{black!10}10.17 & \cellcolor{black!10}14.59 & \cellcolor{black!10}83.71 & \cellcolor{black!10}86.76 & \cellcolor{black!10}7.68K \\
    \midrule
    \multirow{4}[2]{*}{\textbf{gpt2-large}} & \textbf{AdapterCL} & 70.42 & 78.91 & 91.95 & 91.87 & 7421K \\
          & \textbf{SEQ (Cos)} & 6.48  & 20.25 & 91.95 & 93.89 & 12.8K \\
          & \cellcolor{black!10}\textbf{SEQ* (W+FixBC+Cos)} & \cellcolor{black!10}72.95 & \cellcolor{black!10}68.28 & \cellcolor{black!10}91.95 & \cellcolor{black!10}94.06 & \cellcolor{black!10}12.8K \\
          & \cellcolor{black!10}\textbf{SEQ* (P+W+FixBC+Cos)} & \cellcolor{black!10}87.93 & \cellcolor{black!10}91.41 & \cellcolor{black!10}91.95 & \cellcolor{black!10}94.18 & \cellcolor{black!10}12.8K \\
    \bottomrule
    \end{tabular}%
    }
  \label{tab:sota_full_gpt2}%
\end{table*}%

\subsection{CIL with Discriminant Backbones for Sentence-Level Tasks}
We compare SEQ* with two strong rehearsal-based methods, DER++ and CLSER.
Both DER++ and CLSER store teacher models for knowledge distillation and require updating all parameters in PLMs.

From the results in Table \ref{tab:sota_full_bert-large-cased_sentence} and \ref{tab:sota_full_bert-base-cased_sentence}, we find that both DER++ and CLSER bring considerable improvement upon ER.
However, SEQ* again shows competitive or superior performance on all datasets.
It indicates that knowledge distillation is effective for preserving knowledge. However, fully fine-tuning PLMs causes more forgetting.
Therefore, stability is more important than plasticity when designing IL algorithms for PLMs.

Besides, we have the following minor findings:
(1) Pre-allocating future classifiers with discriminant backbones is less effective than generative backbones.
(2) Freezing only old classifiers achieves the best performance on CLINC150 and TACRED. 
It shows that freezing backbone PLMs may not be necessary for some datasets.
The relative position between old class embeddings and features may still be preserved when SEQ with frozen old class embeddings.
(3) Using bert-large-cased may not lead to better results than bert-base-cased. 
We empirically find that training with bert-large-cased is more unstable than bert-base-cased.

\begin{table*}[!t]
  \centering
  \caption{Comparison between SOTA methods and SEQ* on sentence-level classification tasks. The backbone is bert-large-cased. The IL scenario is CIL. Each model stores 1 sample for each class. Other notation is the same as Table \ref{tab:sota_main_gen_pythia410m}.}
  \resizebox{0.8\linewidth}{!}{
    \begin{tabular}{lcccccccccc}
    \toprule
          & \multicolumn{2}{c}{\textbf{Topic3Datasets}} & \multicolumn{2}{c}{\textbf{CLINC150}} & \multicolumn{2}{c}{\textbf{Banking77}} & \multicolumn{2}{c}{\textbf{FewRel}} & \multicolumn{2}{c}{\textbf{TACRED}} \\
    \midrule
          & \boldmath{}\textbf{$\mathcal{A}_T$}\unboldmath{} & \boldmath{}\textbf{$\bar{\mathcal{A}}$}\unboldmath{} & \boldmath{}\textbf{$\mathcal{A}_T$}\unboldmath{} & \boldmath{}\textbf{$\bar{\mathcal{A}}$}\unboldmath{} & \boldmath{}\textbf{$\mathcal{A}_T$}\unboldmath{} & \boldmath{}\textbf{$\bar{\mathcal{A}}$}\unboldmath{} & \boldmath{}\textbf{$\mathcal{A}_T$}\unboldmath{} & \boldmath{}\textbf{$\bar{\mathcal{A}}$}\unboldmath{} & \boldmath{}\textbf{$\mathcal{A}_T$}\unboldmath{} & \boldmath{}\textbf{$\bar{\mathcal{A}}$}\unboldmath{} \\
    \midrule
    \textbf{ER} & 56.65 & 74.93 & 65.32 & 82.57 & 49.07 & 72.79 & 38.94 & 66.23 & 36.36 & 54.69 \\
    \textbf{CLSER} & 56.36 & 74.68 & \underline{73.12} & 86.80  & 50.07 & 70.85 & 45.59 & 67.9  & 42.53 & 57.41 \\
    \textbf{DER++} & 61.49 & 78.83 & 72.7  & 86.49 & 56.15 & 76.67 & 44.41 & 68.32 & 42.5  & \underline{59.05} \\
    \midrule
    \rowcolor{black!10}\textbf{SEQ (Lin)} & 20.28 & 45.4  & 7.62  & 25.45 & 13.19 & 22.78 & 14.72 & 36.11 & 13.6  & 33.08 \\
    \rowcolor{black!10}\textbf{SEQ (Cos)} & 17.55 & 24.17 & 8.33  & 15.56 & 10.04 & 12.4  & 9.55  & 15.8  & 9.46  & 14.79 \\
    \rowcolor{black!10}\textbf{SEQ (FixB+Lin)} & 44.58  & 67.70  & 24.35  & 57.02  & 20.29  & 49.23  & 14.09  & 35.62  & 12.50  & 26.12  \\
    \rowcolor{black!10}\textbf{SEQ (FixC+Lin)} & 58.24  & 76.63  & \textbf{78.86 } & \textbf{89.90 } & 58.34  & 75.17  & 44.36  & 69.77  & \textbf{45.26 } & \textbf{60.46 } \\
    \rowcolor{black!10}\textbf{SEQ (FixBC+Lin)} & 61.20  & 80.52  & 47.11  & 70.17  & 46.29  & 64.24  & 29.43  & 49.85  & 23.14  & 32.73  \\
    \rowcolor{black!10}\textbf{SEQ (W+FixBC+Cos)} & 36.52 & 59.05 & 44.9  & 58.49 & 31.09 & 51.25 & 34.15 & 49.52 & 23.8  & 40.78 \\
    \rowcolor{black!10}\textbf{SEQ (P+W+FixBC+Cos)} & 33.48 & 55.2  & 10.95 & 12.44 & 14.53 & 25.84 & 21.7  & 35.71 & 16.9  & 30.78 \\
    \midrule
    \rowcolor{black!20}\textbf{SEQ* (W+FixBC+Lin)} & \underline{62.63} & \underline{80.63} & \underline{73.12} & \underline{86.81} & \textbf{62.02} & \textbf{80.04} & \textbf{51.57} & \textbf{72.5} & \underline{43.8} & 58.47 \\
    \rowcolor{black!20}\textbf{SEQ* (P+W+FixBC+Lin)} & \textbf{64.39} & \textbf{81.92} & 72.52 & 86.49 & \underline{61.15} & \underline{79.42} & \underline{49.71} & \underline{70.87} & 41.98 & 58.2 \\
    \bottomrule
    \end{tabular}%
    }
  \label{tab:sota_full_bert-large-cased_sentence}%
\end{table*}%

\begin{table*}[!t]
  \centering
  \caption{Comparison between SOTA methods and SEQ* on sentence-level classification tasks. The backbone is bert-base-cased. The IL scenario is CIL. Each model stores 1 sample for each class. Other notation is the same as Table \ref{tab:sota_main_gen_pythia410m}.}
  \resizebox{0.8\linewidth}{!}{
    \begin{tabular}{lcccccccccc}
    \toprule
          & \multicolumn{2}{c}{\textbf{Topic3Datasets}} & \multicolumn{2}{c}{\textbf{CLINC150}} & \multicolumn{2}{c}{\textbf{Banking77}} & \multicolumn{2}{c}{\textbf{FewRel}} & \multicolumn{2}{c}{\textbf{TACRED}} \\
    \midrule
          & \boldmath{}\textbf{$\mathcal{A}_T$}\unboldmath{} & \boldmath{}\textbf{$\bar{\mathcal{A}}$}\unboldmath{} & \boldmath{}\textbf{$\mathcal{A}_T$}\unboldmath{} & \boldmath{}\textbf{$\bar{\mathcal{A}}$}\unboldmath{} & \boldmath{}\textbf{$\mathcal{A}_T$}\unboldmath{} & \boldmath{}\textbf{$\bar{\mathcal{A}}$}\unboldmath{} & \boldmath{}\textbf{$\mathcal{A}_T$}\unboldmath{} & \boldmath{}\textbf{$\bar{\mathcal{A}}$}\unboldmath{} & \boldmath{}\textbf{$\mathcal{A}_T$}\unboldmath{} & \boldmath{}\textbf{$\bar{\mathcal{A}}$}\unboldmath{} \\
    \midrule
    \textbf{ER} & 51.90  & 72.76  & 63.11  & 82.33  & 49.76  & 73.20  & 40.32  & 64.02  & 35.50  & 53.17  \\
    \textbf{CLSER} & 57.23  & 75.67  & \underline{71.77}  & \underline{86.46 } & 51.17  & 74.75  & 43.33  & 67.63  & \underline{42.13 } & 55.24  \\
    \textbf{DER++} & \textbf{64.78 } & 79.29  & 69.05 & 86.13 & 55.84  & 76.63  & 44.67  & 67.73  & 40.58  & 56.11  \\
    \midrule
    \rowcolor{black!10}\textbf{SEQ (Lin)} & 23.62  & 48.38  & 11.09  & 30.88  & 14.07  & 23.78  & 15.16  & 36.29  & 12.77  & 35.60  \\
    \rowcolor{black!10}\textbf{SEQ (Cos)} & 17.65  & 25.84  & 7.06  & 14.77  & 9.48  & 11.80  & 8.54  & 13.78  & 9.98  & 16.54  \\
    \rowcolor{black!10}\textbf{SEQ (FixB+Lin)} & 39.98  & 64.44  & 25.26  & 57.33  & 21.07  & 49.61  & 14.38  & 36.56  & 14.36  & 26.20  \\
    \rowcolor{black!10}\textbf{SEQ (FixC+Lin)} & 53.90  & 74.28  & \textbf{78.08}  & \textbf{89.03 }  & 59.64  & 77.19  & 46.06  & 66.87  & \textbf{45.65}  & \textbf{57.79}  \\
    \rowcolor{black!10}\textbf{SEQ (FixBC+Lin)} & 61.81  & 79.63  & 44.97  & 69.10  & 40.77  & 63.96  & 32.24  & 50.07  & 21.26  & 29.28  \\
    \rowcolor{black!10}\textbf{SEQ (W+FixBC+Cos)} & 31.91  & 54.39  & 13.11  & 14.09  & 15.74  & 25.82  & 18.47  & 31.04  & 22.53  & 40.77  \\
    \rowcolor{black!10}\textbf{SEQ (P+W+FixBC+Cos)} & 26.45  & 51.71  & 40.23  & 53.29  & 30.73  & 49.88  & 33.96  & 49.64  & 14.86  & 23.29  \\
    \midrule
    \rowcolor{black!20}\textbf{SEQ* (W+FixBC+Lin)} & 63.12  & \textbf{81.77 } & 69.30  & 84.21  & \textbf{61.95 } & \underline{80.18} & \textbf{53.66 } & \textbf{72.56 } & 40.63  & 56.44 \\
    \rowcolor{black!20}\textbf{SEQ* (P+W+FixBC+Lin)} & \underline{63.58} & \underline{80.39} & 68.50  & 83.24  & \underline{61.94} & \textbf{80.20 } & \underline{52.76} & \underline{72.41} & 42.08 & \underline{56.95 } \\
    \bottomrule
    \end{tabular}%
    }
  \label{tab:sota_full_bert-base-cased_sentence}%
\end{table*}%

\subsection{CIL with Discriminant Backbones for Word-Level Tasks}
We evaluate SEQ* on class-incremental named entity recognition since its popularity.
In class-incremental named entity recognition, the non-entity tokens (``Other'' tokens) can be regarded as old samples \cite{zheng-etal-2022-distilling}.
Existing methods such as \cite{monaikul2021continual,zheng-etal-2022-distilling} utilize the Other tokens for knowledge distillation.
And \citet{zhang2023task}, \citet{zhang2023decomposing}, \citet{zhang2023continual} also verify the effectiveness of applying knowledge distillation.
However, the results in Table \ref{tab:sota_full_bert-large-cased_word}, \ref{tab:sota_full_bert-base-cased_word} show that SEQ* outperforms them by a large margin on OntoNotes5 and I2B2.
On the challenging dataset Few-NERD, SEQ* beat 7 SOTA methods for class-incremental named entity recognition.

When using bert-large-cased, the performance of CFNER, DLD, and SpanKL becomes more unstable.
In contrast, SEQ* are more robust to the choice of PLMs and show consistent superior performance.

Finally, we also find that the choice of classifier is crucial for class-incremental named entity recognition.
When using cosine linear classifiers, the performance is degraded significantly.
Therefore, we use linear classifiers for all baselines and SEQ* for a fair comparison, although CFNER, DLD, RDP, and CPFD adopt cosine linear classifiers in their original implementation.
We speculate the reason is that using cosine linear classifiers prevents models from learning the prior distribution between classes, which is crucial for class-imbalanced tasks. 

The result in Table \ref{tab:compare_linear_probing_sota_dis} shows that the linear probing performance of both SEQ* and SOTA methods increases.
It indicates that all models enable PLMs to adapt to downstream tasks.
Furthermore, SEQ*, SelfTrain and ExtendNER have higher linear probing performance than SEQ (Lin). 
It indicates that forgetting causes the degradation of linear probing performance, and it can be alleviated by freezing PLMs or knowledge distillation.

\begin{table*}[!t]
  \centering
  \caption{Comparison between SOTA methods and SEQ* on word-level classification tasks. The backbone is bert-large-cased. The IL scenario is CIL. No old samples are stored for all models. Other notation is the same as Table \ref{tab:sota_main_gen_pythia410m}.}
  \resizebox{0.8\linewidth}{!}{
    \begin{tabular}{lcccccc}
        \toprule
          & \multicolumn{2}{c}{\textbf{Few-NERD}} & \multicolumn{2}{c}{\textbf{OntoNotes5}} & \multicolumn{2}{c}{\textbf{I2B2}} \\
    \midrule
          & \boldmath{}\textbf{$\mathcal{A}_T$}\unboldmath{} & \boldmath{}\textbf{$\bar{\mathcal{A}}$}\unboldmath{} & \boldmath{}\textbf{$\mathcal{A}_T$}\unboldmath{} & \boldmath{}\textbf{$\bar{\mathcal{A}}$}\unboldmath{} & \boldmath{}\textbf{$\mathcal{A}_T$}\unboldmath{} & \boldmath{}\textbf{$\bar{\mathcal{A}}$}\unboldmath{} \\
    \midrule
    \textbf{SpanKL} & 0.00    & 0.00    & 0.00  & 0.00  & 0.00  & 1.98  \\
    \textbf{OCILNER} & 22.83  & 31.78  & 37.67  & 53.58  & 36.94  & 52.12  \\
    \textbf{ExtendNER} & 20.69  & 31.77  & 46.46  & 57.65  & 26.22  & 44.62  \\
    \textbf{DLD} & 21.46  & 38.42  & 48.96  & 58.17  & 0.00  & 28.75  \\
    \textbf{SelfTrain} & 25.73  & 40.48  & 49.31  & 57.70  & 36.83  & 55.00  \\
    \textbf{RDP} & \underline{30.66} & \underline{45.93} & 54.41  & 64.30  & 42.96  & 61.59  \\
    \textbf{CPFD} & \textbf{35.65 } & \textbf{48.88 } & 59.02  & 64.51  & 24.75  & 52.98  \\
    \textbf{ICE\_O} & 25.61  & 29.35  & 47.19  & 48.21  & 53.03  & 55.07  \\
    \textbf{ICE\_PLO} & 17.54  & 22.11  & 42.53  & 46.16  & 47.76  & 53.21  \\
    \textbf{CFNER} & 29.90  & 44.02  & 48.62  & 57.96  & 1.23  & 27.05  \\
    \midrule
     \rowcolor{black!10}\textbf{SEQ (Lin)} & 3.31  & 16.49  & 4.42  & 21.42  & 4.77  & 25.46  \\
     \rowcolor{black!10}\textbf{SEQ (W+FixBC+Cos)} & 9.21  & 21.40  & 32.60  & 50.51  & 43.64  & 59.50  \\
     \rowcolor{black!10}\textbf{SEQ (P+W+FixBC+Cos)} & 5.09  & 18.25  & 33.95  & 50.94  & 46.25  & 61.71  \\
    \midrule
     \rowcolor{black!20}\textbf{SEQ* (W+FixBC+Lin)} & 27.72  & 42.27  & \underline{67.21} & \underline{73.23} & \textbf{74.76 } & \textbf{77.17 } \\
     \rowcolor{black!20}\textbf{SEQ* (P+W+FixBC+Lin)} & 28.52  & 42.57  & \textbf{68.87 } & \textbf{73.70 } & \underline{72.66} & \underline{75.79} \\
    \bottomrule
    \end{tabular}%
    }
  \label{tab:sota_full_bert-large-cased_word}%
\end{table*}%

\begin{table*}[!t]
  \centering
  \caption{Comparison between SOTA methods and SEQ* on word-level classification tasks. The backbone is bert-base-cased. The IL scenario is CIL. No old samples are stored for all models. Other notation is the same as Table \ref{tab:sota_main_gen_pythia410m}.}
  \resizebox{0.8\linewidth}{!}{
    \begin{tabular}{lcccccc}
        \toprule
          & \multicolumn{2}{c}{\textbf{Few-NERD}} & \multicolumn{2}{c}{\textbf{OntoNotes5}} & \multicolumn{2}{c}{\textbf{I2B2}} \\
    \midrule
          & \boldmath{}\textbf{$\mathcal{A}_T$}\unboldmath{} & \boldmath{}\textbf{$\bar{\mathcal{A}}$}\unboldmath{} & \boldmath{}\textbf{$\mathcal{A}_T$}\unboldmath{} & \boldmath{}\textbf{$\bar{\mathcal{A}}$}\unboldmath{} & \boldmath{}\textbf{$\mathcal{A}_T$}\unboldmath{} & \boldmath{}\textbf{$\bar{\mathcal{A}}$}\unboldmath{} \\
    \midrule
    \textbf{SpanKL} & 18.26  & 39.28  & 40.10  & 53.88  & 6.12  & 34.84  \\
    \textbf{OCILNER} & 18.44  & 29.80  & 39.99  & 54.70  & 27.27  & 47.74  \\
    \textbf{ExtendNER} & 20.02  & 36.34  & 48.08  & 57.02  & 20.02  & 36.34  \\
    \textbf{DLD} & 20.75  & 35.53  & 47.23  & 58.04  & 30.50  & 48.03  \\
    \textbf{SelfTrain} & 23.46  & 39.88  & 51.08  & 57.41  & 23.60  & 39.49  \\
    \textbf{RDP} & 27.08  & \underline{43.43} & 50.45  & 60.32  & 40.38  & 58.12  \\
    \textbf{CPFD} & \textbf{34.65 } & \textbf{46.92 } & 55.58  & 59.34  & 43.52  & 56.15  \\
    \textbf{ICE\_O} & \underline{28.98} & 33.94  & 51.81  & 53.06  & 49.12  & 54.56  \\
    \textbf{ICE\_PLO} & 19.94  & 29.31  & 46.52  & 49.79  & 47.76  & 53.35  \\
    \textbf{CFNER} & 27.70  & 42.13  & 58.07  & 63.76  & 35.42  & 51.44  \\
    \midrule
    \rowcolor{black!10}\textbf{SEQ (Lin)} & 2.97  & 16.21  & 4.38  & 21.20  & 5.26  & 25.26  \\
    \rowcolor{black!10}\textbf{SEQ (W+FixBC+Cos)} & 7.26  & 19.72  & 29.12  & 47.60  & 45.95  & 61.29  \\
    \rowcolor{black!10}\textbf{SEQ (P+W+FixBC+Cos)} & 3.17  & 19.65  & 29.70  & 48.30  & 47.10  & 60.42  \\
    \midrule
    \rowcolor{black!20}\textbf{SEQ* (W+FixBC+Lin)} & 28.13  & 42.72  & \underline{66.99} & \underline{71.80} & \underline{71.76} & \underline{73.71} \\
    \rowcolor{black!20}\textbf{SEQ* (P+W+FixBC+Lin)} & 28.21  & 43.06  & \textbf{67.39 } & \textbf{72.27 } & \textbf{72.51 } & \textbf{75.48 } \\
    \bottomrule
    \end{tabular}%
    }
  \label{tab:sota_full_bert-base-cased_word}%
\end{table*}%

\begin{table*}[!t]
  \centering
  \caption{The linear probing performance with backbone bert-base-cased. Other settings are the same as Table \ref{tab:sota_full_bert-base-cased_word}.}
  \resizebox{0.8\linewidth}{!}{
    \begin{tabular}{lcccc}
    \toprule
          & \multicolumn{2}{c}{\textbf{OntoNotes5}} & \multicolumn{2}{c}{\textbf{I2B2}} \\
\cmidrule{2-5}          & \textbf{Before IL} & \textbf{After IL} & \textbf{Before IL} & \textbf{After IL} \\
    \midrule
    \textbf{SEQ (Lin)} & \multirow{4}[2]{*}{52.89\small{±0.41}} & 71.93\small{±1.04} & \multirow{4}[2]{*}{58.18\small{±0.92}} & 73.40\small{±0.94} \\
    \textbf{SelfTrain} &       & 74.29\small{±0.75} &       & 74.54\small{±0.57} \\
    \textbf{ExtendNER} &       & 73.96\small{±0.85} &       & 76.40\small{±0.55} \\
    \textbf{SEQ* (P+W+FixBC+Lin)} &       & 74.05\small{±1.02} &       & 75.08\small{±1.15} \\
    \bottomrule
    \end{tabular}%
    }
  \label{tab:compare_linear_probing_sota_dis}%
\end{table*}%

\subsection{TIL for Sentence-Level Tasks}
We consider LoRA and ProgPrompt for the baselines in TIL.
LoRA has been widely adopted for adapting PLMs to downstream tasks.
The result in Table \ref{tab:sota_full_til} shows that SEQ* achieves competitive performance but requires much fewer new parameters to learn compared with LoRA. 
ProgPrompt relies heavily on the choice of PLMs and shows poor performance on Pythia-410m and 160m.
Since the scenario of TIL is much simpler than that of CIL, and all methods achieve high accuracy, we did not conduct as many experiments in TIL as in CIL.

\begin{table*}[!t]
  \centering
  \caption{Comparison between SOTA methods and SEQ* on CLINC150 and FewRel. The backbone is Pythia-410m, Pythia-160m, bert-large-cased and bert-base-cased. The IL scenario is TIL. No old samples are stored for all models. Other notation is the same as Table \ref{tab:sota_main_gen_pythia410m}.}
  \resizebox{0.8\linewidth}{!}{
        \begin{tabular}{clcccccc}
    \toprule
    \multirow{2}[4]{*}{\textbf{Backbone}} & \multicolumn{1}{c}{\multirow{2}[4]{*}{\textbf{Method}}} & \multicolumn{3}{c}{\textbf{CLINC150}} & \multicolumn{3}{c}{\textbf{FewRel}} \\
\cmidrule{3-8}          &       & \boldmath{}\textbf{$\mathcal{A}_T$}\unboldmath{} & \boldmath{}\textbf{$\bar{\mathcal{A}}$}\unboldmath{} & \textbf{\# New Params per Task} & \boldmath{}\textbf{$\mathcal{A}_T$}\unboldmath{} & \boldmath{}\textbf{$\bar{\mathcal{A}}$}\unboldmath{} & \textbf{\# New Params per Task} \\
    \midrule
    \multirow{4}[2]{*}{\textbf{Pythia-410m}} & \textbf{LoRA} & 97.04  & 98.60  & 393K  & 95.29  & 96.62  & 393K \\
          & \textbf{ProgPrompt} & 57.91  & 98.00  & 5.12K & 46.73  & 55.05  & 5.12K \\
          & \cellcolor{black!10}\textbf{SEQ* (W+FixBC+Lin)} & \cellcolor{black!10}98.04  & \cellcolor{black!10}98.01  & \cellcolor{black!10}10.24K & \cellcolor{black!10}90.02  & \cellcolor{black!10}93.03  & \cellcolor{black!10}10.24K \\
          & \cellcolor{black!10}\textbf{SEQ* (P+W+FixBC+Lin)} & \cellcolor{black!10}98.27  & \cellcolor{black!10}98.02  & \cellcolor{black!10}10.24K & \cellcolor{black!10}91.25  & \cellcolor{black!10}93.61  & \cellcolor{black!10}10.24K \\
    \midrule
    \multirow{4}[2]{*}{\textbf{Pythia-160m}} & \textbf{LoRA} & 51.62  & 50.81  & 147K  & 83.93  & 87.35  & 147K \\
          & \textbf{ProgPrompt} & 13.00  & 27.03  & 3.84K & 14.79  & 22.36  & 3.84K \\
          & \cellcolor{black!10}\textbf{SEQ* (W+FixBC+Lin)} & \cellcolor{black!10}95.94  & \cellcolor{black!10}96.75  & \cellcolor{black!10}7.68K & \cellcolor{black!10}88.42  & \cellcolor{black!10}91.35  &\cellcolor{black!10} 7.68K \\
          & \cellcolor{black!10}\textbf{SEQ* (P+W+FixBC+Lin)} & \cellcolor{black!10}96.42  & \cellcolor{black!10}97.09  & \cellcolor{black!10}7.68K & \cellcolor{black!10}88.62  & \cellcolor{black!10}92.23  & \cellcolor{black!10}7.68K \\
    \midrule
    \multirow{4}[2]{*}{\textbf{bert-large-cased}} & \textbf{LoRA} & 98.86  & 98.97  & 393K  & 93.63  & 95.27  & 393K \\
          & \boldmath{}\textbf{ProgPrompt$^{\dagger}$}\unboldmath{} & 94.82  & 96.85  & 15.36K & 90.96  & 92.90  & 15.36K \\
          & \cellcolor{black!10}\textbf{SEQ* (W+FixBC+Lin)} & \cellcolor{black!10}97.80  & \cellcolor{black!10}97.80  & \cellcolor{black!10}10.24K & \cellcolor{black!10}85.56  & \cellcolor{black!10}87.75  & \cellcolor{black!10}10.24K \\
          & \cellcolor{black!10}\textbf{SEQ* (P+W+FixBC+Lin)} & \cellcolor{black!10}97.70  & \cellcolor{black!10}97.64  & \cellcolor{black!10}10.24K & \cellcolor{black!10}85.93  & \cellcolor{black!10}88.29  & \cellcolor{black!10}10.24K \\
    \midrule
    \multirow{4}[2]{*}{\textbf{bert-base-cased}} & \textbf{LoRA} & 98.53  & 98.46  & 147K  & 94.00  & 95.42  & 147K \\
          & \boldmath{}\textbf{ProgPrompt$^{\dagger}$}\unboldmath{} & 98.15  & 98.25  & 11.52K & 92.38  & 94.03  & 11.52K \\
          & \cellcolor{black!10}\textbf{SEQ* (W+FixBC+Lin)} & \cellcolor{black!10}96.71  & \cellcolor{black!10}96.64  & \cellcolor{black!10}7.68K & \cellcolor{black!10}86.50  & \cellcolor{black!10}88.64  & \cellcolor{black!10}7.68K \\
          & \cellcolor{black!10}\textbf{SEQ* (P+W+FixBC+Lin)} & \cellcolor{black!10}96.30  & \cellcolor{black!10}96.20  & \cellcolor{black!10}7.68K & \cellcolor{black!10}86.65  & \cellcolor{black!10}88.73  & \cellcolor{black!10}7.68K \\
    \bottomrule
    \end{tabular}%
    }
  \label{tab:sota_full_til}%
\end{table*}%

\end{document}